\documentclass{article}

\usepackage{PRIMEarxiv}

\usepackage[utf8]{inputenc} 
\usepackage[T1]{fontenc}    
\usepackage{hyperref}       
\usepackage{url}  
\usepackage{amsmath}
\usepackage[ruled,vlined]{algorithm2e}
\usepackage{algorithmic}
\usepackage{booktabs}       
\usepackage{amsfonts}       
\usepackage{nicefrac} 
\usepackage{xcolor}
\usepackage{microtype}      
\usepackage{longtable}
\usepackage{siunitx}
\usepackage{subcaption}
\usepackage{lipsum}
\newsavebox{\algobox}
\usepackage{float} 
\usepackage{graphicx}
\graphicspath{{media/}}  

\title{Predicting Stock Prices using Permutation Decision Trees and Strategic Trailing
}

\author{
  Vishrut Ramraj \\
  Department of Computer Science and Information Systems \\
  BITS Pilani K K Birla Goa Campus \\
  Goa, 403726, India \\
  \texttt{vishrut172@gmail.com} \\
  \And
  Nithin Nagaraj \\
  Complex Systems Programme \\
  National Institute of Advanced Studies, IISc Campus \\
  Bengaluru, 560012, Karnataka, India \\
  \texttt{nithin@nias.res.in} \\
  \And
  Harikrishnan N B \\
  Department of Computer Science and Information Systems \\
  BITS Pilani K K Birla Goa Campus \\
  Goa, 403726, India \\
  Adjunct Faculty, Consciousness Studies Programme \\
  National Institute of Advanced Studies, IISc Campus \\
  Bengaluru, 560012, Karnataka, India \\
  \texttt{harikrishnannb@goa.bits-pilani.ac.in} \\
}

\begin{document}
\maketitle

\begin{abstract}
In this paper, we explore the application of Permutation Decision Trees (PDT) and strategic trailing for predicting stock market movements and executing profitable trades in the Indian stock market. We focus on high-frequency data using 5-minute candlesticks for the top 50 stocks listed in the NIFTY 50 index and Forex pairs such as XAUUSD and EURUSD. We implement a trading strategy that aims to buy stocks at lower prices and sell them at higher prices, capitalizing on short-term market fluctuations. Due to regulatory constraints in India, short selling is not considered in our strategy. The model incorporates various technical indicators and employs hyperparameters such as the trailing stop-loss value and support thresholds to manage risk effectively. We trained and tested data on a 3 month dataset provided by Yahoo Finance. Our bot based on Permutation Decision Tree achieved a profit of 1.1802\% over the testing period, where as a bot based on LSTM gave a return of 0.557\% over the testing period and a bot based on RNN gave a return of 0.5896\% over the testing period. All of the bots outperform the buy-and-hold strategy, which resulted in a loss of 2.29\%.
\end{abstract}


\section{Introduction}
 Stock market trading has garnered enormous interest due to its potential for generating revenue. Successful trading combines both intuition, developed through studying market dynamics, and an algorithmic approach that leverages data-driven decision-making. The algorithmic approach involves extracting meaningful features from stock market data—typically time series data—and using this information to decide whether to buy or sell a stock. Advancements in Machine Learning and Deep Learning have significantly expanded the applicability of learning algorithms in stock market price prediction and classification, aiding traders in making informed decisions. Several studies have focused on predicting NIFTY 50 stock prices using time series models. Fathali et al. (2022) \cite{fathali2022stock} compared the performance of Recurrent Neural Networks (RNN), Long Short-Term Memory (LSTM), and Convolutional Neural Networks (CNN) for this task. Their findings indicated that an LSTM model utilizing High/Low/Open/Close (HLOC) features and a ReLU activation function achieved the best results, reporting a Root Mean Squared Error (RMSE) of 0.002 and an R-squared ($R^2$) value of 0.537 on the testing dataset. Other research has explored hybrid approaches. Manjunath et al. (2023) \cite{manjunath2023analysis} proposed a Principal Component Analysis (PCA) combined with various ML models, including Artificial Neural Networks (ANN), Support Vector Machines (SVM) with a Radial Basis Function (RBF) kernel, Naive Bayes, and Random Forest, to predict NIFTY 50 trends. Their hybrid PCA-ANN model achieved a remarkable accuracy and F1-score of 0.9984, while PCA combined with SVM (RBF) and Random Forest yielded an Area Under the Curve (AUC) of 1. Traditional neural network architectures have also been investigated. Kumar et al. (2019) \cite{kumar2019stock} employed a Multi-Layer Perceptron (MLP) with backpropagation to forecast the next day's Open (O), High (H), Low (L), and Close (C) data for the NIFTY 50, reporting an average accuracy of 99.21\% with an RMSE of 0.0079. Beyond standard neural networks, more sophisticated techniques like the Adaptive Neuro-Fuzzy Inference System (ANFIS) have been applied. Shankar et al. (2020) \cite{shankar2020systematic} demonstrated the effectiveness of ANFIS in modeling risk-return patterns and generating accurate predictions for selected NIFTY 50 blue-chip companies based on adjusted close price, volume, and net profit. In contrast, Singh et al. (2022) \cite{singh2022machine} evaluated a broader range of eight supervised learning models, including ANN, SVM, Stochastic Gradient Descent (SGD), Linear Regression (LR), AdaBoost, Random Forest (RF), k-Nearest Neighbors (kNN), and Decision Trees (DT), on NIFTY 50 data. Their analysis concluded that Linear Regression and ANN exhibited promising results, while Decision Trees showed signs of overfitting and poor generalization. Volatility forecasting has also been a subject of study. Mahajan et al. (2022) \cite{mahajan2022modeling} compared symmetric (GARCH) and asymmetric (EGARCH, TARCH) Generalized Autoregressive Conditional Heteroskedasticity models with an LSTM model for predicting NIFTY 50 volatility using the India VIX. Their findings indicated that the EGARCH model achieved the best forecasting accuracy, with a Mean Absolute Percentage Error (MAPE) of 16.99\% and an RMSE of 5.05, while the LSTM model underperformed across all metrics. Further exploration of ANN architectures includes the work by Lamba et al. (2021) \cite{lamba2021comparative}, who compared Feedforward Neural Networks (FFNN), Generalized Regression Neural Networks (GRNN), Radial Basis Neural Networks (RBNN), and Enhanced Radial Basis Neural Networks (ERBNN) for predicting NIFTY 50 closing prices. The ERBNN model achieved the highest accuracy of 99.31\%, followed by RBNN at 98.17\%. Studies comparing traditional ML models have also been conducted. Eswar et al. (2023) \cite{eswar2023prediction} compared SVM and Random Forest for NIFTY 50 price prediction using historical data, with Random Forest demonstrating superior performance, achieving an RMSE of 9.75 and an $R^2$ of 99.2\%. Similarly, Sen et al. (2021) \cite{sen2021stock} compared eight ML models and four LSTM variants, identifying a univariate LSTM model utilizing one week of prior data as the most accurate, with an RMSE/mean ratio of 0.0311. Deep learning models, particularly LSTMs, have gained significant attention. A study by \cite{9754148} applied an LSTM-based deep learning model to predict the prices of 10 NIFTY 50 stocks, achieving a peak accuracy of 83.88\% for State Bank of India (SBI) and an average accuracy of approximately 83\% across all analyzed stocks. Furthermore, Jafar et al. (2023) \cite{jafar2023forecasting} proposed a Backward Elimination-enhanced LSTM (BE-LSTM) model for NIFTY 50 price prediction, reporting superior performance compared to a standard LSTM with 95\% accuracy, 3.54\% MAPE, and an RMSE of 619.35 after 50 training epochs. Traditional statistical models have also been revisited in conjunction with ML. Patel et al. (2015) \cite{patel2015predicting} compared ANN, SVM, Random Forest, and Naive Bayes models for predicting NIFTY 50 and other indices using trend deterministic data. Their analysis revealed that the Naive Bayes (Multivariate Bernoulli) model achieved the best accuracy of 90.19\%, closely followed by Random Forest at 89.98\%. Jain et al. (2018) \cite{jain2018analysis} utilized an Artificial Neural Network with backpropagation to predict individual stock prices within the NIFTY 50 financial sector. The study reported moderate accuracy, with error percentages ranging from 6.13\% to 16.13\% across the different stocks analyzed.

In summary, the reviewed literature demonstrates a significant interest in applying various machine learning models to the NIFTY 50 stock market for prediction and analysis. Despite the extensive application of machine learning (ML) techniques in stock market prediction, most existing methods struggle to achieve robust and consistent returns over short time horizons. Deep learning models such as Long Short-Term Memory (LSTM) networks and Recurrent Neural Networks (RNNs), while adept at capturing sequential dependencies, face several challenges

\begin{itemize}
    \item \textbf{Overfitting:} 
    High model complexity requires vast amounts of data; intraday trading data can be noisy and insufficient to train deep networks effectively without overfitting.
    \item \textbf{Black-Box Interpretability:}
    Many high-performing ML models (e.g., deep neural networks) lack transparency in their decision-making, making it difficult for traders to trust or refine the outputs.
    \item \textbf{Inconsistent Short-Term Returns:}
    Volatile market movements lead to unpredictable performance when existing ML models are used for intraday (5-minute) timeframes. Gains from certain periods are often offset by losses in others.
\end{itemize}

 This difficulty arises due to inherent nonlinearity, abrupt fluctuations, and the influence of multiple interdependent factors driving stock market dynamics. In this paper, we focus on high-frequency data using 5-minute candlesticks for the 50 stocks listed in the NIFTY 50 index.
\subsection{Problem Definition}
\label{sec:problem_definition}

The central challenge in algorithmic trading lies in developing a system that not only demonstrates reliability and high performance but also offers interpretability, particularly when utilizing high-frequency data. Traditional approaches often struggle to consistently outperform naive strategies, such as a buy-and-hold approach. This necessitates the development of a robust trading system capable of exploiting short-term (5-minute) price movements.

To address these challenges, this study proposes the application of the \emph{Permutation Decision Tree} (PDT) algorithm by Harikrishnan et al. ~\cite{b2024permutationdecisiontrees}. PDT, a decision-tree framework driven by the Effort to Compress (\textit{ETC})~\cite{nagaraj2013complexity} metric, is hypothesized to effectively model patterns in high-frequency stock data. The core hypothesis is that, compared to black-box deep learning models, \emph{PDT can mitigate overfitting risks and offer enhanced interpretability}, thereby improving \emph{risk-adjusted returns} for short-term trading in the Indian stock market.

\section{Dataset Description and Feature Engineering}
For this study, we utilized intraday 5-minute candlestick data for the NIFTY 50 index's underlying constituent stocks (the list of stocks is available in the Appendix Section~\ref{sec:appendix}), a major Forex currency–currency pair (EURUSD), and a major Forex commodity–currency pair (XAUUSD). For NIFTY 50 index's underlying stocks, the data was sourced from Yahoo Finance\footnote{\url{https://finance.yahoo.com}} dated from 2024-10-18, 09:15:00 IST to 2025-01-14, 15:25:00 IST. For the Forex pairs, both XAUUSD and EURUSD, the data was sourced from Stooq\footnote{\url{https://stooq.com/db/h/}} dated from 2025-03-17 to 2025-08-11.
\paragraph{Dataset Overview (NIFTY 50 index's underlying stocks):}
Each stock has 4,426 data points, resulting in a total of 221,300 data points across all 50 stocks.
\paragraph{Dataset Overview (Forex pairs, XAUUSD and EURUSD):}
EURUSD has 29,711 data points and XAUUSD has 28,631 data points.

We have considered raw features and features derived from the raw features. They are as follows:
\begin{itemize}
    \item \textbf{Raw Features}
    \begin{enumerate}
        \item \textbf{Time}: The timestamp for the data point.
    \item \textbf{Open (O)}: The opening price of the stock during the 5-minute interval.
    \item \textbf{High (H)}: The highest price of the stock during the 5-minute interval.
    \item \textbf{Low (L)}: The lowest price of the stock during the 5-minute interval.
    \item \textbf{Close (C)}: The closing price of the stock during the 5-minute interval.
    \end{enumerate}
    \item \textbf{Derived Features}
    \begin{enumerate}
    \item \textbf{Swing High}: The most recent swing high price.
    \item \textbf{Swing Low}: The most recent swing low price.
    \begin{itemize}
        \item Swing highs and swing lows are critical concepts in technical analysis that help traders identify key turning points in the market. These points represent areas of psychological significance, often acting as resistance (for swing highs) or support (for swing lows).

\paragraph{Definitions:}
\begin{itemize}
    \item \textbf{Swing High:} A price point that is higher than the prices immediately preceding and following it. Swing highs indicate a local peak in the price chart.
    \item \textbf{Swing Low:} A price point that is lower than the prices immediately preceding and following it. Swing lows indicate a local trough in the price chart.
\end{itemize}

\paragraph{Importance of Swing Highs and Lows:}
\begin{itemize}
    \item Swing highs often act as \textbf{resistance levels}, where prices struggle to rise above.
    \item Swing lows often act as \textbf{support levels}, where prices struggle to fall below.
    \item These points are used to identify trends and reversals, helping traders make informed decisions.
\end{itemize}

Swing highs and lows are identified by comparing a price point with its neighbors:
\begin{itemize}
    \item A price point is a swing high if:
    \[
    P_i > P_{i-1} \quad \text{and} \quad P_i > P_{i+1}
    \]
    \item A price point is a swing low if:
    \[
    P_i < P_{i-1} \quad \text{and} \quad P_i < P_{i+1}
    \]
\end{itemize}

    \end{itemize}
    \item \textbf{Distance to Swing High}: The difference between the current closing price and the swing high price.
    \item \textbf{Distance to Swing Low}: The difference between the current closing price and the swing low price.
    \item \textbf{Order Block Indicator}: Binary value indicating the presence of an order block.
    \begin{itemize}
        \item An \textbf{order block} is a price zone where large institutional investors (such as mutual funds, hedge funds, or banks) have placed significant buy or sell orders. These zones often act as strong levels of \textbf{support} or \textbf{resistance} because of the high volume of trades executed by these bigger players.

\paragraph{Characteristics of Order Blocks:}
\begin{itemize}
    \item \textbf{Consolidation Zone:} Before a large price movement, the stock price often consolidates within a narrow range. This is where large investors are accumulating (buying) or distributing (selling) their positions.
    \item \textbf{Breakout:} After the accumulation or distribution is complete, the stock breaks out of the range:
    \begin{itemize}
        \item If the breakout is \textbf{upward}, it indicates buying activity by large investors.
        \item If the breakout is \textbf{downward}, it indicates selling activity by large investors.
    \end{itemize}
    \item \textbf{Future Price Action:} After a breakout, the order block acts as:
    \begin{itemize}
        \item A \textbf{support level} if the price retraces to the order block after an upward breakout.
        \item A \textbf{resistance level} if the price retraces to the order block after a downward breakout.
    \end{itemize}
    \paragraph{Support and Resistance:}
\begin{itemize}
    \item \textbf{Support:} A price level where buying pressure is strong enough to prevent the price from falling further. It acts as a “floor” for the price.
    \item \textbf{Resistance:} A price level where selling pressure is strong enough to prevent the price from rising further. It acts as a “ceiling” for the price.
\end{itemize}

\end{itemize}

\paragraph{How to Identify Order Blocks:}
Order blocks are identified by finding periods of low price volatility followed by high price volatility. In this study, the order block is computed using:
\begin{enumerate}
    \item Calculating the \textbf{rolling high} (maximum price) and \textbf{rolling low} (minimum price) over a fixed window (e.g., 5 candlesticks).
    \item Checking whether the range between the rolling high and low is smaller than a percentage (e.g., 0.2\%) of the current closing price.
\end{enumerate}

    \end{itemize}
    \item \textbf{20-Period Moving Average (MA\_20)}: The simple moving average of the closing prices over the last 20 intervals.
    \item \textbf{50-Period Moving Average (MA\_50)}: The simple moving average of the closing prices over the last 50 intervals.
    \item \textbf{Diff}: The simple difference between MA\_20 and MA\_50.
    \item \textbf{Future Close}: This is the closing price from a variable time steps into the future for each current row.
    \item \textbf{Relative Strength Index (RSI)}:
    Momentum oscillator measuring the speed and change of price movements.
    \begin{itemize}
        \item The \textbf{Relative Strength Index (RSI)} is a momentum indicator that measures the speed and magnitude of price changes to determine whether a stock is \textbf{overbought} or \textbf{oversold}. RSI values range from 0 to 100:
\begin{itemize}
    \item \textbf{Overbought ($RSI > 70$):} The stock price has risen too much and might fall soon.
    \item \textbf{Oversold ($RSI < 30$):} The stock price has fallen too much and might rise soon.
\end{itemize}

\paragraph{Formula for RSI:}
The RSI is calculated using the following formula:
\[
\text{RSI} = 100 - \left( \frac{100}{1 + RS} \right)
\]
where \( RS \) (Relative Strength) is defined as:
\[
RS = \frac{\text{Average Gain over } n \text{ periods}}{\text{Average Loss over } n \text{ periods}}
\]

\paragraph{Steps to Calculate RSI:}
\begin{enumerate}
    \item Compute the price changes for each interval (\( \Delta P = P_{\text{current}} - P_{\text{previous}} \)).
    \item Separate the gains (\( \Delta P > 0 \)) and losses (\( \Delta P < 0 \)).
    \item Calculate the average gain and average loss over \( n \) periods (usually \( n = 14 \)).
    \item Compute \( RS = \frac{\text{Average Gain}}{\text{Average Loss}} \).
    \item Use the \( RS \) to calculate RSI.
\end{enumerate}

    \end{itemize}
    \end{enumerate}

    \item \textbf{Label}: Binary value indicating whether the stock price is expected to increase (1) or decrease (0) over the next 50 intervals (methodology of calculating the target is mentioned in Section~\ref{sec:labelmethod}.).
\end{itemize}

Table~\ref{tab:raw_data} and Table~\ref{tab:features} provides a sample dataset for Reliance Industries, highlighting the features used in our experiments.
The raw data was obtained from Yahoo Finance, containing the essential price information for each 5-minute interval. Table~\ref{tab:raw_data} provides a sample of this raw data for Reliance Industries.
From the raw data, we calculated several \emph{derived features} which includes technical indicators and features to enhance the predictive power of the model. Table~\ref{tab:features} provides a sample of the derived features.

\begin{table}[!h]
\centering
\caption{Raw Data for Reliance Industries (5-minute Candlestick Data)}
\label{tab:raw_data}
\begin{tabular}{cccccc}
\toprule
\textbf{Time} & \textbf{Open} & \textbf{High} & \textbf{Low} & \textbf{Close} \\
\midrule
2024-10-18 13:25 & 1364.65 & 1364.90 & 1363.07 & 1364.53 \\
2024-10-18 13:30 & 1364.72 & 1366.80 & 1363.70 & 1366.00 \\
2024-10-18 13:35 & 1366.20 & 1366.50 & 1365.28 & 1365.82 \\
2024-10-18 13:40 & 1365.75 & 1365.75 & 1364.53 & 1364.78 \\
2024-10-18 13:45 & 1364.78 & 1366.55 & 1364.53 & 1365.25 \\
2024-10-18 13:50 & 1365.25 & 1365.25 & 1362.50 & 1363.95 \\
2024-10-18 13:55 & 1363.75 & 1364.80 & 1363.50 & 1364.65 \\
2024-10-18 14:00 & 1364.65 & 1365.00 & 1363.95 & 1364.35 \\
2024-10-18 14:05 & 1364.20 & 1365.45 & 1363.55 & 1365.25 \\
2024-10-18 14:10 & 1365.03 & 1365.90 & 1364.57 & 1365.50 \\
\bottomrule
\end{tabular}
\end{table}

\begin{table}[!h]
\centering
\caption{Derived Features for Reliance Industries}
\label{tab:features}
\resizebox{\textwidth}{!}{%
\begin{tabular}{cccccccccc}
\toprule
\textbf{Time} & \textbf{SH} & \textbf{SL} & \textbf{Dist\_SH} & \textbf{Dist\_SL} & \textbf{OB} & \textbf{MA\_20} & \textbf{MA\_50} & \textbf{diff} & \textbf{RSI} \\
\midrule
2024-10-18 13:25 & 1365.50 & 1362.50 & -0.97 & 2.03 & 0 & 1362.73 & 1356.42 & 6.31 & 57.27 \\
2024-10-18 13:30 & 1366.80 & 1362.50 & -0.80 & 3.50 & 0 & 1363.06 & 1356.80 & 6.26 & 56.79 \\
2024-10-18 13:35 & 1366.80 & 1362.50 & -0.98 & 3.32 & 0 & 1363.38 & 1357.28 & 6.10 & 60.97 \\
2024-10-18 13:40 & 1366.80 & 1362.50 & -2.03 & 2.28 & 0 & 1363.69 & 1357.74 & 5.95 & 60.22 \\
2024-10-18 13:45 & 1366.80 & 1362.50 & -1.55 & 2.75 & 0 & 1363.94 & 1358.11 & 5.82 & 55.82 \\
2024-10-18 13:50 & 1366.80 & 1362.50 & -2.85 & 1.45 & 0 & 1364.18 & 1358.48 & 5.70 & 57.34 \\
2024-10-18 13:55 & 1366.80 & 1362.50 & -2.15 & 2.15 & 0 & 1364.36 & 1358.83 & 5.54 & 52.06 \\
2024-10-18 14:00 & 1366.80 & 1362.50 & -2.45 & 1.85 & 0 & 1364.58 & 1359.14 & 5.44 & 54.49 \\
2024-10-18 14:05 & 1365.45 & 1362.50 & -0.20 & 2.75 & 0 & 1364.69 & 1359.54 & 5.15 & 53.24 \\
2024-10-18 14:10 & 1365.90 & 1362.50 & -0.40 & 3.00 & 0 & 1364.62 & 1359.91 & 4.71 & 56.46 \\
\bottomrule
\end{tabular}
}
\end{table}

\paragraph{Abbreviations Used:}
The following abbreviations are used in the dataset to denote various features:

\begin{table}[!h]
\centering
\caption{Abbreviations and Their Descriptions}
\label{tab:abbreviations}
\begin{tabular}{lp{10cm}}
\toprule
\textbf{Abbreviation} & \textbf{Description} \\
\midrule
\textbf{SL}           & Swing Low: The most recent significant low price. \\
\textbf{SH}           & Swing High: The most recent significant high price. \\
\textbf{Dist\_SL}     & Distance to Swing Low: Difference between current close price and Swing Low. \\
\textbf{Dist\_SH}     & Distance to Swing High: Difference between current close price and Swing High. \\
\textbf{OB}           & Order Block Indicator: Binary value indicating presence of order block. \\
\textbf{MA\_20}       & 20-Period Moving Average. \\
\textbf{MA\_50}       & 50-Period Moving Average. \\
\textbf{RSI}          & Relative Strength Index: Momentum oscillator for price movement. \\
\bottomrule
\end{tabular}
\end{table}

\subsection{Preprocessing}

Prior to feature engineering and model training, the raw data underwent a series of preprocessing steps. These procedures ensured data cleanliness, consistency, and alignment across all stocks.

\begin{enumerate}

\item Feature Filtering:
From the raw data, only the necessary features were retained for further analysis. These include only Close prices
\item Handling Missing Data: In our dataset, we have never encountered missing data, thus ensuring consistency across analysis.
\item Timeframe Consistency:
To ensure uniformity, all stock data points were aligned to the same 5-minute intervals.
\item After feature calculations, we removed the rows with missing data.

\end{enumerate}

\section{Proposed Method}~\label{sec:ProposedMethod}
\subsection{Problem Formulation}

\subsubsection{Data and Notation}
Let $(O_t,H_t,L_t,C_t)$ denote the 5-minute Open, High, Low and Close sequence for a given instrument. We construct a feature vector at each index $t$ from price-derived signals:
\[
\mathbf{x}_t=\big[\text{DistSH}_t,\ \text{DistSL}_t,\ \text{OB}_t,\ \text{MA20}_t,\ \text{MA50}_t,\ \text{RSI}_t,\ \text{Diff}_t
\big],
\]
where (i) $\text{DistSH}_t = C_t-\text{SH}_t$, $\text{DistSL}_t = C_t-\text{SL}_t$; (ii) $\text{OB}_t\in\{0,1\}$ indicates order-block proximity; (iii) $\text{MA20}_t,\text{MA50}_t$ are simple moving averages; (iv) $\text{RSI}_t$ is the Relative Strength Index; and (v) $\text{Diff}_t=\text{MA20}_t-\text{MA50}_t$. All rolling statistics use only past data up to $t$ (no leakage).

\subsection{Training and Testing Splits}\label{sec:labelmethod}
We perform a \emph{temopral} split of the time series into non overlapping blocks: the first block for training and the subsequent block for testing. Let $\mathcal{T}_{\text{train}}$ and $\mathcal{T}_{\text{test}}$ denote the index sets corresponding to the training and testing periods, respectively (We define the exact splitting in the upcoming subsections).

\begin{itemize}
    \item \textbf{Training labels}
    For each training index $t$, we construct the binary label
\[
y_t=
\begin{cases}
1, & \text{if } C_{t+h}>C_t,\\
0, & \text{otherwise},
\end{cases}
\]

where $C_t$ is the close at time $t$ and $h$ is the forecast horizon in bars (in our case, $h=50$, this value is a hyper parameter). These labels are denoted $\{y_t\}$ and are used to fit the classifier $f$. Since we are accessing the $Close$ prices of the future to calculate the label, there is a chance of data leakage, i.e. while we calculate the labels for the train data, we might access the data which is beyond the training segment (test segment). This issue has been addressed in the upcoming subsections.
\item \textbf{Test predictions} For testing, at time $t$, we compute the features from the closing prices available before or at $t$ (for our case, to generate the features, we roll back to at most 50 steps behind to calculate features, like the $MA\_50$, the parameters for the moving averages can be controlled) to generate the predicted labels:
\[
\hat{y}_t = f(\mathbf{x}_t),\qquad t\in \mathcal{T}_{\text{test}}.
\]
\item \textbf{Leakage control at the split}: Since training labels use $C_{t+h}$, the last $h$ training points would reference the \textit{Close} prices that falls inside the test window. To eliminate any cross-boundary look-ahead, we \emph{discard the first $h$ samples of the test block} before evaluation:
\[
\mathcal{T}_{\text{test}} \;=\; \{\, t\in \mathcal{T}_{\text{test}} \mid t \ge t_{\text{test,start}} + h \,\}.
\]
All rolling statistics and features are computed with data $\le t$ only, and the model selection/evaluation uses the leakage-free set $\mathcal{T}_{\text{test, eval}}$.
For train-test splitting, we are first splitting the time series data in a 80:20 ratio for training and testing respectively. Post splitting, we are discarding the first $h$ (in our case $h=50$) from test data and evaluating the strategy. For all stocks under NIFTY 50, we are directly applying the splitting algorithm to the complete data corresponding to each stocks. But for the Forex pairs, we are breaking down the dataset (maintaining the time order) into chunks of size 3225. So there will be 3225 rows available in each chunk, and this train/test is experimented on each chunk and results are available in the appendix. The reason we are doing this kind of chunking is due to the large time that PDT takes to learn for bigger training data.

\end{itemize}
\subsubsection{Objective}~\label{sec:objective}
Learn a classifier $f:\mathbb{R}^d\!\to\!\{0,1\}$ where $d$ is the dimension of the feature vector (7 in our case). We instantiate $f$ as a Permutation Decision Tree (PDT) using the Effort-To-Compress (\textit{ETC}) impurity on the \emph{time-ordered} label sequence at each node. In addition, we use a rule-based execution policy that maps $f$’s predictions into trades as follows:
\begin{itemize}
    \item \textbf{Entry}:
If $\hat{y}_t=f(\mathbf{x}_t)=1$, initiate a long position at the close $C_t$ \,(buy the maximum number of shares with available cash); otherwise, remain flat \,(hold no position and place no orders).
    \item \textbf{Exit:}
    We decide to sell all our positions if at least one of the following conditions are satisfied:
    \begin{itemize}
    \item \textbf{Trailing stop-loss}: Let $t_e$ be the entry time and running time is defined as follows:
\[
M_t=\max_{u\in[t_e,t]} C_u,
\]
where $t$ is the current time and $t_e$ is the time at which we had purchased the stocks (Entry condition).
For a trail parameter $\lambda\in(0,1)$, the stop level is ($S_t$):
\[
S_t=(1-\lambda)\,M_t,
\]
In our use case, we have used the value of $\lambda=0.005$. This value can be changed in order to control the behavior of the strategy.

Exit occurs at the first $t$ such that $C_t\le S_t$ (filled at $S_t$).
\item \textbf{Model Prediction}: We also exit if the prediction flips, $\hat{y}_t=0$.
\end{itemize}
\item \textbf{Objective function}: We select model and policy parameters (tree depth, node size, $\lambda$, $h$) to \textit{maximize profits}.

\end{itemize}

\subsubsection{\textit{ETC} Impurity and Split Gain.}
For a node containing the ordered labels $S = (y_{t_1},\dots,y_{t_m})$, let $ETC(S)$ be its \textit{ETC} value. For a candidate split $(j,\tau)$ on feature $x^{(j)}$ at threshold $\tau$, let $S_L,S_R$ be the left/right label subsequences (preserving the original temporal order). The \textit{ETC gain} is
\[
\text{Gain}(j,\tau)= ETC(S) - \frac{|S_L|}{|S|}ETC(S_L) - \frac{|S_R|}{|S|}ETC(S_R).
\]
We choose $
(j,\tau) = \arg\max_{\substack{(j,\;\tau)}} \; \text{Gain}(j,\tau).
$

\subsubsection{Tree Growth and Stopping}
Top–down recursion of the Permutation Decision Tree construction continues until any of the following holds: (i) $ETC(S)=0$ (homogeneous), (ii) node size $< n_{\min}$, (iii) depth $= D_{\max}$.

\subsubsection{Method to calculate \textit{ETC}.}
At any node of the tree, we take the \emph{time-ordered} sequence of labels that reach that node, e.g.
\[
S=(y_{t_1},y_{t_2},\ldots,y_{t_m}),\qquad y_{t_i}\in\{0,1\}.
\]
We compute the Effort-To-Compress (\textit{ETC}) of this sequence using the Non-Sequential Recursive Pair Substitution (NSRPS)~\cite{ebeling1980grammars} procedure: repeatedly (a) count all consecutive pairs, (b) among the most frequent pairs pick the one that \emph{appears first} when scanning left-to-right, and (c) replace all \emph{non-overlapping} occurrences of that pair with a new symbol. The \textit{ETC} value is the number of such iterations needed to make the sequence homogeneous. This is the impurity at that node; when we try a split, we compare the node’s \textit{ETC} to the (size-weighted) \textit{ETC} of the left/right label subsequences to get the \textit{ETC} Gain.

\textbf{Worked example (8 consecutive 5-minute labels).}
Suppose the label sequence for 8 successive 5-minute bars is
\[
S=[0,0,1,0,1,1,0,1].
\]
Applying NSRPS step-by-step (ties broken by the first \emph{occurring} max-frequency pair; replacements are non-overlapping):

\begin{enumerate}
\item \emph{Iter 1.} Pair counts: $(0,1)\times 3,\ (1,0)\times 2,\ (0,0)\times 1,\ (1,1)\times 1$.\\
First max pair encountered is $(0,1)$. Replace all non-overlapping $(0,1)\Rightarrow 2$:
\[
[0,0,1,0,1,1,0,1]\ \Rightarrow\ [0,2,2,1,2].
\]

\item \emph{Iter 2.} Pair counts: all frequency $1$; first encountered is $(0,2)$. Replace $(0,2)\Rightarrow 3$:
\[
[0,2,2,1,2]\ \Rightarrow\ [3,2,1,2].
\]

\item \emph{Iter 3.} First encountered max pair is $(3,2)$. Replace $(3,2)\Rightarrow 4$:
\[
[3,2,1,2]\ \Rightarrow\ [4,1,2].
\]

\item \emph{Iter 4.} First encountered max pair is $(4,1)$. Replace $(4,1)\Rightarrow 5$:
\[
[4,1,2]\ \Rightarrow\ [5,2].
\]

\item \emph{Iter 5.} Only pair is $(5,2)$. Replace $(5,2)\Rightarrow 6$:
\[
[5,2]\ \Rightarrow\ [6].
\]
\end{enumerate}
The sequence is now homogeneous, so $\mathrm{\textit{ETC}}(S)=5$. This is exactly the computation we perform on the \emph{entire} time-ordered label set at the root (and on each node’s label subsequence thereafter) when evaluating splits via \textit{ETC} Gain.




\subsection{Permutation Decision Tree}
A Permutation Decision Tree (PDT) is a decision-tree-based algorithm that measures the 
\emph{uniformity} of target labels using a novel metric called \textit{Effort To Compress (ETC)}~\cite{balasubramanian2016aging,nagaraj2013complexity} instead of traditional criteria (e.g., Gini impurity). By focusing on \textit{ETC}, a PDT locates 
splits that make the target labels (0s and 1s) in each node more “compressible,” 
revealing threshold-based patterns that conventional splitting criteria may not detect.

\vspace{2mm}
\noindent
\textbf{Key Concepts in PDT:}
\begin{itemize}
    \item \textbf{Effort To Compress (\textit{ETC}):} 
    A measure of how many pair-replacement iterations are needed to compress a label sequence 
    into a single repeated symbol. The fewer the iterations, the more uniform (or “pure”) the 
    sequence is. See Algorithm~\ref{alg:calculate_etc}. Unlike Shannon entropy and Gini impurity, \textit{ETC} does not assume independent and identical distribution (i.i.d). This makes \textit{ETC} as an impurity measure ideal for datasets having temporal dependencies across data instances~\cite{b2024permutationdecisiontrees}.
    \item \textbf{Split Criterion (\textit{ETC Gain}):} 
    When choosing a threshold on a given feature, we compute the drop in \textit{ETC} (weighted by the 
    sizes of the child nodes) from the parent node to its left/right children. 
    A higher reduction (i.e., higher \textit{ETC Gain}) implies a more effective split. 
    See Algorithm~\ref{alg:etc_gain}. 
\end{itemize}

Thus, PDT explicitly seeks splits that yield subsets of target labels that can be “compressed” 
more easily. This can reveal threshold-based patterns that might not be captured by 
conventional impurity metrics.

\begin{algorithm}[!h]
\caption{\textsc{CalculateETC} -- Computes the \textit{ETC} Value of a Sequence.}
\label{alg:calculate_etc}
\begin{algorithmic}[1]
    \REQUIRE A sequence \(\mathit{seq}\) (list or string of digits)
    \ENSURE \(\mathrm{ETC}(\mathit{seq})\): number of pair-replacement iterations until uniformity
    \STATE \(\mathit{itrs} \gets 0\)
    \STATE Convert all elements in \(\mathit{seq}\) to integers
    \STATE \(\mathit{max\_symbol} \gets \max(\mathit{seq})\)
    \WHILE{\(|\text{set}(\mathit{seq})| > 1\)}
        \STATE Count all consecutive pairs \((\mathit{seq}[i], \mathit{seq}[i+1])\) for \(i = 0, 1, \dots, |\mathit{seq}| - 2\)
        \STATE \(\mathit{highest\_freq\_pair} \gets \text{pair with max frequency}\)
        \STATE \(\mathit{max\_symbol} \gets \mathit{max\_symbol} + 1\)
        \STATE \(\mathit{new\_symbol} \gets \mathit{max\_symbol}\)
        \STATE \(\mathit{temp\_seq} \gets [\,]\)
        \STATE \(i \gets 0\)
        \WHILE{\(i < |\mathit{seq}|\)}
            \IF{\(i < |\mathit{seq}| - 1\) \AND \((\mathit{seq}[i], \mathit{seq}[i+1]) = \mathit{highest\_freq\_pair}\)}
                \STATE Append \(\mathit{new\_symbol}\) to \(\mathit{temp\_seq}\)
                \STATE \(i \gets i + 2\)
            \ELSE
                \STATE Append \(\mathit{seq}[i]\) to \(\mathit{temp\_seq}\)
                \STATE \(i \gets i + 1\)
            \ENDIF
        \ENDWHILE
        \STATE \(\mathit{seq} \gets \mathit{temp\_seq}\)
        \STATE \(\mathit{itrs} \gets \mathit{itrs} + 1\)
    \ENDWHILE
    \RETURN \(\mathit{itrs}\)
\end{algorithmic}
\end{algorithm}

\begin{algorithm}[!h]
\caption{\textsc{\textit{ETC\_Gain}} -- Finds the Best Threshold Split for a Single Feature.}
\label{alg:etc_gain}
\begin{algorithmic}[1]
    \REQUIRE 
      \(\mathit{data}\): 2D array of shape \((n, d)\) \\
      \(\mathit{labels}\): 1D array of length \(n\) \\
      \(\mathit{feature\_index}\): Index of feature to split on
    \ENSURE \(\bigl(\mathit{best\_gain}, \mathit{best\_threshold}\bigr)\)

    \STATE Convert \(\mathit{labels}\) to integer array
    \STATE \(\mathit{totalETC} \gets \textsc{CalculateETC}(\mathit{labels})\)
    \STATE \(\mathit{feature\_values} \gets \mathit{data}[:, \mathit{feature\_index}]\)
    \STATE \(\mathit{unique\_vals} \gets \text{sorted unique values of } \mathit{feature\_values}\)

    \IF{\(|\mathit{unique\_vals}| = 1\)}
        \RETURN \(({-}\infty, \text{None})\)
    \ENDIF

    \STATE \(\mathit{thresholds} \gets \text{midpoints between consecutive elements of } \mathit{unique\_vals}\)

    \STATE \(\mathit{best\_gain} \gets {-}\infty\)
    \STATE \(\mathit{best\_threshold} \gets \text{None}\)

    \FOR{each threshold \(t \in \mathit{thresholds}\)}
        \STATE \(\mathit{left\_mask} \gets \mathit{feature\_values} \le t\)
        \STATE \(\mathit{right\_mask} \gets \text{not } \mathit{left\_mask}\)
        \STATE \(\mathit{left\_labels} \gets \mathit{labels}[\mathit{left\_mask}]\)
        \STATE \(\mathit{right\_labels} \gets \mathit{labels}[\mathit{right\_mask}]\)

        \STATE \(\mathit{etcL} \gets \textsc{CalculateETC}(\mathit{left\_labels}) \text{ if } |\mathit{left\_labels}| > 0 \text{ else } 0\)
        \STATE \(\mathit{etcR} \gets \textsc{CalculateETC}(\mathit{right\_labels}) \text{ if } |\mathit{right\_labels}| > 0 \text{ else } 0\)

        \STATE \(\mathit{wL} \gets \frac{|\mathit{left\_labels}|}{|\mathit{labels}|}\)
        \STATE \(\mathit{wR} \gets \frac{|\mathit{right\_labels}|}{|\mathit{labels}|}\)

        \STATE \(\mathit{weightedETC} \gets \mathit{wL} \times \mathit{etcL} + \mathit{wR} \times \mathit{etcR}\)
        \STATE \(\mathit{gain} \gets \mathit{totalETC} - \mathit{weightedETC}\)

        \IF{\(\mathit{gain} > \mathit{best\_gain}\)}
            \STATE \(\mathit{best\_gain} \gets \mathit{gain}\)
            \STATE \(\mathit{best\_threshold} \gets t\)
        \ENDIF
    \ENDFOR

    \RETURN \((\mathit{best\_gain}, \mathit{best\_threshold})\)
\end{algorithmic}
\end{algorithm}

Once we determine the “best threshold” for a feature, the next step is to see which feature 
\emph{overall} (among many) yields the highest \textit{ETC Gain}. This is handled by \textsc{Find\_Best\_Feature}, 
shown in Algorithm~\ref{alg:find_best_feature}.
\begin{algorithm}[!h]
\caption{\textsc{Find\_Best\_Feature} -- Identifies the Feature and Threshold with Maximum \textit{ETC Gain}.}
\label{alg:find_best_feature}
\begin{algorithmic}[1]
    \REQUIRE 
       \(\mathit{data}\): 2D array of shape \((n, d)\) \\
       \(\mathit{labels}\): 1D array of length \(n\)
    \ENSURE \(\bigl(\mathit{bestFeature}, \mathit{bestThreshold}\bigr)\)

    \STATE Convert \(\mathit{data}, \mathit{labels}\) to arrays (if not already)
    \STATE \(\mathit{numFeatures} \gets \text{number of columns in } \mathit{data}\)
    \STATE \(\mathit{bestFeature} \gets \text{None}\)
    \STATE \(\mathit{bestThreshold} \gets \text{None}\)
    \STATE \(\mathit{bestGain} \gets -\infty\)

    \FOR{\(f\_idx = 0 \text{ to } \mathit{numFeatures} - 1\)}
        \STATE \((\mathit{gain}, \mathit{threshold}) \gets \textsc{ETC\_Gain}(\mathit{data}, \mathit{labels}, f\_idx)\)
        \IF{\(\mathit{gain} > \mathit{bestGain}\)}
            \STATE \(\mathit{bestGain} \gets \mathit{gain}\)
            \STATE \(\mathit{bestFeature} \gets f\_idx\)
            \STATE \(\mathit{bestThreshold} \gets \mathit{threshold}\)
        \ENDIF
    \ENDFOR

    \RETURN \((\mathit{bestFeature}, \mathit{bestThreshold})\)
\end{algorithmic}
\end{algorithm}

Finally, the overall PDT is constructed \emph{recursively} by \textsc{Build\_PDT} (Algorithm~\ref{alg:build_pdt}). We keep splitting until labels become uniform or the maximum depth of the tree (stopping criteria) is met.
\newpage
\subsection{Framework Using PDT for Stock Classification}
\label{sec:pdt_framework}

\begin{figure}[ht]
    \centering
    \includegraphics[height=0.5\linewidth]{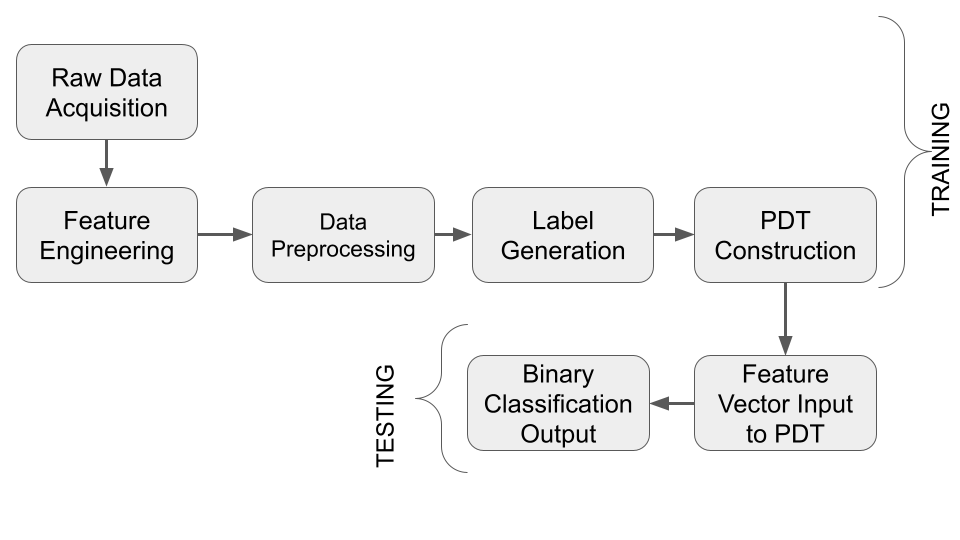}
    \caption{Architecture Diagram of the Proposed PDT-Based Stock Classification Pipeline.
             Data is collected and preprocessed, then used to build a Permutation 
             Decision Tree, which predicts stock movements for future intervals.}
    \label{fig:pdt_pipeline}
\end{figure}

To classify short-term stock movements (up or down), we integrate PDT 
within the following pipeline (illustrated in Figure~\ref{fig:pdt_pipeline}):

\begin{enumerate}

    \item \textbf{Building the PDT Model:}
          \begin{itemize}
              \item Calculate \textit{ETC} of the entire (parent) label set.
              \item Search thresholds for each feature to find the best \textit{ETC Gain} split 
                    (Algorithms~\ref{alg:etc_gain}~and~\ref{alg:find_best_feature}).
              \item Recursively grow left/right child nodes until purity or maximum depth 
                    (Algorithm~\ref{alg:build_pdt}).
          \end{itemize}
    \item \textbf{Prediction:}
          For a new feature vector (e.g., the next 5-minute candle and indicators):
          \begin{itemize}
              \item Traverse the PDT from the root, comparing feature values to thresholds.
              \item Stop at a leaf, which outputs \(\hat{y} \in \{0,1\}\).
              \item If \(\hat{y} = 1\), we anticipate a price increase; if \(\hat{y} = 0\), 
                    we expect no increase (or a drop).
          \end{itemize}

\end{enumerate}

\subsection{Evaluation Metrics}

To measure the performance and effectiveness of the proposed trading strategy, 
we use the following evaluation metrics:

\subsubsection{Growth (\%)}
The growth measures the overall percentage gain or loss achieved during the 
trading period. It is calculated as:
\[
\text{Growth \%} 
= \frac{V_{\text{final}} - V_{\text{initial}}}{V_{\text{initial}}} 
  \times 100,
\]
where \( V_{\text{final}} \) is the final portfolio value at the end of the 
trading period and \( V_{\text{initial}} \) (in our case \( V_{\text{initial}} \)=10,000 units of currency, INR for Indian stocks and USD for Forex pairs) is the initial portfolio value.

\subsubsection{Maximum Drawdown (\%)}
The maximum drawdown quantifies the largest percentage decline in 
the portfolio from its peak to its lowest point during the trading period. 
It is calculated as:
\[
\text{Maximum Drawdown} 
= \max \left( 
   \frac{V_{\text{peak}} - V_t}{V_{\text{peak}}} 
   \times 100 \right),
\]
where \( V_{\text{peak}} \) is the highest portfolio value observed at or before time $t$ and \( V_t \) is the portfolio value at time \( t \).

These metrics collectively provide a comprehensive assessment of both 
\emph{profitability} and \emph{risk management} for any trading approach 
driven by PDT-based classification.

\section{Experimental Setup}
This section explains the procedure used to simulate trading strategies based on the predictions generated by the PDT model.

\begin{lrbox}{\algobox}
\begin{minipage}{0.97\textwidth}
\scriptsize
\begin{algorithm}[H]
\caption{\textsc{Build\_PDT} -- Recursively Constructs a Permutation Decision Tree.}
\label{alg:build_pdt}
\begin{algorithmic}[1]
    \REQUIRE 
      \(\mathit{data}\): Feature matrix, \(\mathit{labels}\): target labels \\
      \(\mathit{depth}\): current depth, \(\mathit{maxDepth}\): maximum allowed depth \\
      \(\mathit{pbar}\): optional progress tracker
    \ENSURE A PDT node (dictionary) or a leaf label

    \STATE Convert \(\mathit{data}\) and \(\mathit{labels}\) to NumPy arrays
    \IF{\(|\mathit{data}| = 0\)}
        \IF{\(\mathit{pbar} \neq \text{None}\)} \STATE \(\mathit{pbar.update(1)}\) \ENDIF
        \RETURN \text{None}
    \ENDIF

    \IF{all labels in \(\mathit{labels}\) are identical}
        \IF{\(\mathit{pbar} \neq \text{None}\)} \STATE \(\mathit{pbar.update(1)}\) \ENDIF
        \RETURN that identical label
    \ENDIF

    \IF{\(\mathit{depth} \ge \mathit{maxDepth}\)}
        \IF{\(\mathit{pbar} \neq \text{None}\)} \STATE \(\mathit{pbar.update(1)}\) \ENDIF
        \STATE Return majority label in \(\mathit{labels}\)
    \ENDIF

    \STATE \((\mathit{bestFeature}, \mathit{bestThreshold}) \gets \textsc{Find\_Best\_Feature}(\mathit{data}, \mathit{labels})\)

    \IF{\(\mathit{bestThreshold} = \text{None}\)}
        \IF{\(\mathit{pbar} \neq \text{None}\)} \STATE \(\mathit{pbar.update(1)}\) \ENDIF
        \STATE Return majority label in \(\mathit{labels}\)
    \ENDIF

    \STATE \(\mathit{tree} \gets \{\texttt{feature\_index}: \mathit{bestFeature}, \texttt{threshold}: \mathit{bestThreshold}, \texttt{children}: \{\}\}\)

    \STATE \(\mathit{leftMask} \gets \mathit{data}[:, \mathit{bestFeature}] \le \mathit{bestThreshold}\)
    \STATE \(\mathit{rightMask} \gets \text{not } \mathit{leftMask}\)
    \STATE \(\mathit{leftData}, \mathit{leftLabels} \gets \mathit{data}[leftMask], \mathit{labels}[leftMask]\)
    \STATE \(\mathit{rightData}, \mathit{rightLabels} \gets \mathit{data}[rightMask], \mathit{labels}[rightMask]\)

    \STATE \(\mathit{tree.children.left} \gets \textsc{Build\_PDT}(\mathit{leftData}, \mathit{leftLabels}, \mathit{depth}+1, \mathit{maxDepth}, \mathit{pbar})\)
    \STATE \(\mathit{tree.children.right} \gets \textsc{Build\_PDT}(\mathit{rightData}, \mathit{rightLabels}, \mathit{depth}+1, \mathit{maxDepth}, \mathit{pbar})\)

    \IF{\(\mathit{pbar} \neq \text{None}\)}
        \STATE \(\mathit{pbar.update(1)}\)
    \ENDIF

    \RETURN \(\mathit{tree}\)
\end{algorithmic}
\end{algorithm}
\end{minipage}
\end{lrbox}
\noindent\resizebox{\textwidth}{!}{\usebox{\algobox}}

\subsection{Key Components of the Simulation}
The simulation involves the following steps:
\begin{enumerate}
    \item \textbf{Initializing Portfolio:} Start with an initial balance (in our setup, we have started with 10,000INR for indian stocks and 10,000USD for forex pairs) and no positions.
    \item \textbf{Iterative Trading:} For each time interval:
    \begin{itemize}
        \item Evaluate the model's prediction for the current interval.
        \item Execute buy, sell, or hold decisions based on the trading rules defined in Section~\ref{sec:objective}.
    \end{itemize}
    \item \textbf{Risk Management:} Use a trailing stop-loss mechanism (defined in Section~\ref{sec:objective}) to limit losses and lock in profits.
    \item \textbf{Portfolio Value Update:} Update the portfolio balance and positions after each trade.
\end{enumerate}
\subsection{Structure of Results DataFrame}
The \textbf{results\_df} is a key input to the simulation algorithm. It contains the following columns:
\begin{itemize}
    \item \textbf{Datetime:} The timestamp of the corresponding stock price data.
    \item \textbf{Actual:} The actual label indicating whether the price increased (\texttt{1}) or decreased (\texttt{0}).
    \item \textbf{Predicted:} The label predicted by the PDT model (\texttt{1} for increase, \texttt{0} for decrease).
    \item \textbf{Close:} The closing price of the stock at the specified timestamp.
\end{itemize}

An example of the \textbf{results\_df} is shown in Table~\ref{tab:results_df_example}, which illustrates the structure of this DataFrame.

\begin{table}[!h]
\centering
\caption{Example Structure of Results DataFrame (\texttt{results\_df})}
\label{tab:results_df_example}
\begin{tabular}{|c|c|c|c|}
\hline
\textbf{Datetime} & \textbf{Actual} & \textbf{Predicted} & \textbf{Close} \\ \hline
2024-10-25 13:25:00 & 0 & 0 & 1487.87 \\ \hline
2024-10-25 13:30:00 & 1 & 1 & 1488.50 \\ \hline
2024-10-25 13:35:00 & 0 & 0 & 1488.37 \\ \hline
2024-10-25 13:40:00 & 1 & 1 & 1488.45 \\ \hline
2024-10-25 13:45:00 & 0 & 0 & 1491.30 \\ \hline
\end{tabular}
\end{table}
\section{Results and Discussions}

In this section, we describe how each model (PDT, LSTM, RNN) and the Buy-and-Hold baseline 
was evaluated using our \textit{simulation process} (Algorithm~\ref{alg:simulation_process}), 
followed by a detailed discussion of the Permutation Decision Tree (PDT) results.
For PDT, LSTM and RNN, we have evaluated their performances across all 50 stocks that come under the NIFTY 50 index and aggregated the stocks performance by taking an average for each model.
Over and above this, we also have tested PDT on Forex pairs EURUSD and XAUUSD as mentioned in the begining of this paper.
The results for each stock under NIFTY 50 for PDT is available in the appendix.

\subsection{Simulation Algorithm (for All Models)}
\label{sec:sim_algo}

Before reviewing individual model performance, we provide the core simulation algorithm 
(Algorithm \ref{alg:simulation_process}) that executes trades based on \emph{Predicted} signals.


\begin{lrbox}{\algobox}
\begin{minipage}{0.97\textwidth}
\scriptsize
\begin{algorithm}[H]
\caption{\textsc{Simulation Process} -- Executes Buy/Sell Decisions Based on PDT Predictions.}
\label{alg:simulation_process}
\begin{algorithmic}[1]
    \REQUIRE 
        \(\mathit{stock\_symbol}\),
        \(\mathit{results\_df}\) with columns \{\texttt{Datetime, Actual, Predicted, Close}\},
        Initial balance \(B_0\),
        Trailing stop-loss \(\delta\)

    \ENSURE Array of portfolio values \(\mathit{P} = [P_1, P_2, \ldots, P_T]\) and final metrics

    \STATE \(B \gets B_0\) \quad // \emph{current balance}
    \STATE \(\mathit{positions} \gets 0\)
    \STATE \(\mathit{trade\_count} \gets 0\)
    \STATE \(\mathit{successful\_trades} \gets 0\)
    \STATE \(\mathit{P} \gets [\,]\) \quad // \emph{list of portfolio values over time}
    \STATE \(\mathit{entry\_price} \gets \text{None}\)
    \STATE \(\mathit{trailing\_stop} \gets \text{None}\)

    \FOR{\(t = 1 \text{ to } T\)}
        \STATE \(\mathit{pred} \gets \mathit{results\_df}[t][\text{Predicted}]\)
        \STATE \(\mathit{price} \gets \mathit{results\_df}[t][\text{Close}]\)

        \IF{\(\mathit{pred} = 1 \wedge \mathit{positions} = 0\)}
            \STATE \(\mathit{positions} \gets \lfloor B / \mathit{price} \rfloor\)
            \STATE \(\mathit{entry\_price} \gets \mathit{price}\)
            \STATE \(B \gets B - (\mathit{positions} \times \mathit{price})\)
            \STATE \(\mathit{trailing\_stop} \gets \mathit{entry\_price} \times (1 - \delta)\)
        \ENDIF

        \IF{\(\mathit{positions} > 0\)}
            \STATE \(\mathit{trailing\_stop} \gets \max(\mathit{trailing\_stop},\; \mathit{price} \times (1 - \delta))\)

            \IF{\(\mathit{price} \le \mathit{trailing\_stop} \,\lor\, \mathit{pred} = 0\)}
                \STATE \(\mathit{profit} \gets \mathit{positions} \times (\mathit{price} - \mathit{entry\_price})\)
                \STATE \(B \gets B + (\mathit{positions} \times \mathit{price})\)
                \STATE \(\mathit{trade\_count} \gets \mathit{trade\_count} + 1\)
                \STATE \(\mathit{successful\_trades} \gets \mathit{successful\_trades} + \mathbb{1}_{(\mathit{profit} > 0)}\)
                \STATE \(\mathit{positions} \gets 0\)
            \ENDIF
        \ENDIF

        \STATE \(\mathit{P}_t \gets B + (\mathit{positions} \times \mathit{price})\)
        \STATE Append \(\mathit{P}_t\) to \(\mathit{P}\)
    \ENDFOR

    \IF{\(\mathit{positions} > 0\)}
        \STATE \(\mathit{profit} \gets \mathit{positions} \times (\mathit{price} - \mathit{entry\_price})\)
        \STATE \(B \gets B + (\mathit{positions} \times \mathit{price})\)
        \STATE \(\mathit{positions} \gets 0\)
    \ENDIF

    \STATE \(\text{Total Return \%} \gets \Bigl(\frac{B - B_0}{B_0}\Bigr) \times 100\)
    \STATE \(\text{Trading Accuracy \%} \gets 
            \Bigl(\frac{\mathit{successful\_trades}}{\mathit{trade\_count}}\Bigr) \times 100\)

    \RETURN \(\mathit{P}\) \quad // \emph{Full series of portfolio values}
\end{algorithmic}
\end{algorithm}
\end{minipage}
\end{lrbox}
\noindent\resizebox{\textwidth}{!}{\usebox{\algobox}}

\subsection{Permutation Decision Tree (PDT) Performance}

\noindent
We now present the results of applying the \textbf{Permutation Decision Tree (PDT)} model. 
While this shorter time frame captures only short-term trends, it demonstrates the PDT’s 
capability to adapt quickly to market fluctuations.

\paragraph{Key Observations:}
\begin{itemize}
    \item \textbf{NIFTY 50 Performance aggregation}
    \begin{itemize}
        \item \textbf{PDT Strategy:}
        \begin{itemize}
            \item \textbf{Average Return (Testing period):} 1.1802\% per stock.
        \end{itemize}
        \item \textbf{Buy-and-Hold Strategy:}
        \begin{itemize}
            \item \textbf{Average Return (Testing period):} -2.29\% per stock.
        \end{itemize}
    \end{itemize}
    \item \textbf{EURUSD Performance}
        \begin{itemize}
        \item \textbf{PDT Strategy:}
        \begin{itemize}
            \item \textbf{Return (Testing period):} 2.3105\%.
        \end{itemize}
        \item \textbf{Buy-and-Hold Strategy:}
        \begin{itemize}
            \item \textbf{Return (Testing period):} 1.6246\%.
        \end{itemize}
    \end{itemize}
        \item \textbf{XAUUSD Performance}
        \begin{itemize}
        \item \textbf{PDT Strategy:}
        \begin{itemize}
            \item \textbf{Return (Testing period):} 0.0006\%.
        \end{itemize}
        \item \textbf{Buy-and-Hold Strategy:}
        \begin{itemize}
            \item \textbf{Return (Testing period):} 0.0021\%.
        \end{itemize}
    \end{itemize}
\end{itemize}
\textit{Performance Analysis:}
The PDT-based approach demonstrated \textbf{superior returns} compared to buy-and-hold for NIFTY 50 stocks on an average and in EURUSD, 
indicating its robustness and adaptability for short-term trading. The inclusion of a 
trailing stop-loss in Algorithm~\ref{alg:simulation_process} helped to curtail losses, 
preserving capital during sudden market reversals. Even though PDT underperformed in XAUUSD, we still made a profit (positive return).

\textit{Limitations Due to Limited Data:}
Because the evaluation spanned only around 3 months for NIFTY 50 and around 5 months for the Forex pairs, these findings mainly reflect short-term
performance.

\begin{figure}[!ht]
\centering
\includegraphics[width=\textwidth]{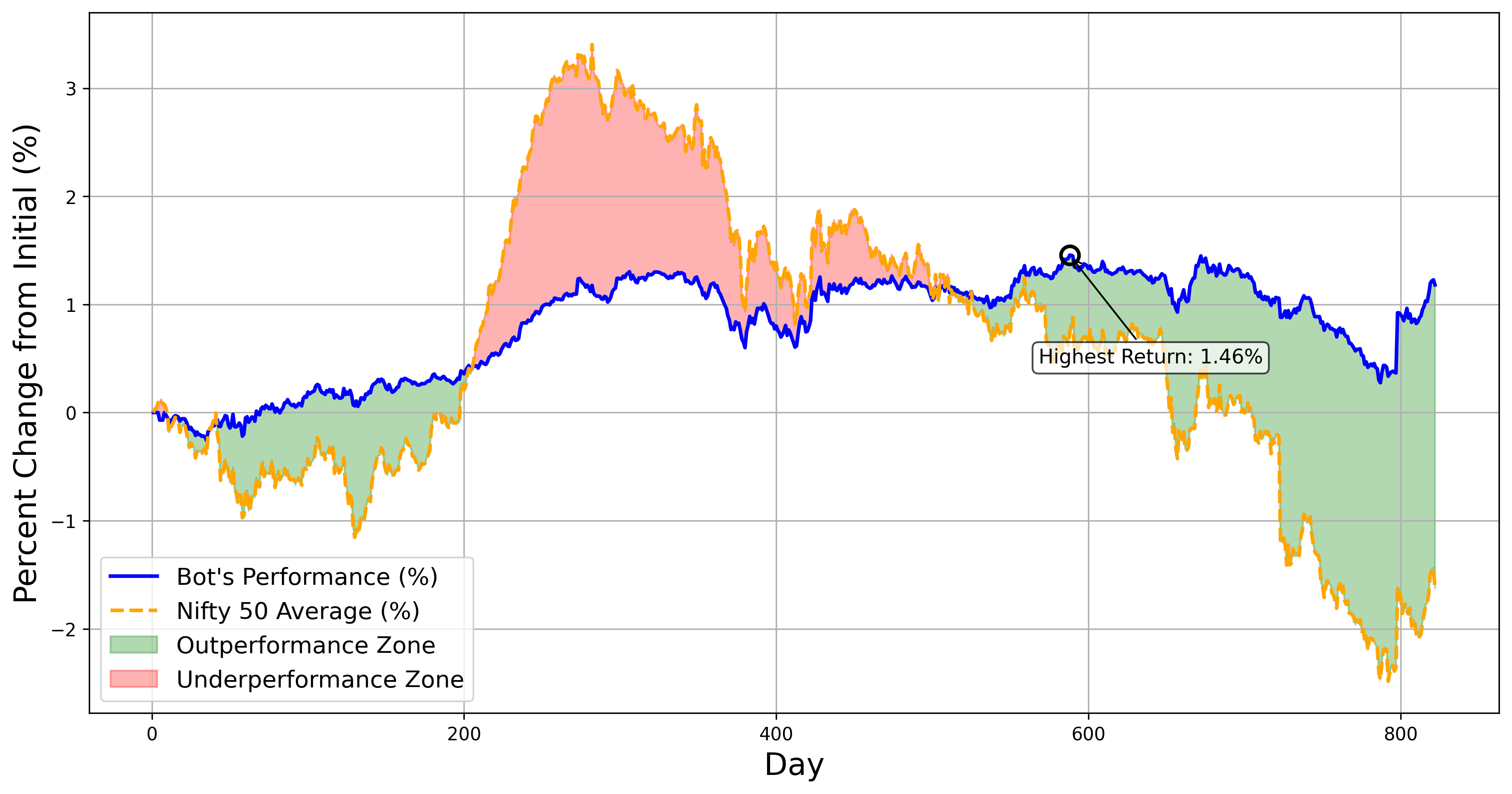}
\caption{Portfolio performance comparison (PDT vs. Buy-and-Hold) over 3 testing-month period. 
         PDT shows \emph{steady growth}, whereas buy-and-hold experiences significant losses.}
\label{fig:portfolio_performance}
\end{figure}

\textit{Visual Representation of Portfolio Performance:}
Figure~\ref{fig:portfolio_performance} compares the \textbf{PDT-based strategy} and 
\textit{buy-and-hold} for NIFTY 50 stocks on an average in terms of portfolio value growth. PDT consistently outperforms, 
incurring far fewer drawdowns.

\begin{figure}[!ht]
    \centering
    \includegraphics[width=\textwidth]{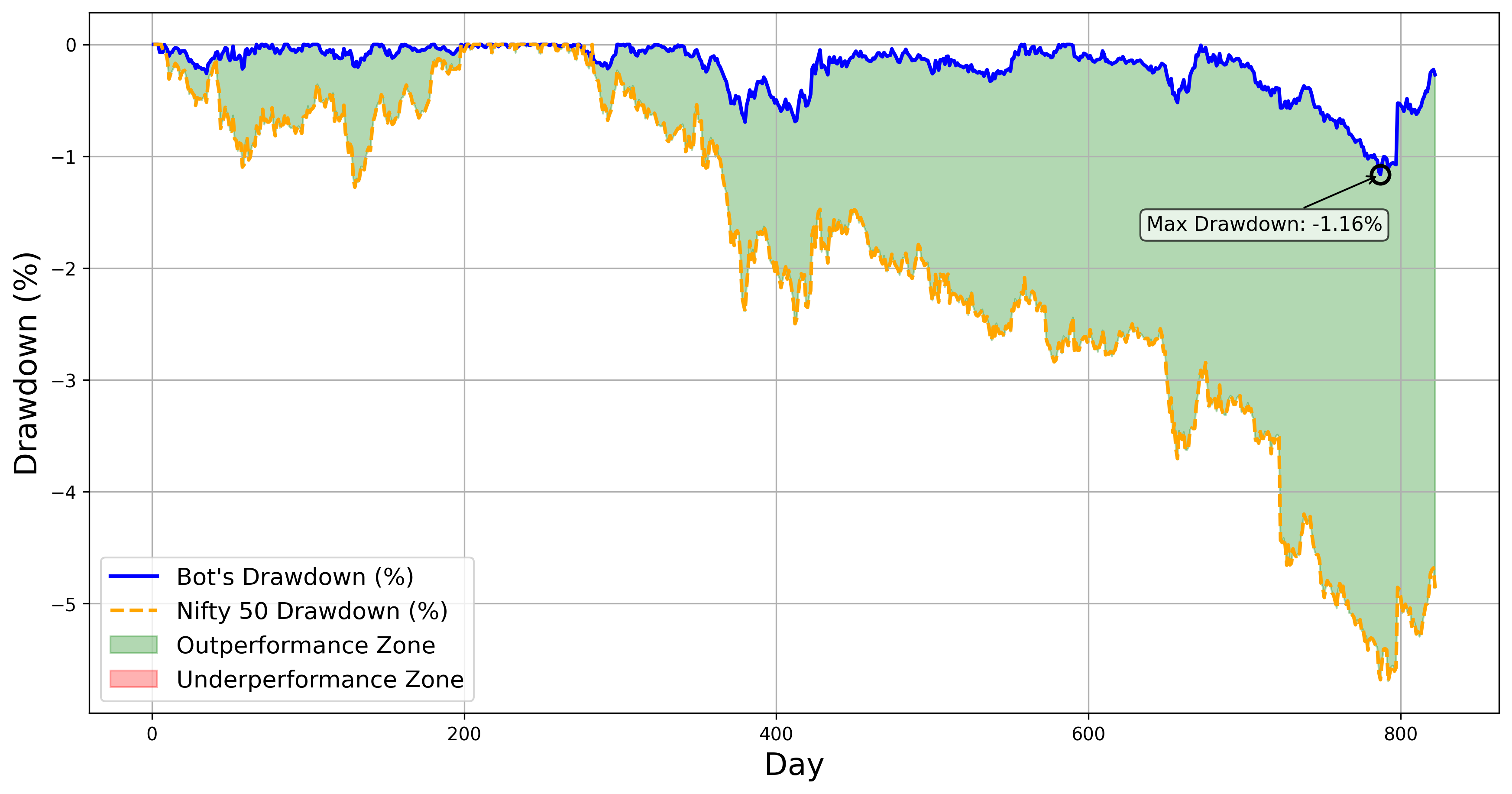}
    \caption{Drawdown for PDT vs. Buy-and-Hold.}
    \label{fig:pdt_drawdown}
\end{figure}

\paragraph{Drawdown Analysis (PDT vs. Buy-and-Hold):}
Figure~\ref{fig:pdt_drawdown} illustrates the drawdown curves for the NIFTY 50 stocks on an average.

\subsection{LSTM and RNN Performance}
We begin by evaluating two deep learning models Long Short-Term Memory (LSTM)
and Recurrent Neural Network (RNN)—on the NIFTY 50 stocks dataset. We have used these models since they are commonly used in stock price predictions.

\paragraph{Hyperparameters and Implementation:}
The hyperparameters for the LSTM and RNN models are chosen to balance 
computational complexity and performance:
\begin{itemize}
    \item \textbf{Number of Units (50):} Provides sufficient capacity for learning patterns.
    \item \textbf{Dense Layer (10 units):} Additional layer to capture complex relationships.
    \item \textbf{Activation Functions:} 
    \begin{itemize}
        \item \textbf{ReLU:} Used in the dense layer.
        \item \textbf{Sigmoid:} Applied to the output layer for binary classification.
    \end{itemize}
    \item \textbf{Loss Function:} Binary cross-entropy.
    \item \textbf{Optimizer:} Adam.
    \item \textbf{Epochs:} 50.
\end{itemize}

\noindent
The training processes for LSTM and RNN are given in 
Algorithms~\ref{alg:train_lstm} and~\ref{alg:train_rnn}.

\begin{algorithm}[ht]
\caption{\textsc{TrainLSTM} -- Builds and Trains an LSTM Model}
\label{alg:train_lstm}
\begin{algorithmic}[1]
    \REQUIRE 
        Feature matrix \(\mathit{X}\), 
        Target labels \(\mathit{y}\), 
        Train-test split index \(\mathit{split\_idx}\)

    \ENSURE Trained LSTM model, predicted test labels \(\mathit{y\_pred}\)

    \STATE Reshape \(\mathit{X}\) into a 3D tensor: \((\text{samples}, \text{timesteps}=1, \text{features})\)
    \STATE Split \(\mathit{X}, \mathit{y}\) into training and testing sets

    \STATE Build an LSTM model:
    \begin{itemize}
        \item One LSTM layer with 50 units
        \item One Dense layer with 10 units and ReLU activation
        \item Output layer with 1 unit and Sigmoid activation
    \end{itemize}

    \STATE Compile using Adam and binary cross-entropy loss
    \STATE Train for 50 epochs, batch size = 32, learning rate = 0.001
    \STATE Predict probabilities on the test set, threshold at 0.5 to get binary labels

    \RETURN \(\mathit{model}, \mathit{y\_pred}\)
\end{algorithmic}
\end{algorithm}

\begin{algorithm}[ht]
\caption{\textsc{TrainRNN} -- Builds and Trains a Simple RNN}
\label{alg:train_rnn}
\begin{algorithmic}[1]
    \REQUIRE
        Feature matrix \(\mathit{X}\),
        Target labels \(\mathit{y}\),
        Train-test split index \(\mathit{split\_idx}\)

    \ENSURE Trained RNN model, predicted test labels \(\mathit{y\_pred}\)

    \STATE Reshape \(\mathit{X}\) into a 3D tensor: \((\text{samples}, \text{timesteps}=1, \text{features})\)
    \STATE Split \(\mathit{X}, \mathit{y}\) into training and testing sets

    \STATE Build an RNN model:
    \begin{itemize}
        \item One SimpleRNN layer with 50 units
        \item One Dense layer with 10 units and ReLU activation
        \item Output layer with 1 unit and Sigmoid activation
    \end{itemize}

    \STATE Compile using Adam and binary cross-entropy loss
    \STATE Train for 50 epochs, batch size = 32, learning rate = 0.001
    \STATE Predict probabilities on the test set, threshold at 0.5 to get binary labels

    \RETURN \(\mathit{model}, \mathit{y\_pred}\)
\end{algorithmic}
\end{algorithm}

\begin{figure}[!ht]
    \centering
    \includegraphics[width=\textwidth]{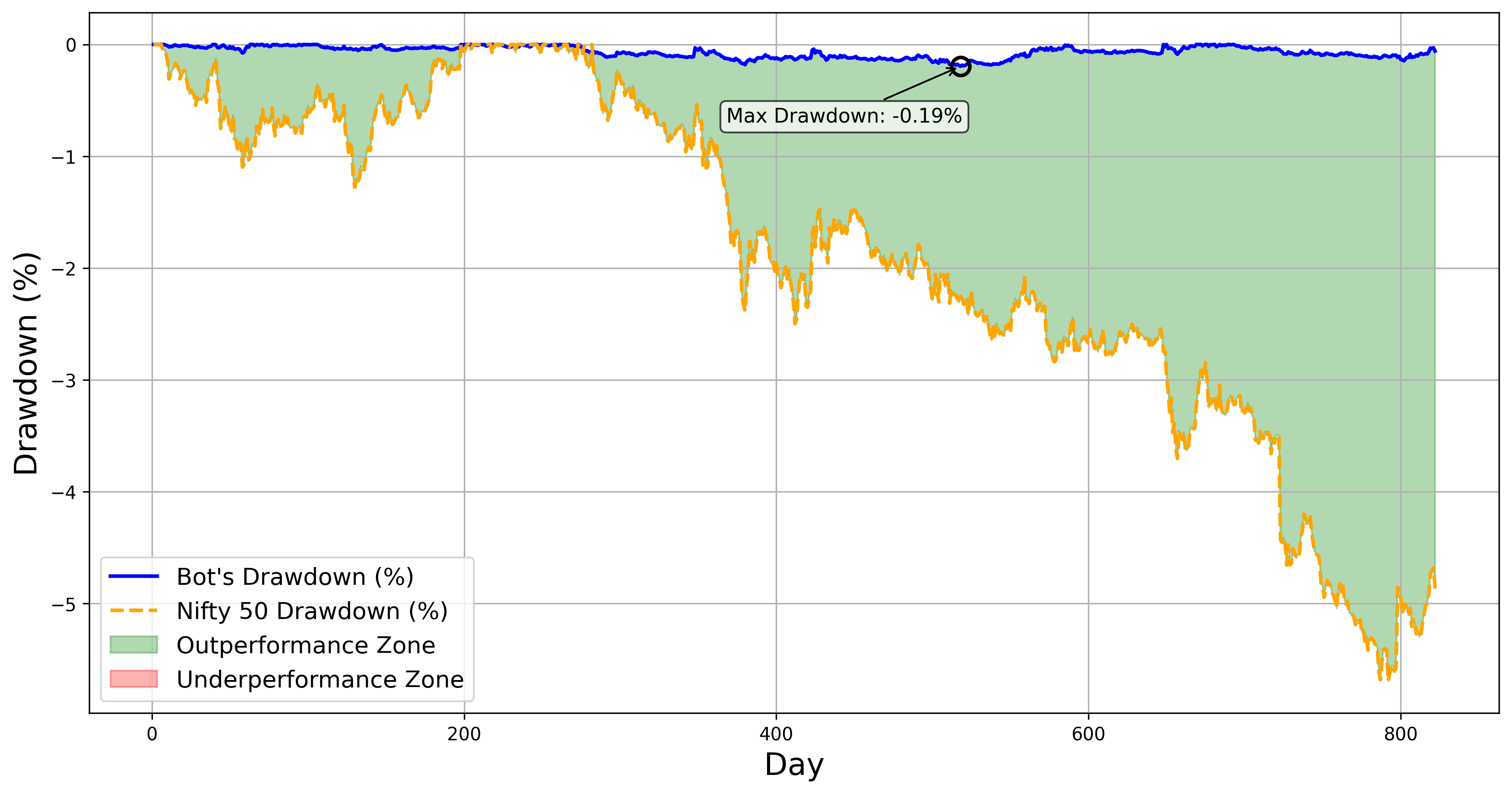}
    \caption{Drawdown for LSTM.}
    \label{fig:lstm_drawdown}
\end{figure}

\begin{figure}[!ht]
    \centering
    \includegraphics[width=\textwidth]{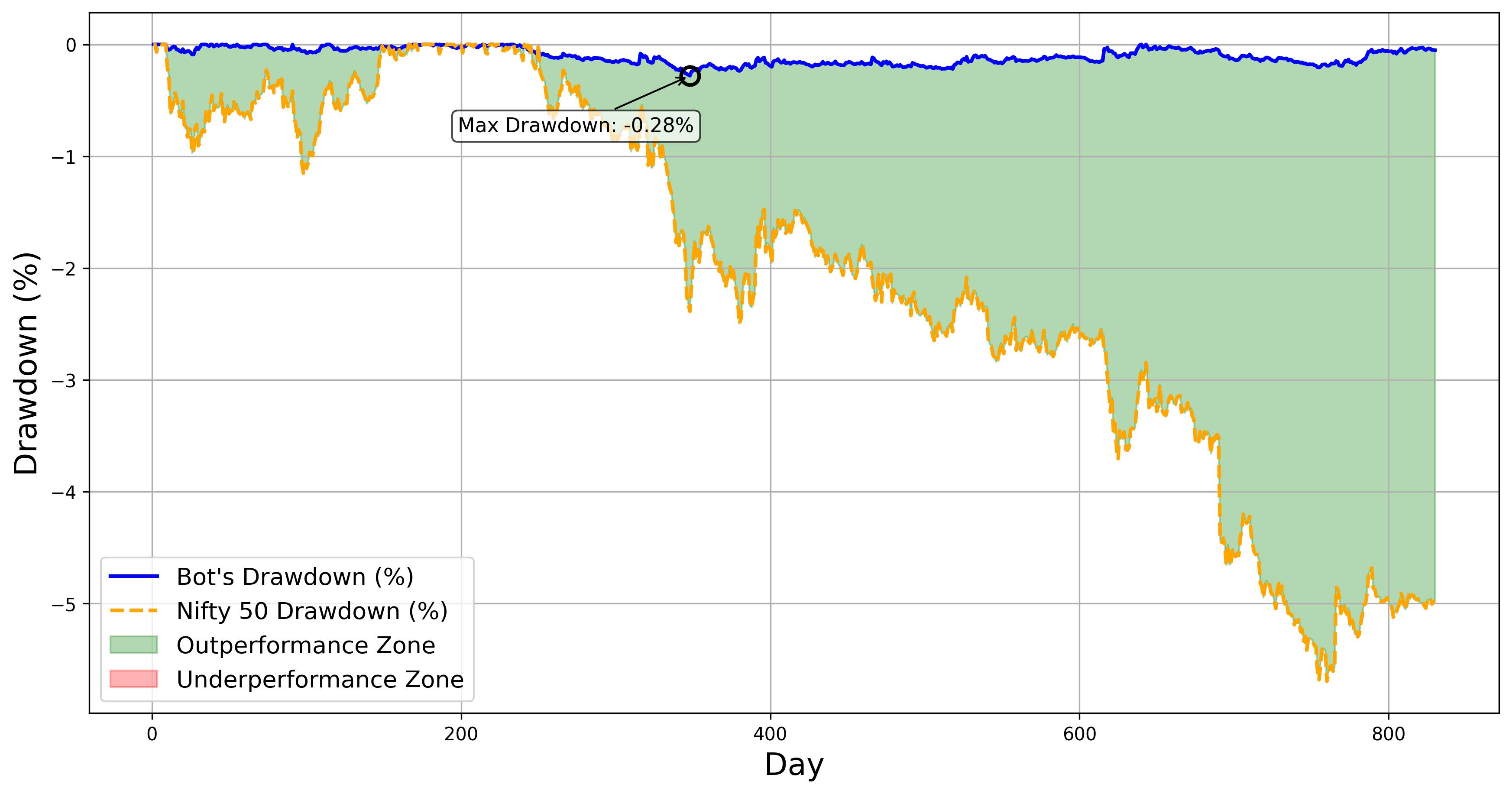}
    \caption{Drawdown for RNN.}
    \label{fig:rnn_drawdown}
\end{figure}

\begin{figure}[!ht]
    \centering
    \includegraphics[width=\textwidth]{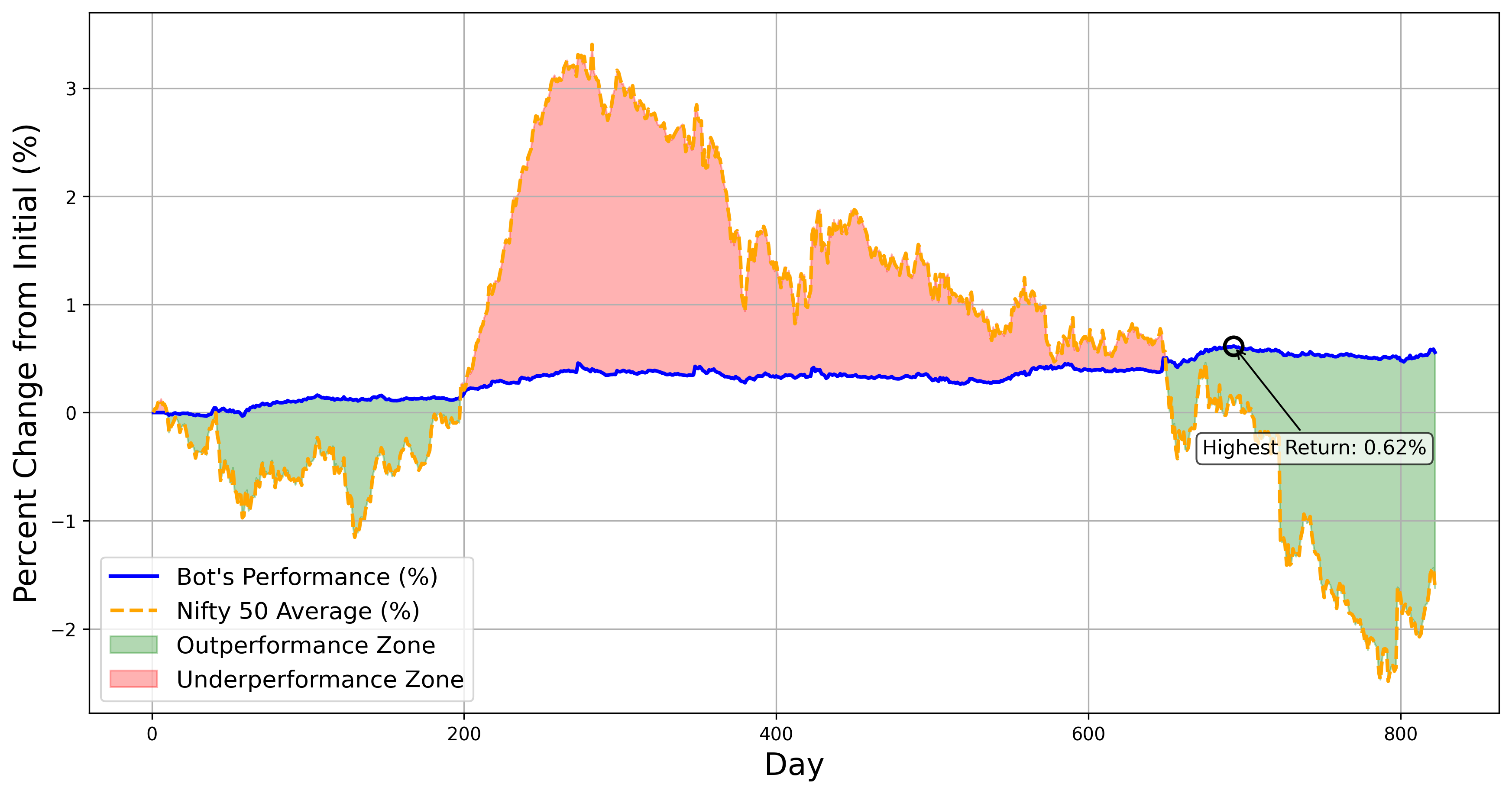}
    \caption{Portfolio Value: LSTM vs. Buy-and-Hold.}
    \label{fig:lstm_vs_buy_and_hold}
\end{figure}

\begin{figure}[!ht]
    \centering
    \includegraphics[width=\textwidth]{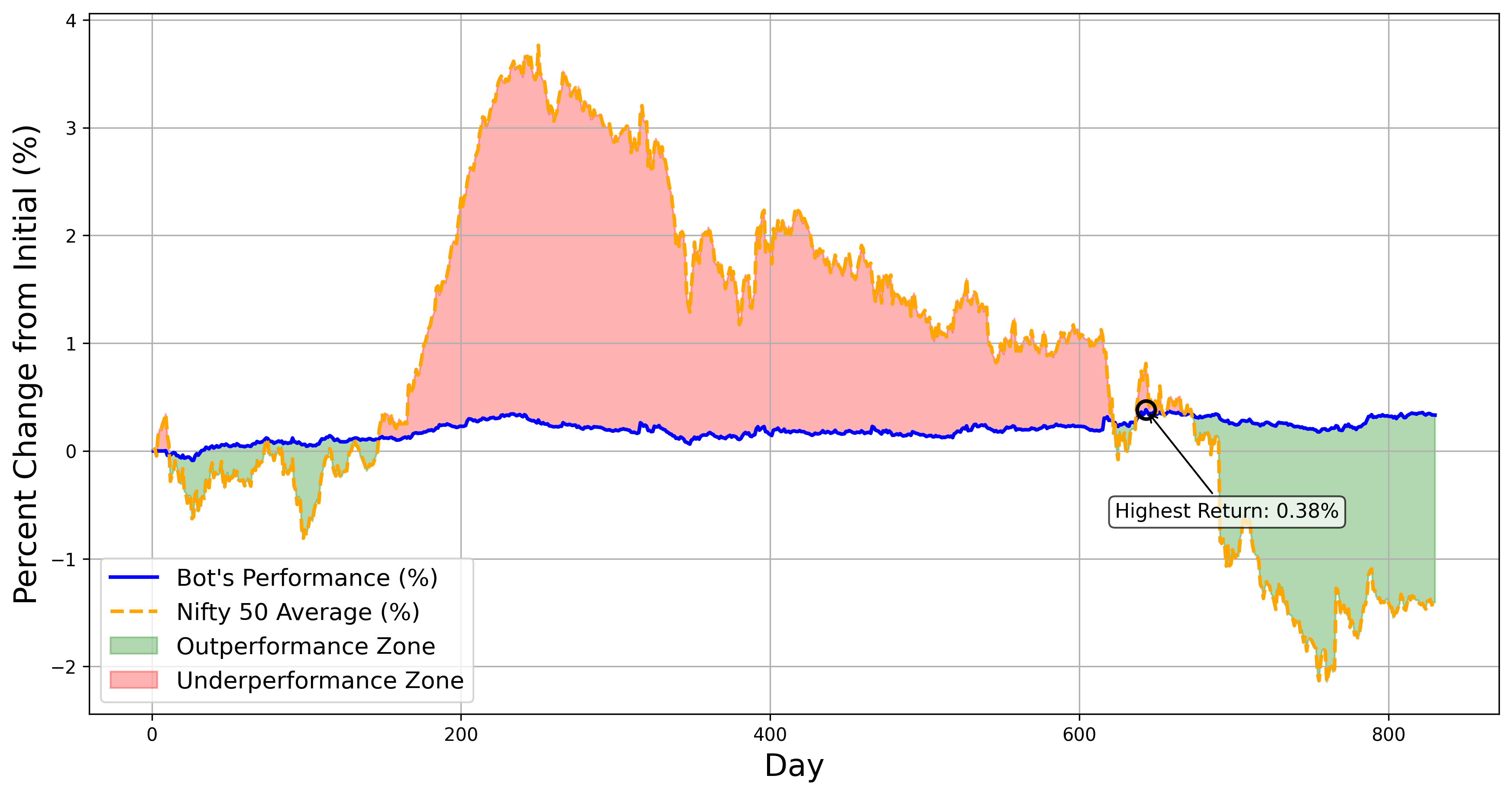}
    \caption{Portfolio Value: RNN vs. Buy-and-Hold.}
    \label{fig:rnn_vs_buy_and_hold}
\end{figure}

\noindent
\textbf{Drawdown and Portfolio Performance:} Figures~\ref{fig:lstm_drawdown} 
and~\ref{fig:rnn_drawdown} illustrate the drawdown curves for LSTM and RNN respectively for the NIFTY 50 stocks on an average, while 
Figures~\ref{fig:lstm_vs_buy_and_hold} and~\ref{fig:rnn_vs_buy_and_hold} compare each model’s 
portfolio value to that of a naive buy-and-hold approach.

\noindent
\textbf{Overall}, LSTM and RNN (NIFTY 50 stocks, on an average):
\begin{itemize}
    \item Provide improved returns over Buy-and-Hold in certain time windows.
    \item Are highly data-dependent and can struggle in sudden market shifts.
\end{itemize}


\subsection{Comparison with Baseline Models}

In addition to Buy-and-Hold, we benchmark PDT against \textbf{LSTM} and \textbf{RNN} 
(Table~\ref{tab:performance_comparison}).

\begin{table}[!h]
\centering
\small
\caption{Performance Comparison of PDT, LSTM, RNN, and Buy-and-Hold.}
\label{tab:performance_comparison}
\begin{tabular}{lcc}
\toprule
\textbf{Model} & \textbf{Testing Period Growth (\%)} & \textbf{Max Drawdown (\%)} \\
\midrule
PDT           & 1.1802& 1.16\\
LSTM          & 0.557& 0.19\\
RNN           & 0.5896& 0.28\\
Buy-and-Hold  & -2.29& 5.6955\\
\bottomrule
\end{tabular}%
\end{table}

\noindent
\subsection{Discussions}
\begin{itemize}
    \item \textbf{PDT vs. LSTM/RNN:} PDT exhibits higher returns (1.1802\% on NIFTY 50 on an average), 
          suggesting that its \emph{Effort To Compress (\textit{ETC})} metric identifies clearer 
          thresholds for stock movement. LSTM/RNN (0.557\% and 0.5896\% on an average, respectively) require larger datasets to fully 
          capture temporal patterns.
    \item \textbf{Risk/Reward Profile:} The trailing stop-loss used in our simulation 
          helps PDT lock in gains and limit downside, whereas LSTM/RNN, in the same 
          simulation, do not show as effective a drawdown control.
    \item \textbf{Buy-and-Hold Vulnerability:} With a negative testing period growth 
          of -2.29\% and large drawdowns, this passive strategy struggles in a volatile 
          3 month market window.
    \item \textbf{Overall Conclusion:} In a short 3 month study, PDT outperforms both 
          deep learning baselines and buy-and-hold, but more extensive tests over 
          multiple market cycles are advised to confirm long-term robustness.
\end{itemize}

\section{Limitations and Future Research Directions}

Permutation Decision Trees (PDT) have shown promising results in stock price prediction; however, several limitations and strengths must be highlighted for balanced evaluation. PDT heavily relies on quality input features; noisy or incomplete data significantly reduces performance. PDTs cannot be incrementally updated. With new data, the entire tree must be reconstructed, limiting adaptability in dynamic markets. Future research should address these limitations, specifically by incorporating dynamic updating of PDT, inclusion of transaction costs and slippage, and better modeling of sequential dependencies.




\section{Conclusion}
In this 3 month study, the Permutation Decision Tree (PDT) demonstrates a superior performance profile compared to LSTM and RNN models. PDT exhibits higher returns suggesting that its Effort To Compress (\textit{ETC}) metric is more effective in identifying key thresholds for stock movement within short-term, high-frequency data. In contrast, LSTM and RNN models appear to require substantially larger datasets to effectively capture the temporal patterns necessary for accurate short-term predictions.
The trailing stop-loss mechanism implemented in our simulation proves beneficial for PDT, enabling it to secure gains and mitigate downside risk. Furthermore, a passive buy-and-hold strategy demonstrates significant vulnerability in the volatile market conditions of the testing period, resulting in a negative growth of -2.25\% and substantial drawdowns. However, given the short timeframe, further research is recommended. Specifically, more extensive testing across multiple market cycles is necessary to validate the long-term robustness and generalizability of the PDT-based trading system.
\section*{Code Availability} 
The codes used in this research paper are available in the following GitHub repository: \url{https://github.com/i-to-the-power-i/pdt-stocks}
\section*{Acknowledgments}
Harikrishnan N. B. gratefully acknowledges the financial support provided by the New Faculty Seed Grant (NFSG/GOA/2024/G0906) from BITS Pilani, K. K. Birla Goa Campus. 

\section*{Appendix}\label{sec:appendix}
\subsection{Comparison of PDT vs Buy \& Hold}

The difference in returns \(\Delta = \text{PDT Return} - \text{Buy \& Hold Return}\) for each stock is shown below.

\begin{longtable}{l
                S[table-format=+2.2]
                S[table-format=+2.2]
                S[table-format=+3.2]}
\caption{PDT vs Buy \& Hold Returns for NIFTY 50 Stocks.}\label{tab:pdt-vs-bh}\\
\hline
\textbf{Stock} & \textbf{PDT Return (\%)} & \textbf{Buy \& Hold Return (\%)} & \textbf{Difference (\%)} \\
\hline
\endfirsthead
\hline
\textbf{Stock} & \textbf{PDT Return (\%)} & \textbf{Buy \& Hold Return (\%)} & \textbf{Difference (\%)} \\
\hline
\endhead
ASIANPAINT.NS  &  0.88 & -0.20 &  1.08 \\
BRITANNIA.NS   &  5.43 &  3.11 &  2.32 \\
CIPLA.NS       &  0.02 & -5.23 &  5.25 \\
EICHERMOT.NS   &  5.61 &  1.93 &  3.68 \\
NESTLEIND.NS   &  4.39 &  2.28 &  2.11 \\
GRASIM.NS      & -5.69 & -5.63 & -0.06 \\
HEROMOTOCO.NS  &  3.09 & -1.96 &  5.05 \\
HINDALCO.NS    &  0.38 & -4.53 &  4.91 \\
HINDUNILVR.NS  &  4.09 &  3.19 &  0.90 \\
ITC.NS         &  0.27 & -8.11 &  8.38 \\
LT.NS          & -0.78 & -3.46 &  2.68 \\
M\&M.NS        &  0.17 &  0.84 & -0.67 \\
RELIANCE.NS    &  2.25 &  1.39 &  0.86 \\
TATACONSUM.NS  &  4.86 &  7.07 & -2.21 \\
TATAMOTORS.NS  &  7.93 &  4.25 &  3.68 \\
TATASTEEL.NS   & -1.42 & -8.99 &  7.57 \\
WIPRO.NS       & -0.21 & -3.81 &  3.60 \\
APOLLOHOSP.NS  &  0.06 & -7.88 &  7.94 \\
DRREDDY.NS     & -4.06 & -2.90 & -1.16 \\
TITAN.NS       &  2.12 &  1.18 &  0.94 \\
SBIN.NS        & -0.39 & -6.64 &  6.25 \\
SHRIRAMFIN.NS  & -2.73 & -9.19 &  6.46 \\
BPCL.NS        &  0.76 & -7.97 &  8.73 \\
KOTAKBANK.NS   &  1.90 & -1.87 &  3.77 \\
INFY.NS        &  3.00 &  2.33 &  0.67 \\
BAJFINANCE.NS  &  3.06 &  5.19 & -2.13 \\
ADANIENT.NS    & -0.23 & -5.35 &  5.12 \\
SUNPHARMA.NS   & -0.14 & -6.82 &  6.68 \\
JSWSTEEL.NS    &  2.55 & -0.42 &  2.97 \\
HDFCBANK.NS    & -0.88 & -9.06 &  8.18 \\
BAJAJFINSV.NS  &  0.71 &  7.73 & -7.02 \\
ULTRACEMCO.NS  &  0.00 & -7.56 &  7.56 \\
DIVISLAB.NS    &  3.03 & -1.81 &  4.84 \\
MARUTI.NS      &  0.00 &  7.29 & -7.29 \\
NTPC.NS        &  0.19 & -7.52 &  7.71 \\
HCLTECH.NS     &  4.02 & -4.30 &  8.32 \\
HDFCLIFE.NS    &  1.61 & -3.33 &  4.94 \\
COALINDIA.NS   & -2.58 & -3.27 &  0.69 \\
TECHM.NS       &  4.44 & -3.88 &  8.32 \\
GAIL.NS        &  3.42 & -7.46 & 10.88 \\
INDUSINDBK.NS  &  1.78 &  0.20 &  1.58 \\
UPL.NS         &  6.87 &  6.68 &  0.19 \\
SHREECEM.NS    &  0.00 & -3.57 &  3.57 \\
POWERGRID.NS   & -0.66 & -5.63 &  4.97 \\
DABUR.NS       &  2.54 &  2.35 &  0.19 \\
ACC.NS         &  1.78 & -7.47 &  9.25 \\
BAJAJ-AUTO.NS  &  0.96 & -3.03 &  3.99 \\
CYIENTDLM.NS   & -1.93 & -12.72 & 10.79 \\
BHARTIARTL.NS  & -2.11 &  1.39 & -3.50 \\
ZEEL.NS        & -1.35 & -1.33 & -0.02 \\
\hline
\end{longtable}
With this table, we can infer that most of the difference is positive which shows that PDT has actually out performed most of the stocks under NIFTY 50 index. In cases where PDT has underperformed with respect to the stocks under NIFTY 50, there are also cases where it still has made profits.

\subsection{PDT Profit vs Buy \& Hold Profit (Contiguous Chunks)}

\begin{longtable}{c S[table-format=+1.5] S[table-format=+1.5]}
\caption{PDT Profit vs Buy \& Hold Profit for XAUUSD Chunks.}\label{tab:chunks}\\
\hline
\textbf{Chunk} & \textbf{PDT Profit} & \textbf{Buy \& Hold Profit} \\
\hline
\endfirsthead
\hline
\textbf{Chunk} & \textbf{PDT Profit} & \textbf{Buy \& Hold Profit} \\
\hline
\endhead
1 &  0.00000 &  0.01633 \\
2 &  0.01164 &  0.02974 \\
3 &  0.01422 &  0.03807 \\
4 &  0.00622 &  0.02491 \\
5 &  0.00958 & -0.01924 \\
6 & -0.01163 & -0.01469 \\
7 & -0.00560 &  0.00098 \\
8 & -0.00797 & -0.01208 \\
\hline
\end{longtable}

\subsection{Drawdown and Portfolio Charts for XAUUSD Chunks}

\begin{figure}[H]
    \centering
    \begin{subfigure}[b]{0.48\textwidth}
        \includegraphics[width=\linewidth]{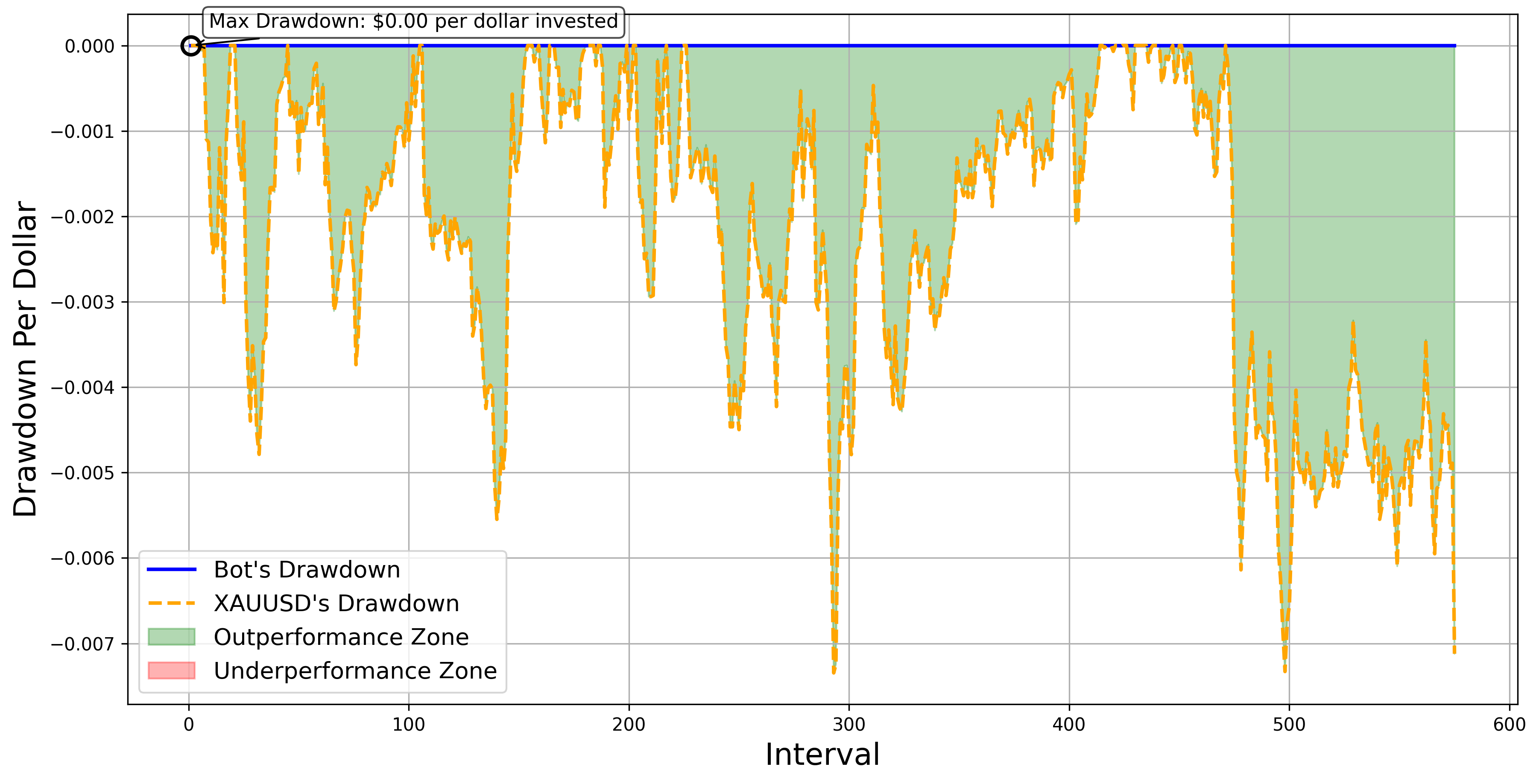}
        \caption{Drawdown}
    \end{subfigure}
    \hfill
    \begin{subfigure}[b]{0.48\textwidth}
        \includegraphics[width=\linewidth]{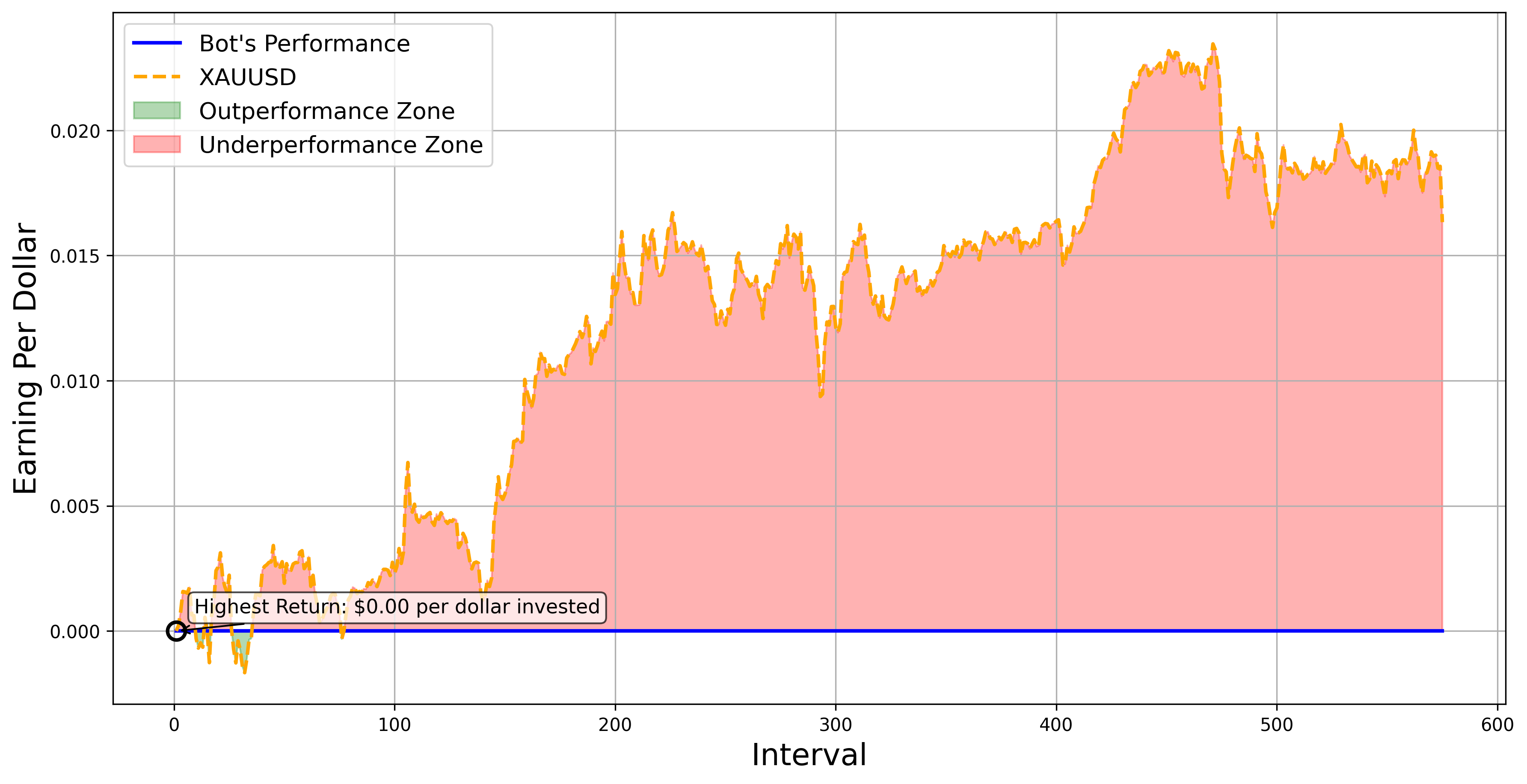}
        \caption{Portfolio}
    \end{subfigure}
    \caption{Chunk 1: Drawdown and Portfolio}
    \label{fig:chunk1}
\end{figure}

\begin{figure}[H]
    \centering
    \begin{subfigure}[b]{0.48\textwidth}
        \includegraphics[width=\linewidth]{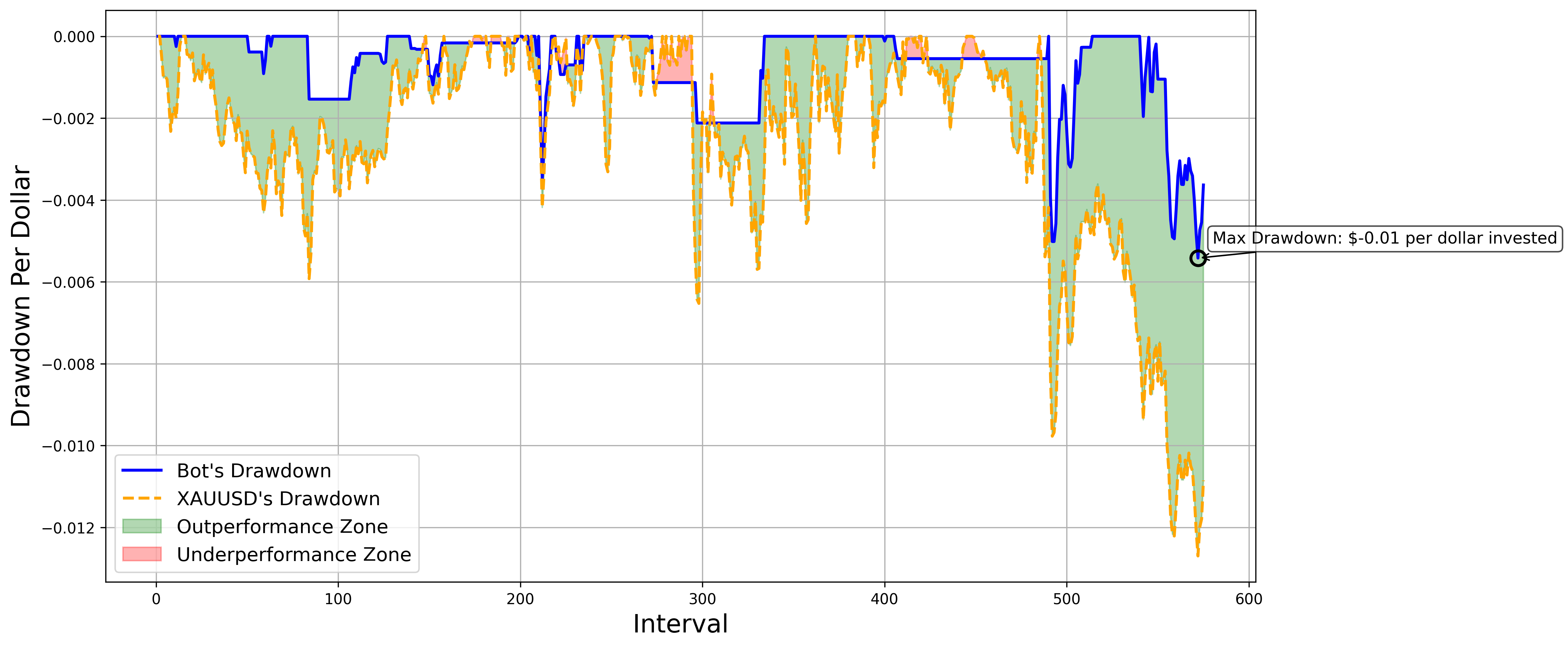}
        \caption{Drawdown}
    \end{subfigure}
    \hfill
    \begin{subfigure}[b]{0.48\textwidth}
        \includegraphics[width=\linewidth]{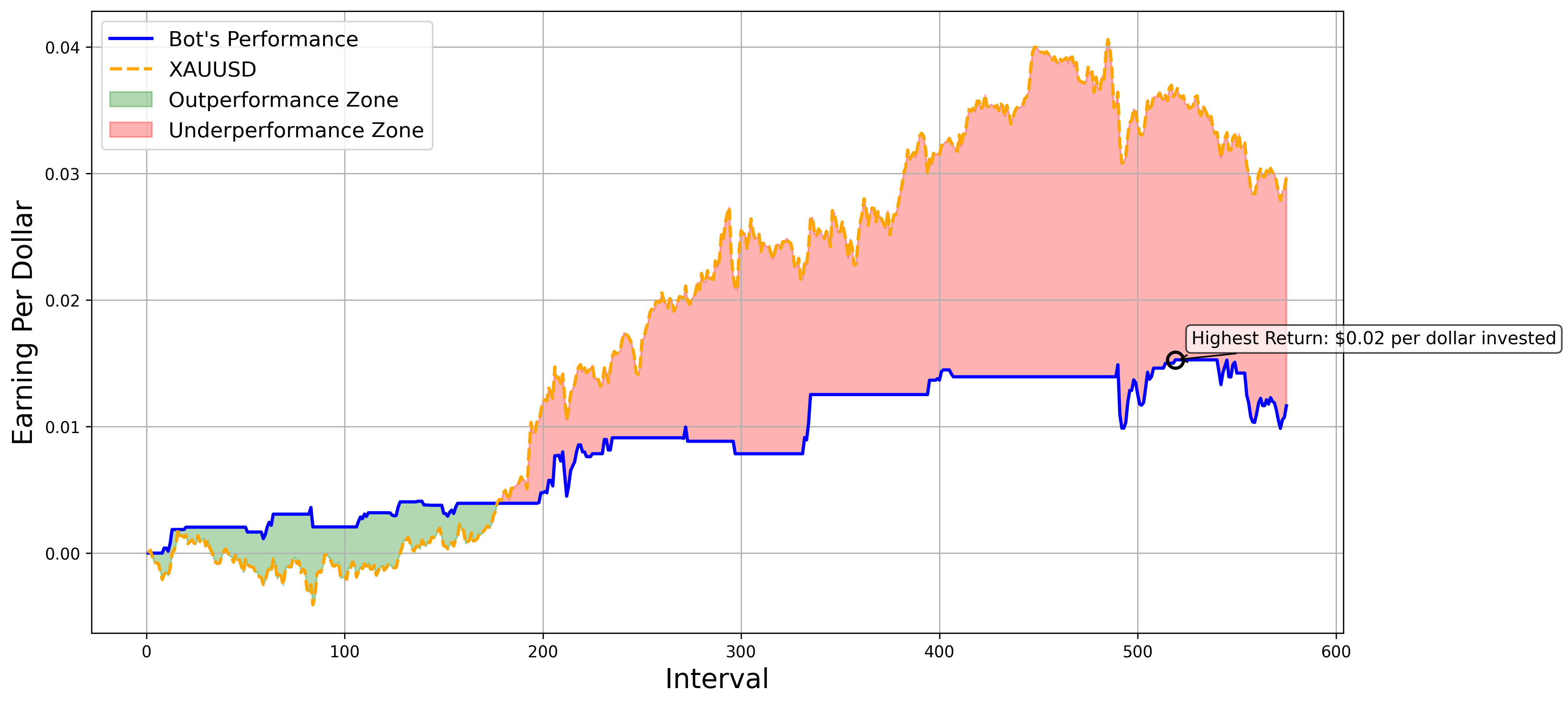}
        \caption{Portfolio}
    \end{subfigure}
    \caption{Chunk 2: Drawdown and Portfolio}
    \label{fig:chunk2}
\end{figure}

\begin{figure}[H]
    \centering
    \begin{subfigure}[b]{0.48\textwidth}
        \includegraphics[width=\linewidth]{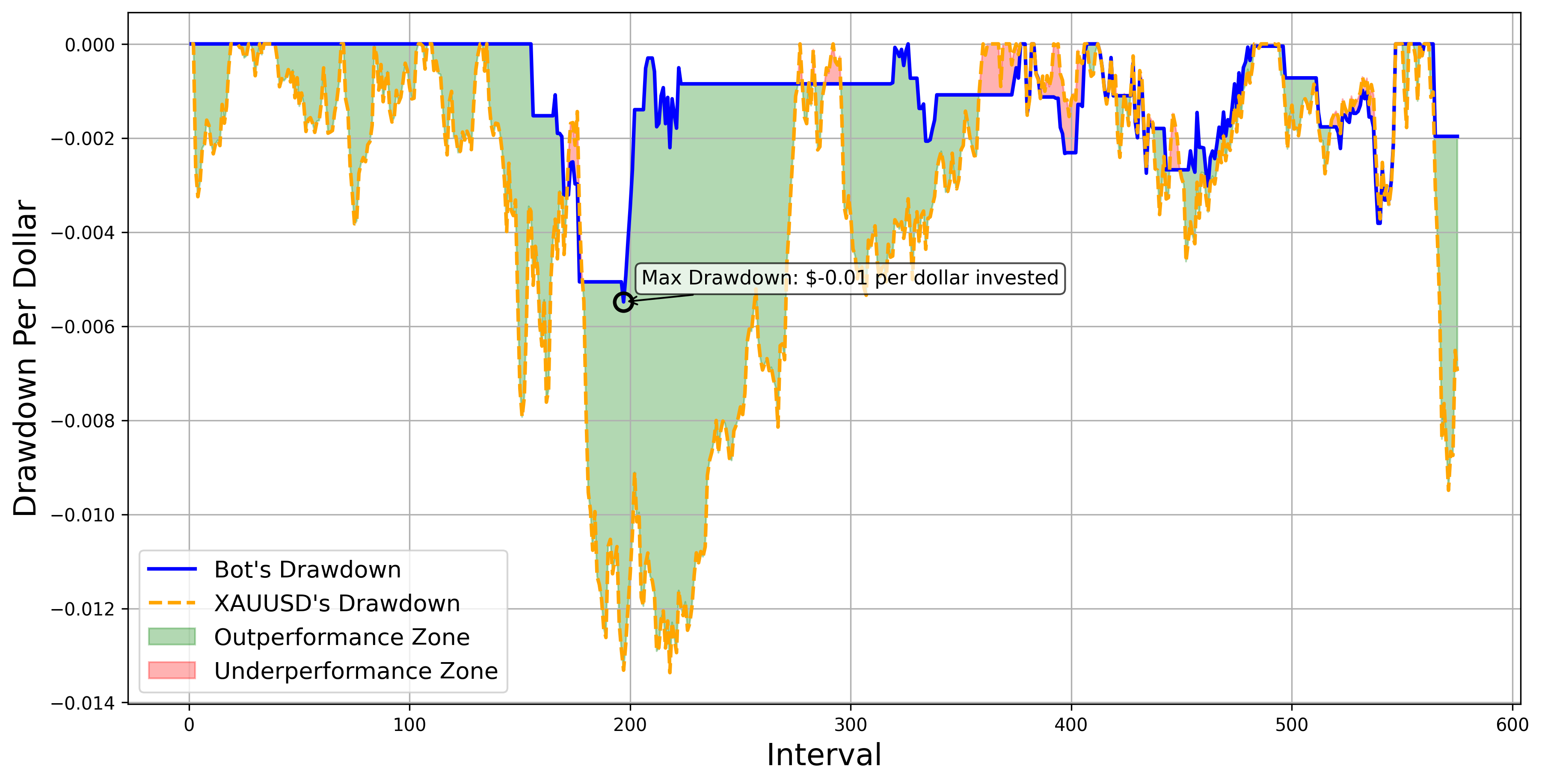}
        \caption{Drawdown}
    \end{subfigure}
    \hfill
    \begin{subfigure}[b]{0.48\textwidth}
        \includegraphics[width=\linewidth]{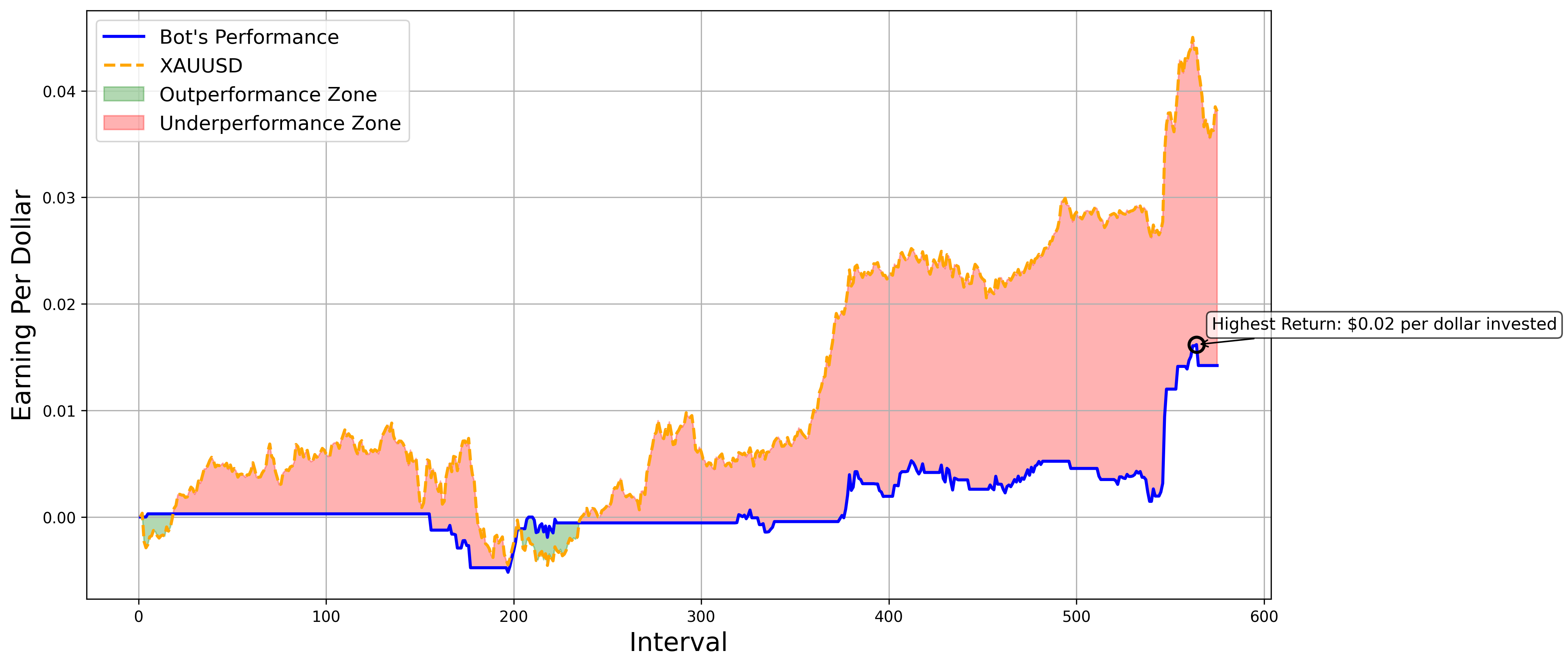}
        \caption{Portfolio}
    \end{subfigure}
    \caption{Chunk 3: Drawdown and Portfolio}
    \label{fig:chunk3}
\end{figure}

\begin{figure}[H]
    \centering
    \begin{subfigure}[b]{0.48\textwidth}
        \includegraphics[width=\linewidth]{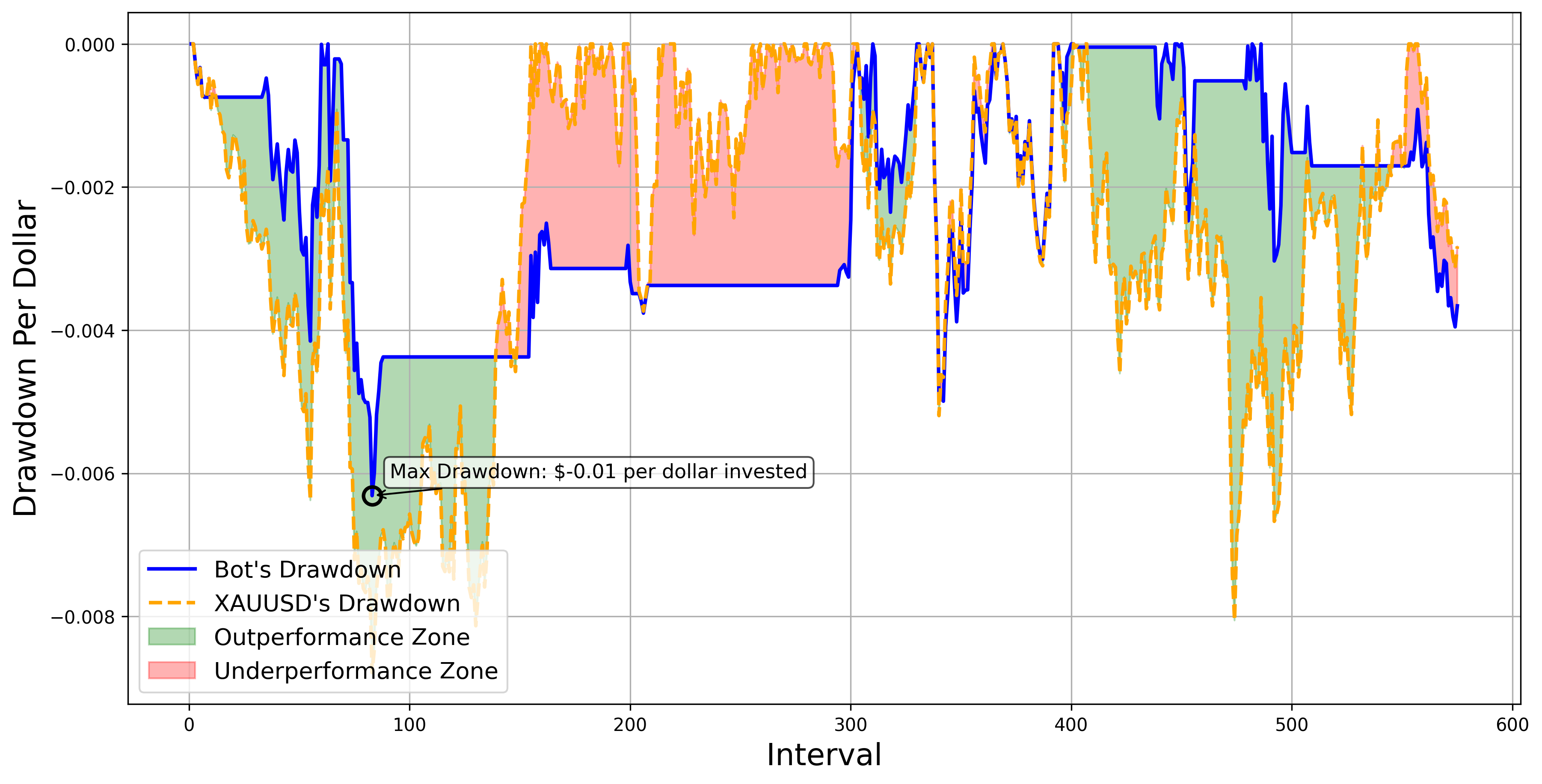}
        \caption{Drawdown}
    \end{subfigure}
    \hfill
    \begin{subfigure}[b]{0.48\textwidth}
        \includegraphics[width=\linewidth]{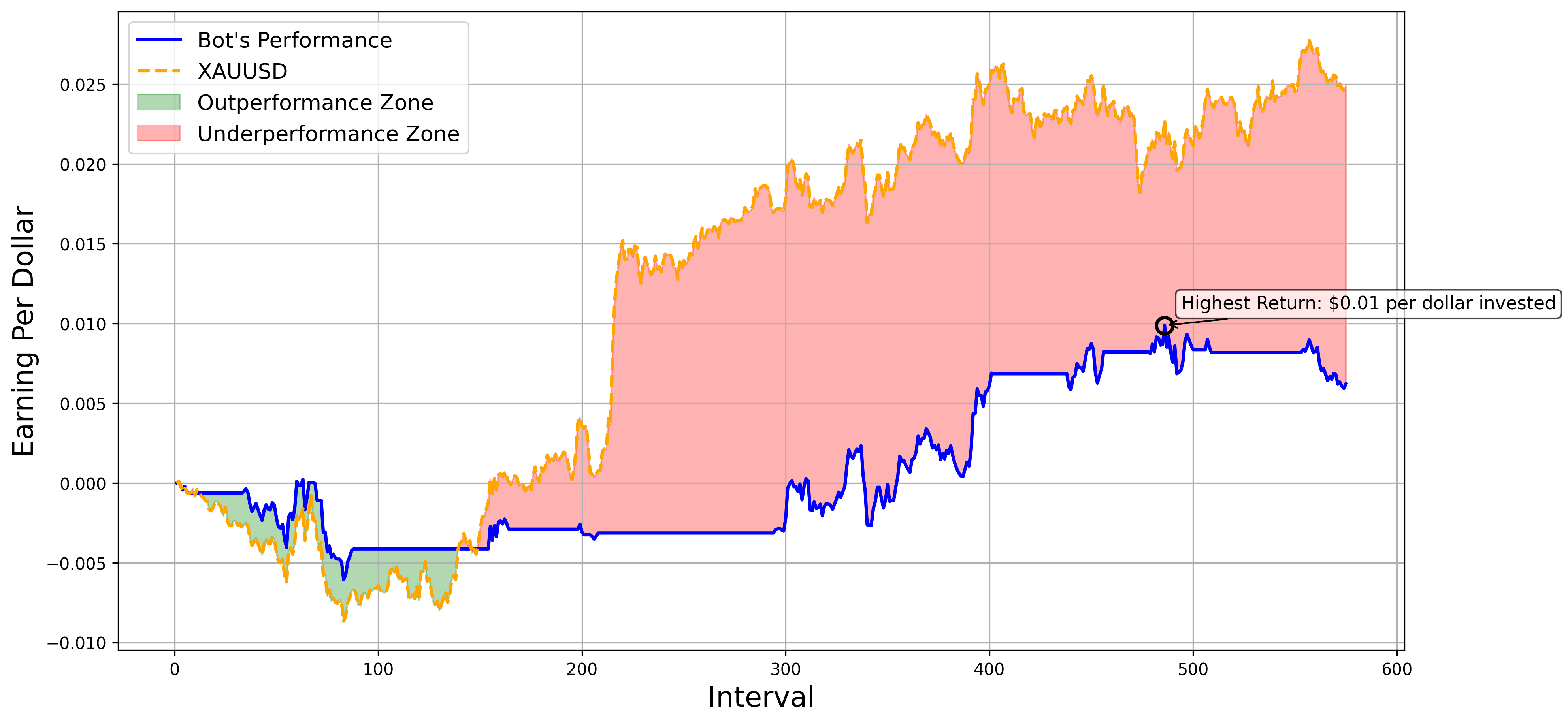}
        \caption{Portfolio}
    \end{subfigure}
    \caption{Chunk 4: Drawdown and Portfolio}
    \label{fig:chunk4}
\end{figure}

\begin{figure}[H]
    \centering
    \begin{subfigure}[b]{0.48\textwidth}
        \includegraphics[width=\linewidth]{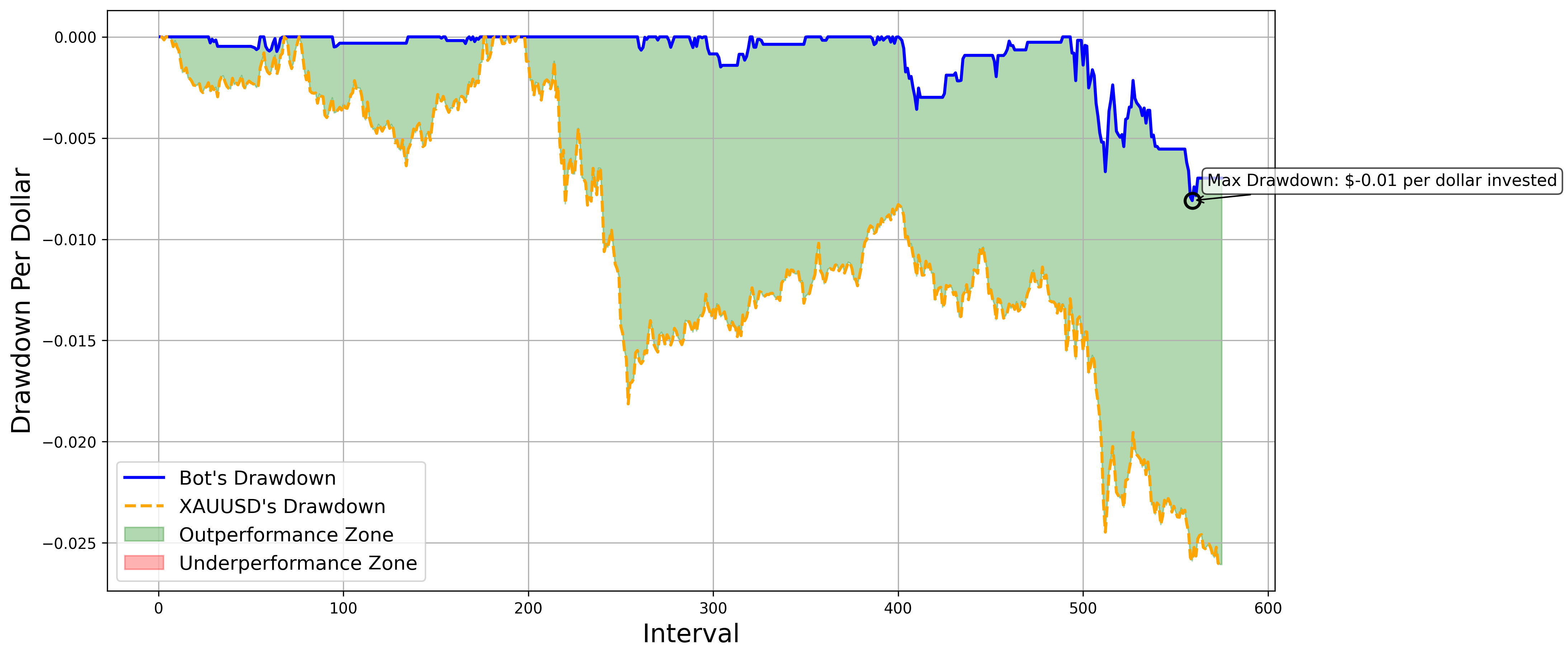}
        \caption{Drawdown}
    \end{subfigure}
    \hfill
    \begin{subfigure}[b]{0.48\textwidth}
        \includegraphics[width=\linewidth]{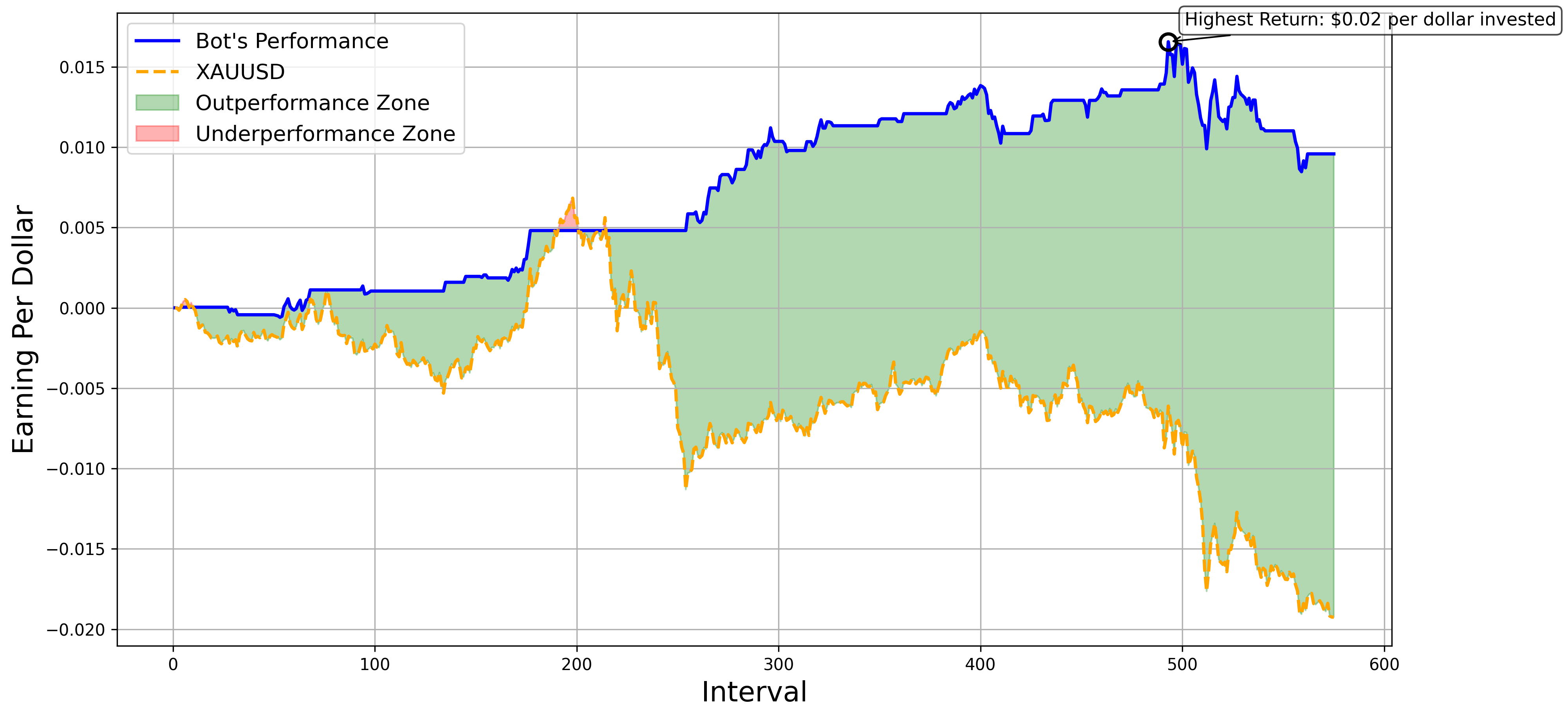}
        \caption{Portfolio}
    \end{subfigure}
    \caption{Chunk 5: Drawdown and Portfolio}
    \label{fig:chunk5}
\end{figure}

\begin{figure}[H]
    \centering
    \begin{subfigure}[b]{0.48\textwidth}
        \includegraphics[width=\linewidth]{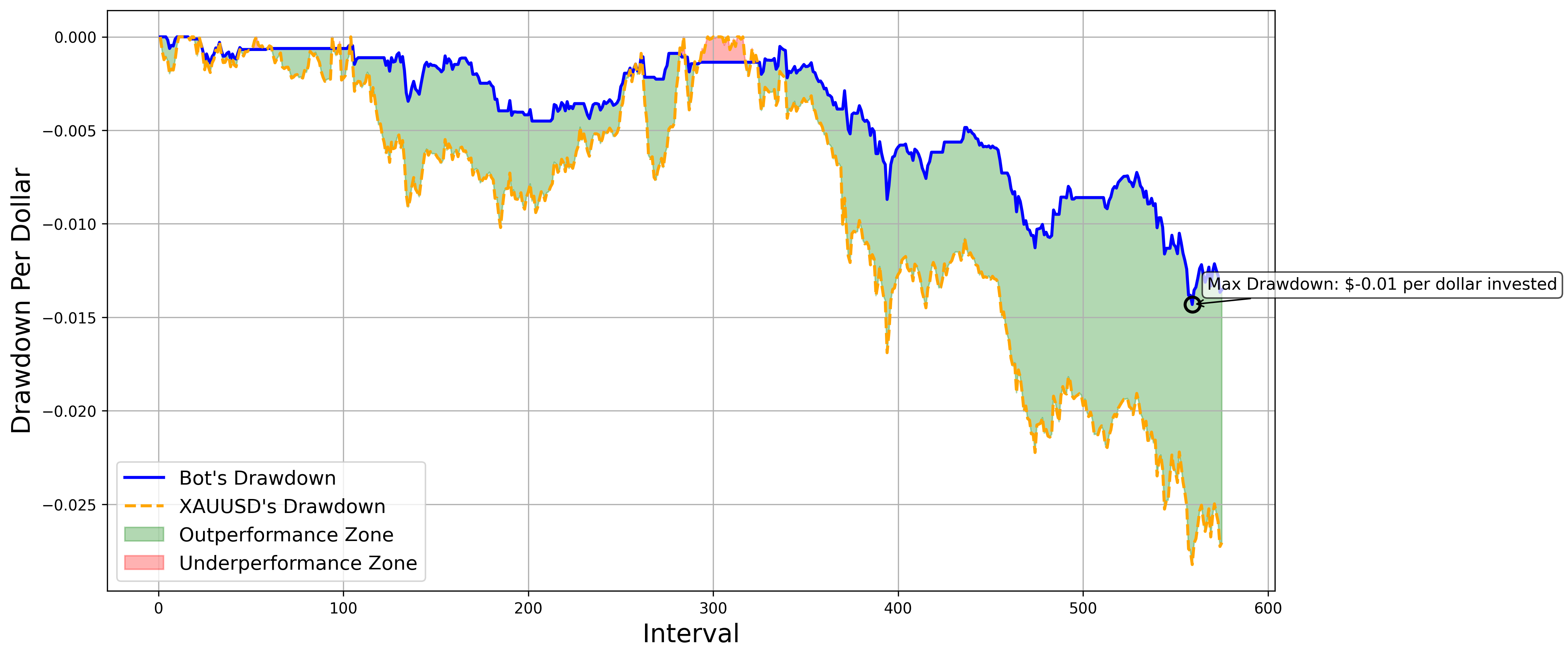}
        \caption{Drawdown}
    \end{subfigure}
    \hfill
    \begin{subfigure}[b]{0.48\textwidth}
        \includegraphics[width=\linewidth]{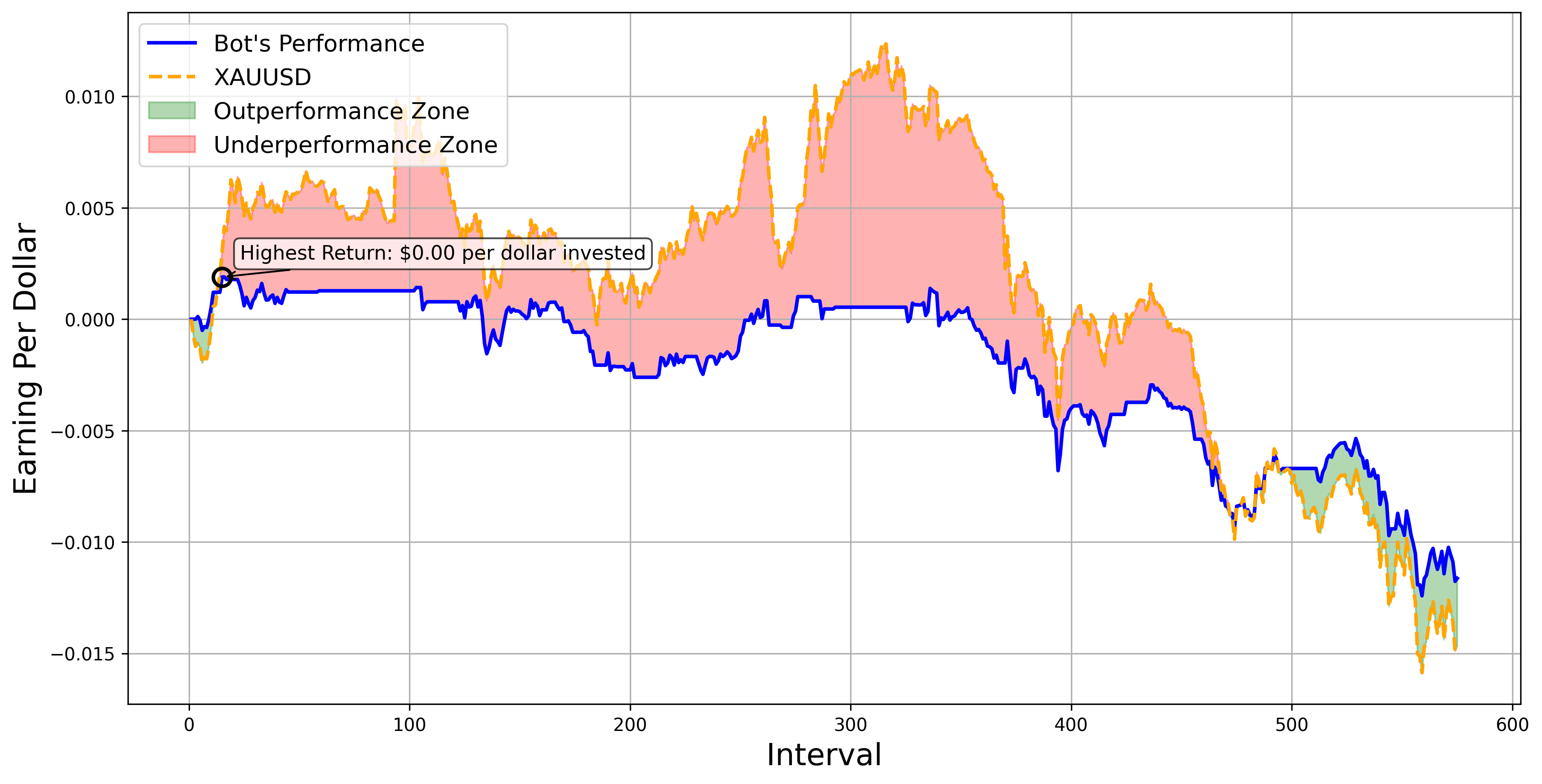}
        \caption{Portfolio}
    \end{subfigure}
    \caption{Chunk 6: Drawdown and Portfolio}
    \label{fig:chunk6}
\end{figure}

\begin{figure}[H]
    \centering
    \begin{subfigure}[b]{0.48\textwidth}
        \includegraphics[width=\linewidth]{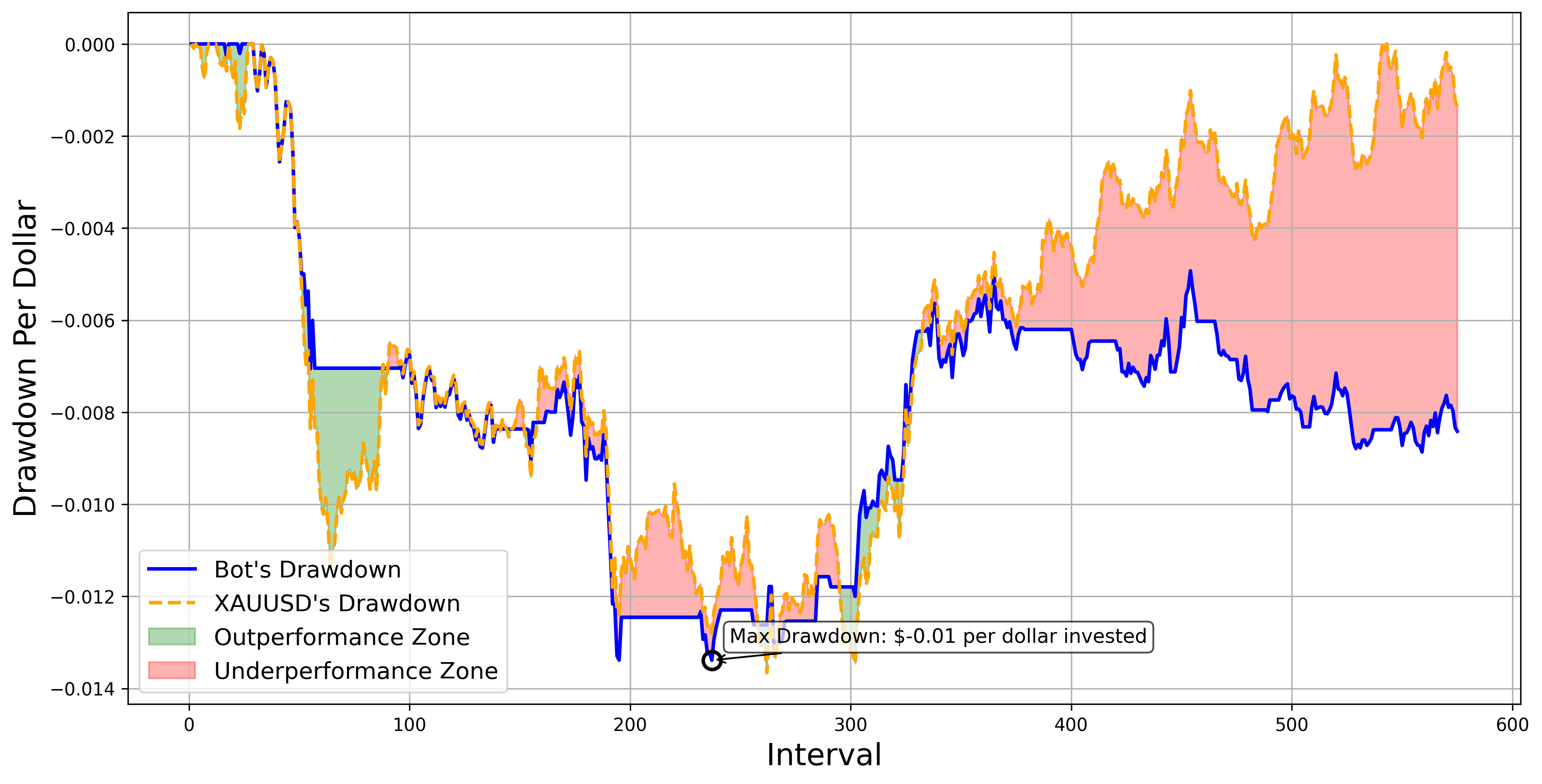}
        \caption{Drawdown}
    \end{subfigure}
    \hfill
    \begin{subfigure}[b]{0.48\textwidth}
        \includegraphics[width=\linewidth]{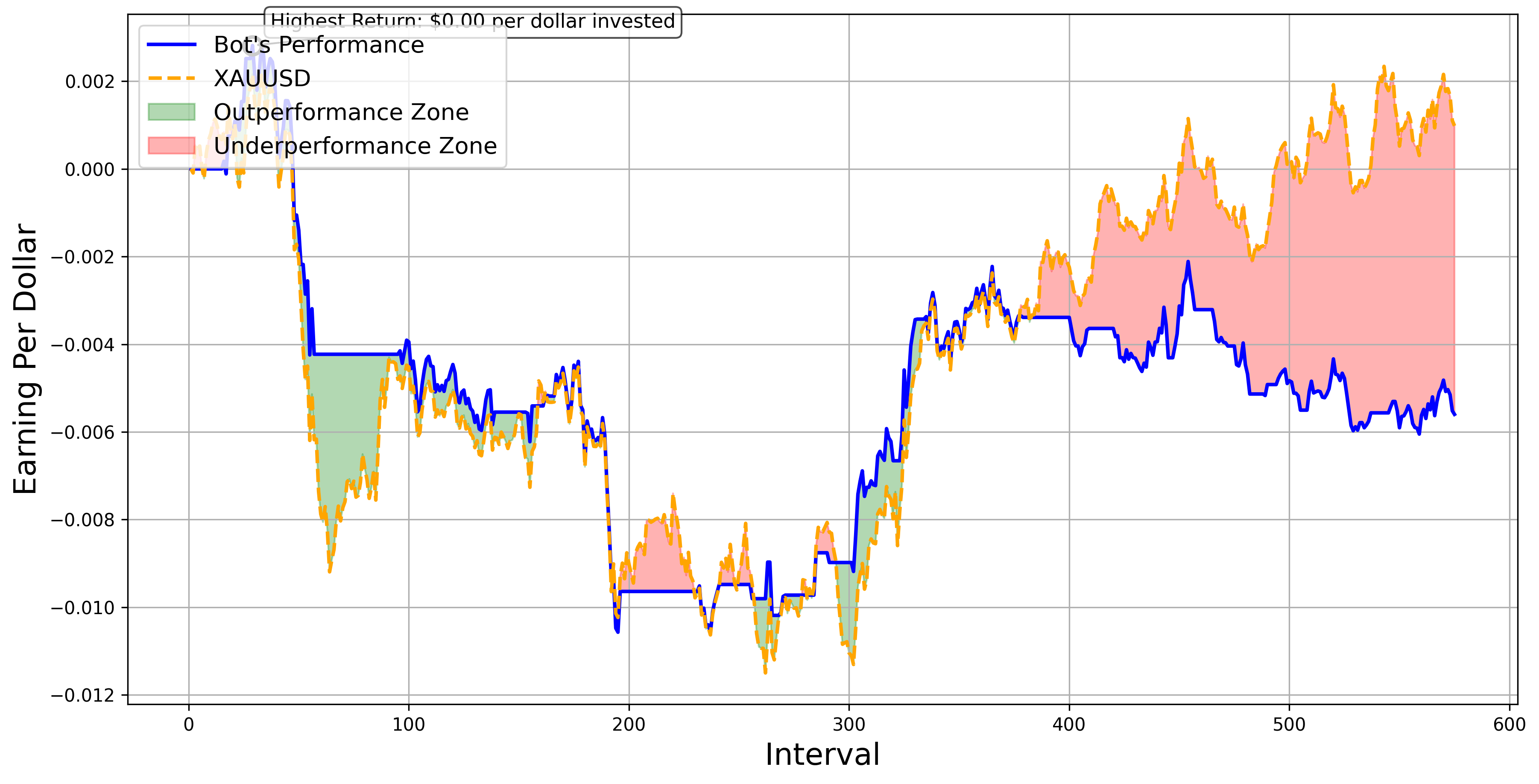}
        \caption{Portfolio}
    \end{subfigure}
    \caption{Chunk 7: Drawdown and Portfolio}
    \label{fig:chunk7}
\end{figure}

\begin{figure}[H]
    \centering
    \begin{subfigure}[b]{0.48\textwidth}
        \includegraphics[width=\linewidth]{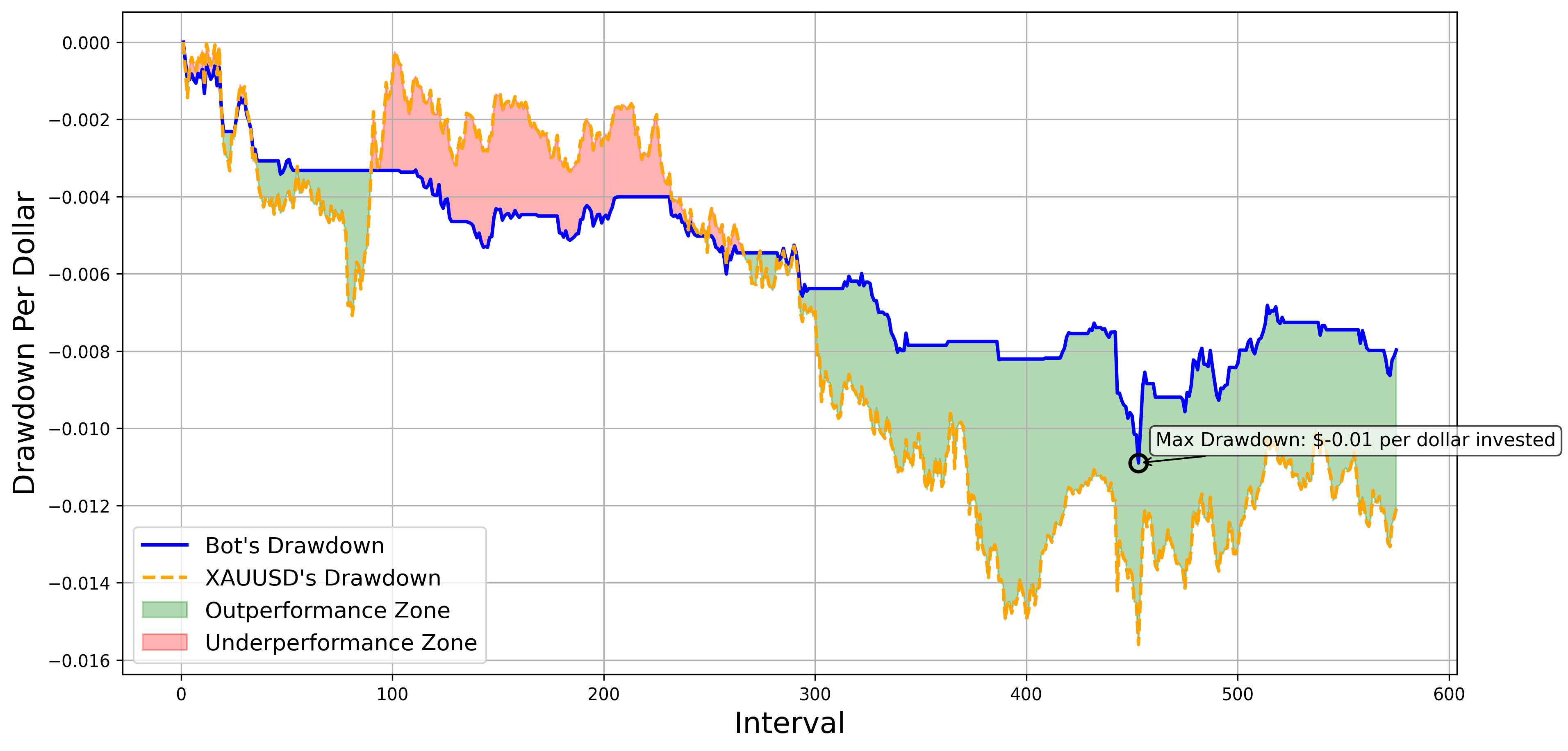}
        \caption{Drawdown}
    \end{subfigure}
    \hfill
    \begin{subfigure}[b]{0.48\textwidth}
        \includegraphics[width=\linewidth]{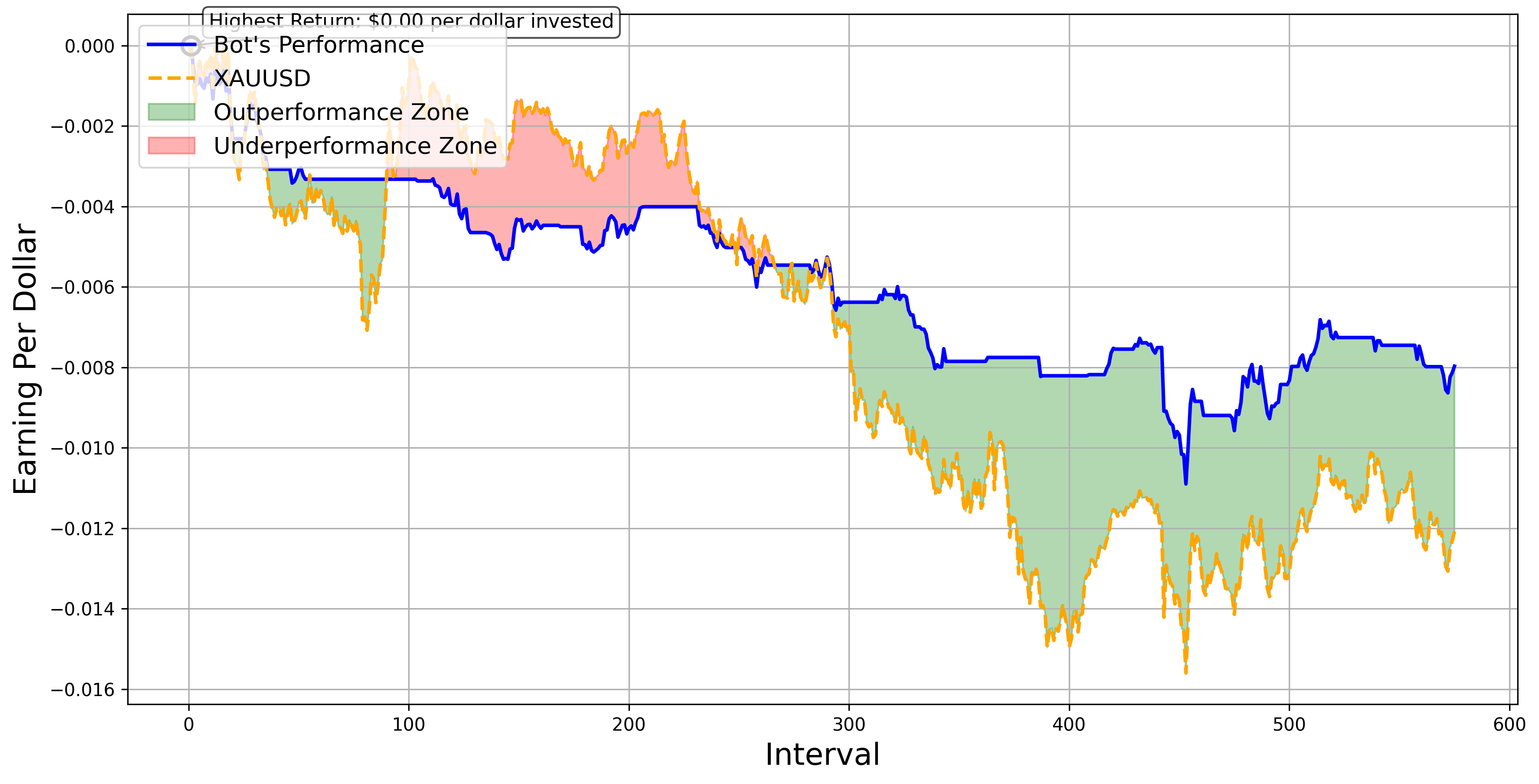}
        \caption{Portfolio}
    \end{subfigure}
    \caption{Chunk 8: Drawdown and Portfolio}
    \label{fig:chunk8}
\end{figure}

\subsection{PDT Profit vs Buy \& Hold Profit (EURUSD Chunks)}

\begin{longtable}{c S[table-format=+1.5] S[table-format=+1.5]}
\caption{PDT Profit vs Buy \& Hold Profit for EURUSD Chunks}\label{tab:eurusd-chunks}\\
\hline
\textbf{Chunk} & \textbf{PDT Profit} & \textbf{Buy \& Hold Profit} \\
\hline
\endfirsthead
\hline
\textbf{Chunk} & \textbf{PDT Profit} & \textbf{Buy \& Hold Profit} \\
\hline
\endhead
1 &  0.00806 &  0.00276 \\
2 &  0.00049 & -0.00229 \\
3 & -0.00183 & -0.00410 \\
4 &  0.00519 &  0.00333 \\
5 &  0.00034 &  0.00067 \\
6 &  0.00180 &  0.00158 \\
7 &  0.00022 & -0.00545 \\
8 &  0.00724 &  0.01032 \\
9 &  0.00363 &  0.01086 \\
\hline
\end{longtable}
\subsection{Drawdown and Portfolio Charts for EURUSD Chunks}

\begin{figure}[H]
    \centering
    \begin{subfigure}[b]{0.48\textwidth}
        \includegraphics[width=\linewidth]{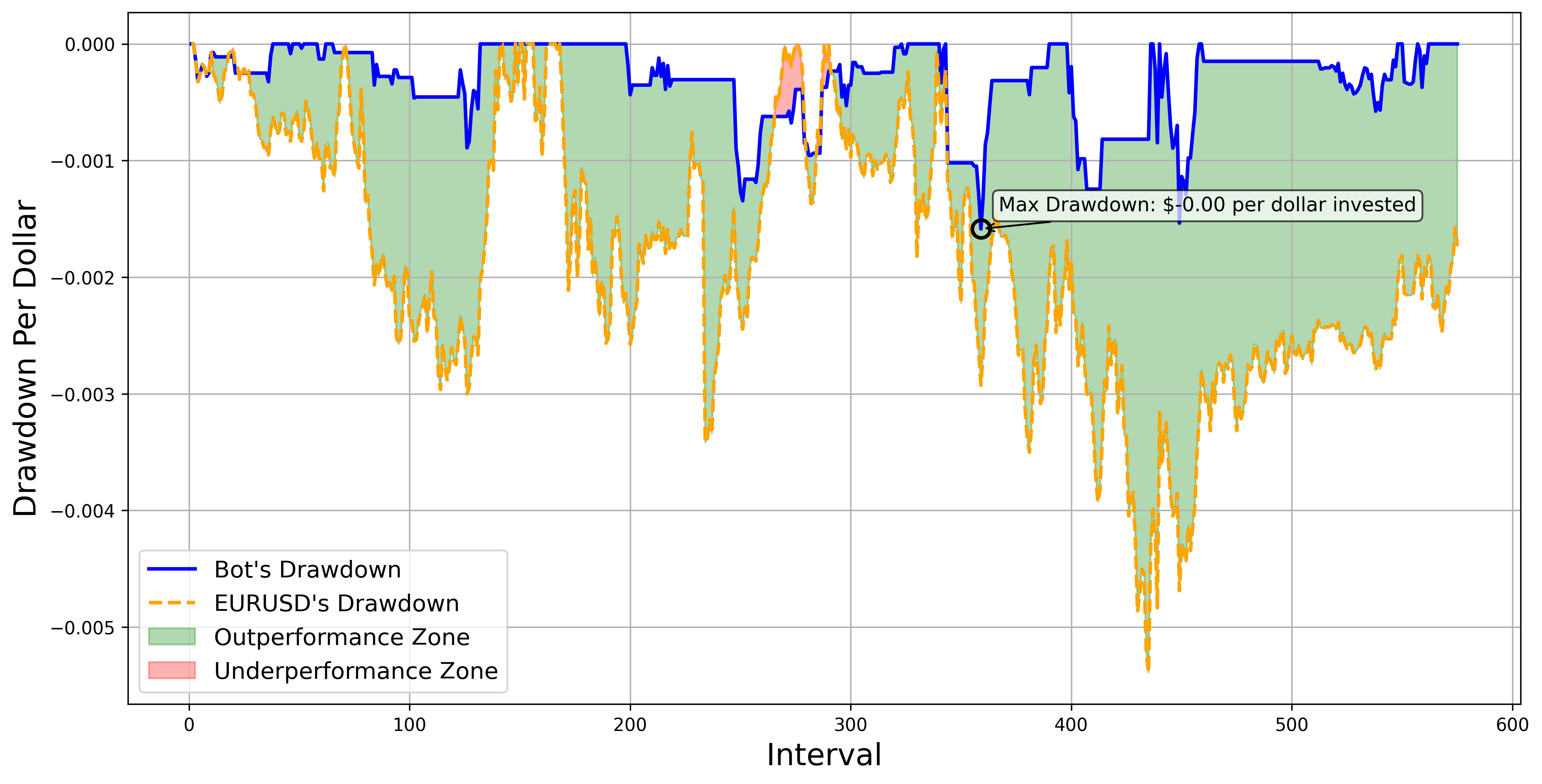}
        \caption{Drawdown}
    \end{subfigure}
    \hfill
    \begin{subfigure}[b]{0.48\textwidth}
        \includegraphics[width=\linewidth]{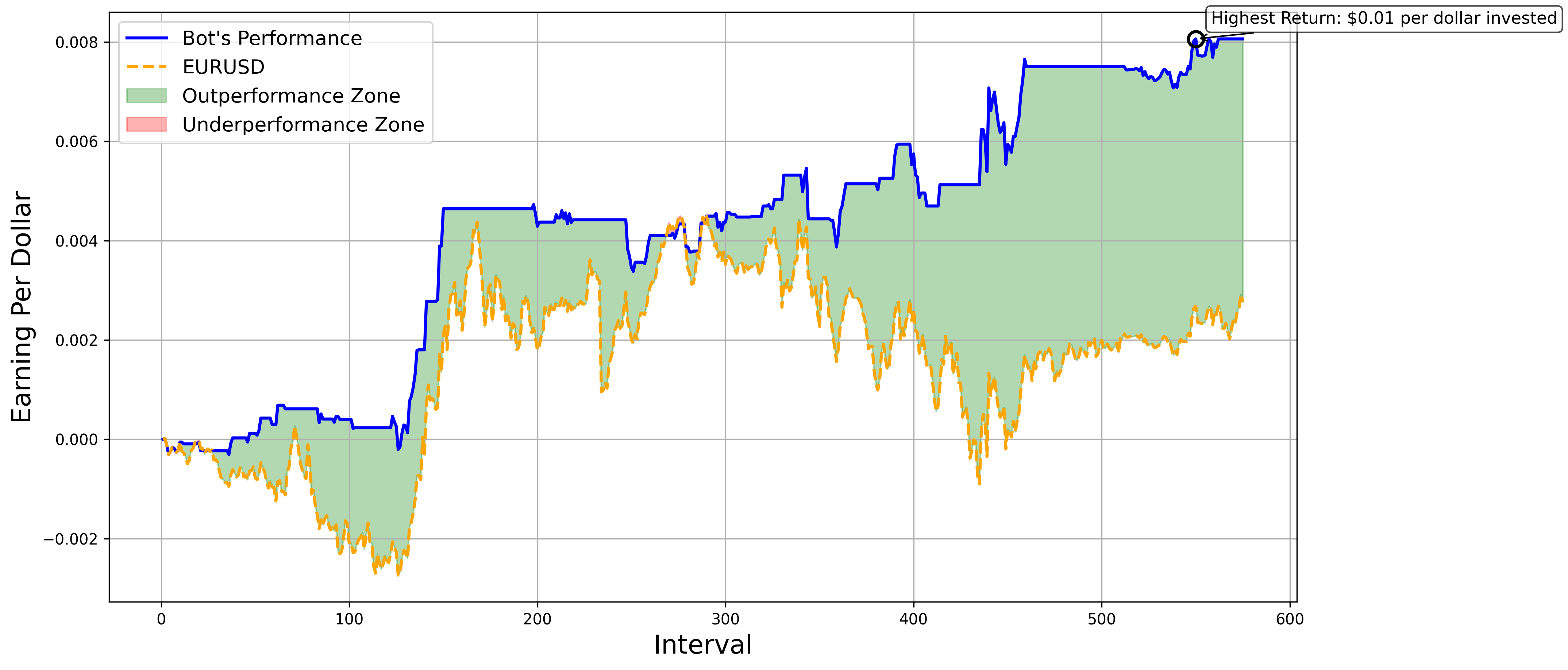}
        \caption{Portfolio}
    \end{subfigure}
    \caption{Chunk 1: Drawdown and Portfolio}
    \label{fig:eurusd-chunk1}
\end{figure}

\begin{figure}[H]
    \centering
    \begin{subfigure}[b]{0.48\textwidth}
        \includegraphics[width=\linewidth]{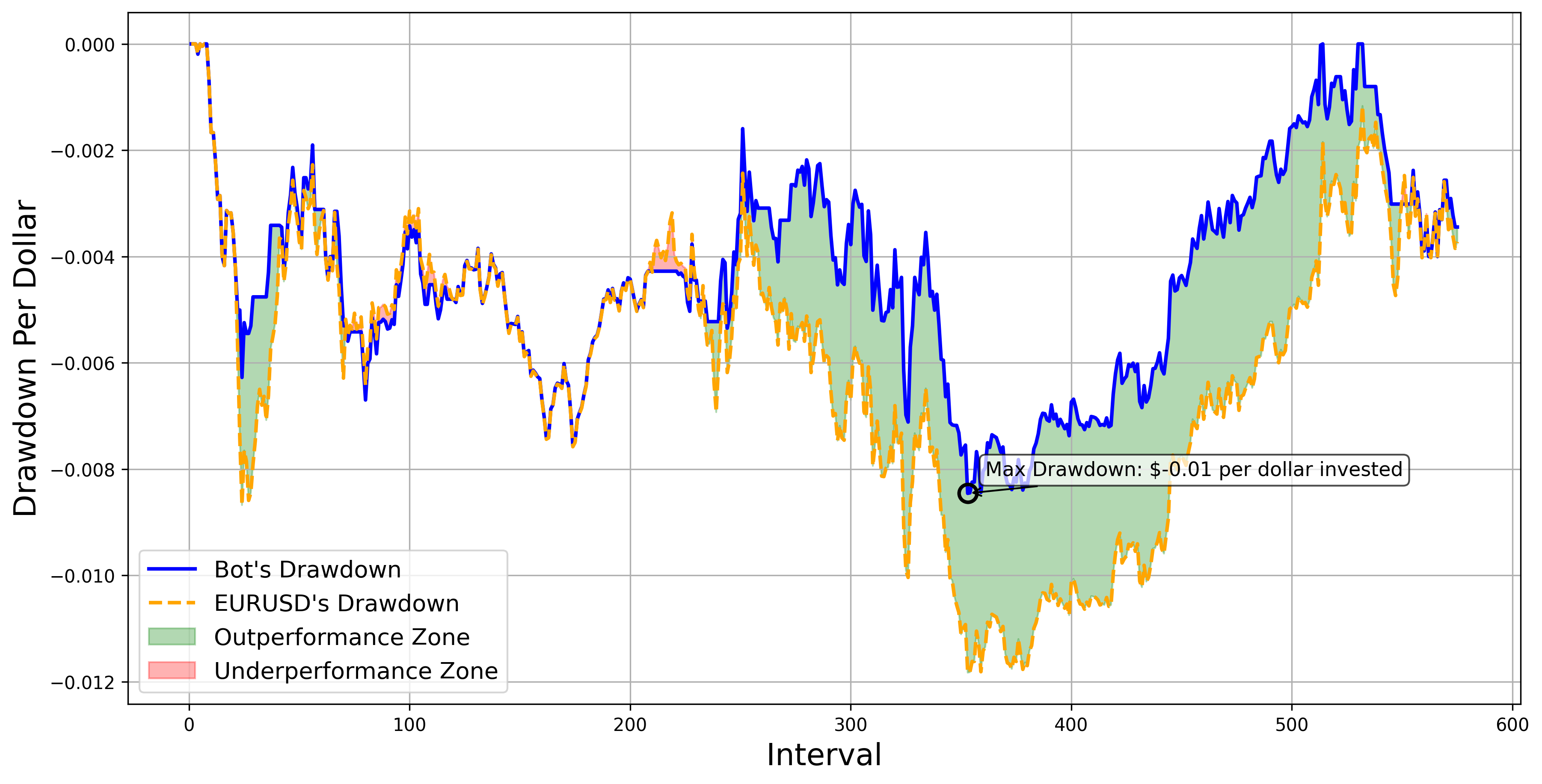}
        \caption{Drawdown}
    \end{subfigure}
    \hfill
    \begin{subfigure}[b]{0.48\textwidth}
        \includegraphics[width=\linewidth]{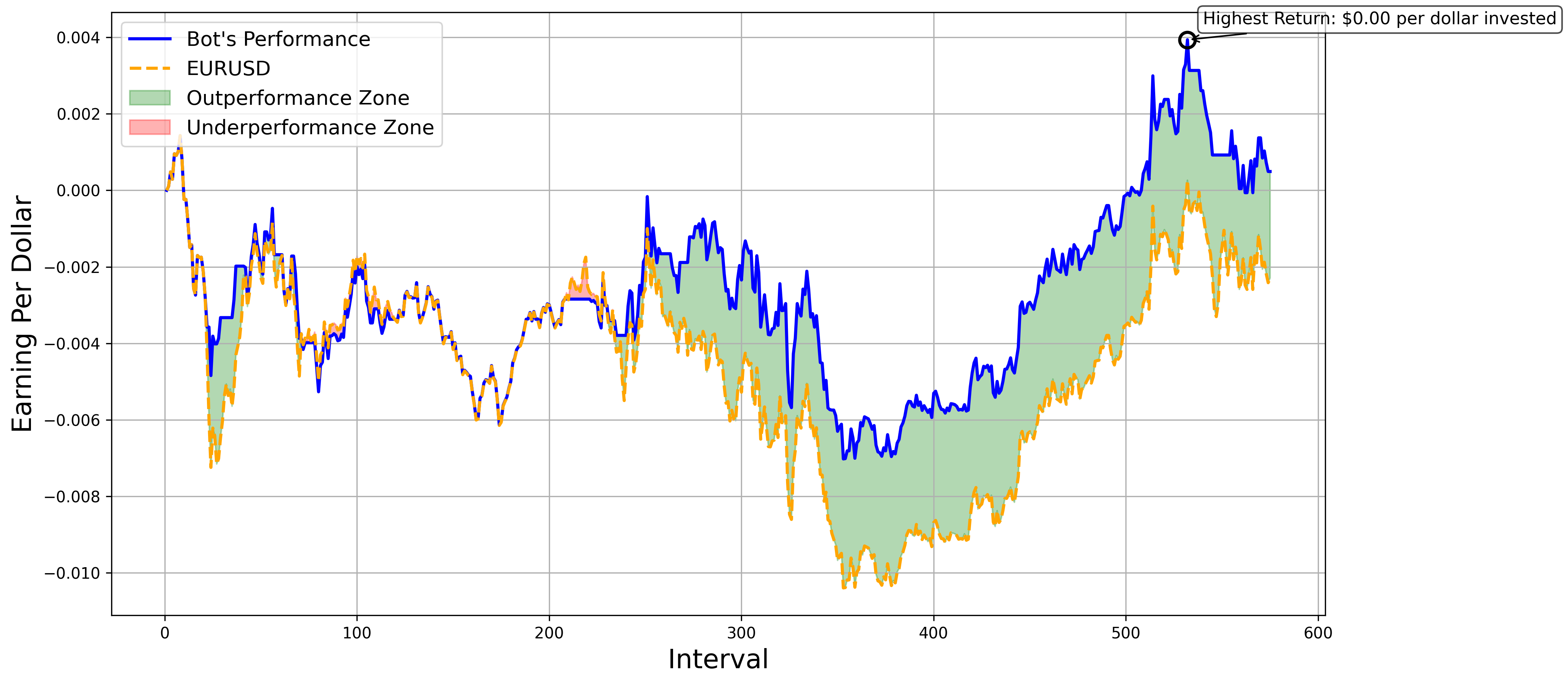}
        \caption{Portfolio}
    \end{subfigure}
    \caption{Chunk 2: Drawdown and Portfolio}
    \label{fig:eurusd-chunk2}
\end{figure}

\begin{figure}[H]
    \centering
    \begin{subfigure}[b]{0.48\textwidth}
        \includegraphics[width=\linewidth]{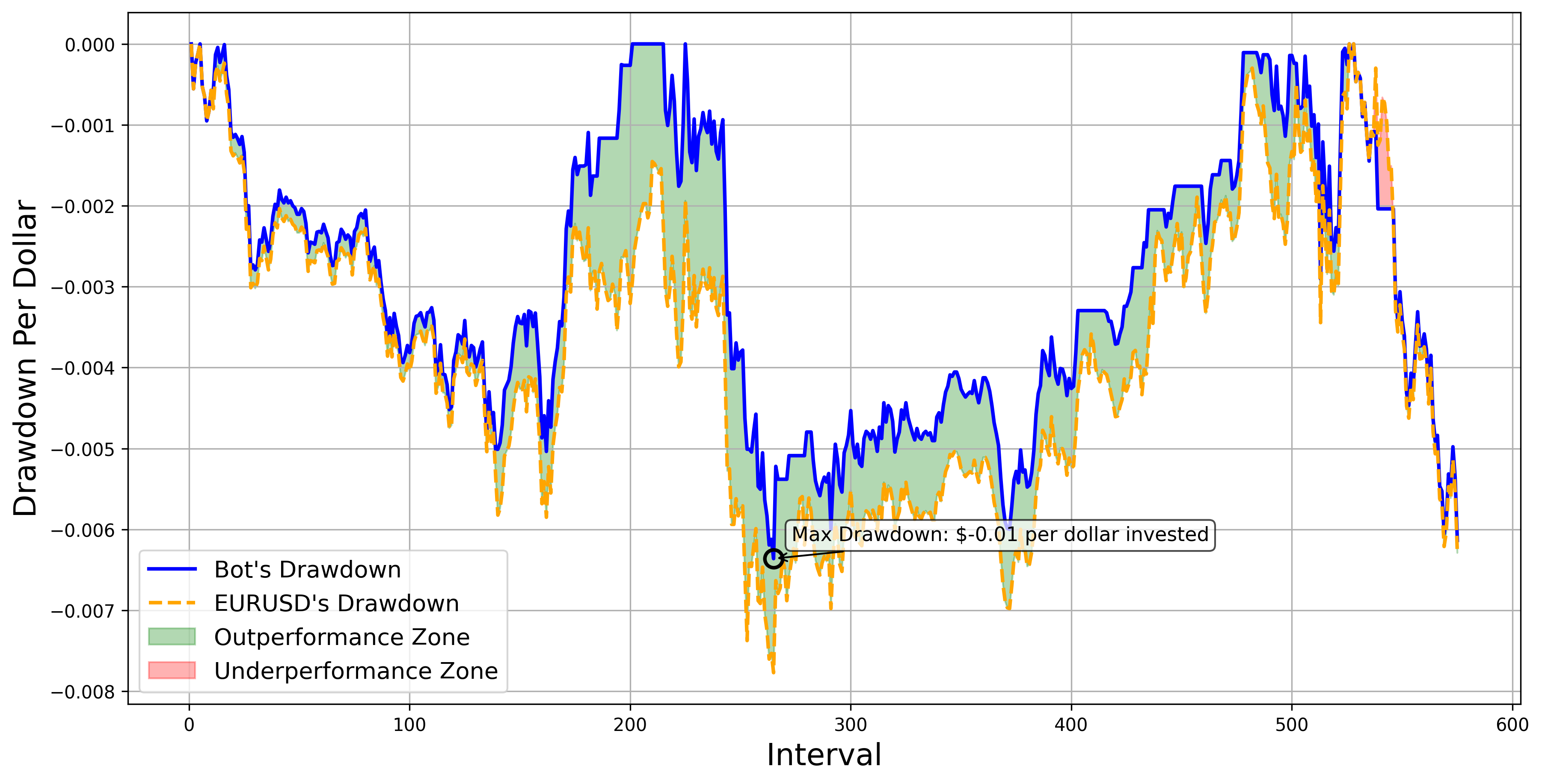}
        \caption{Drawdown}
    \end{subfigure}
    \hfill
    \begin{subfigure}[b]{0.48\textwidth}
        \includegraphics[width=\linewidth]{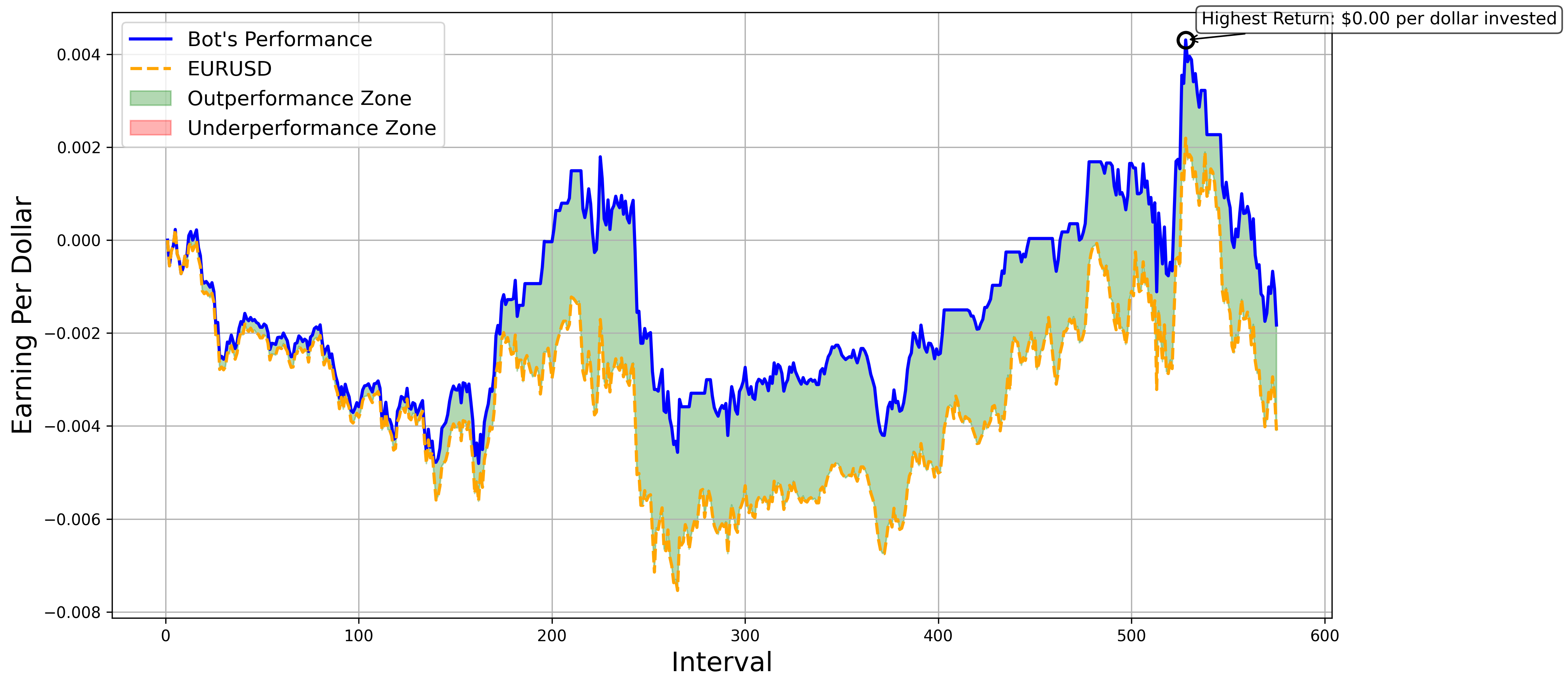}
        \caption{Portfolio}
    \end{subfigure}
    \caption{Chunk 3: Drawdown and Portfolio}
    \label{fig:eurusd-chunk3}
\end{figure}

\begin{figure}[H]
    \centering
    \begin{subfigure}[b]{0.48\textwidth}
        \includegraphics[width=\linewidth]{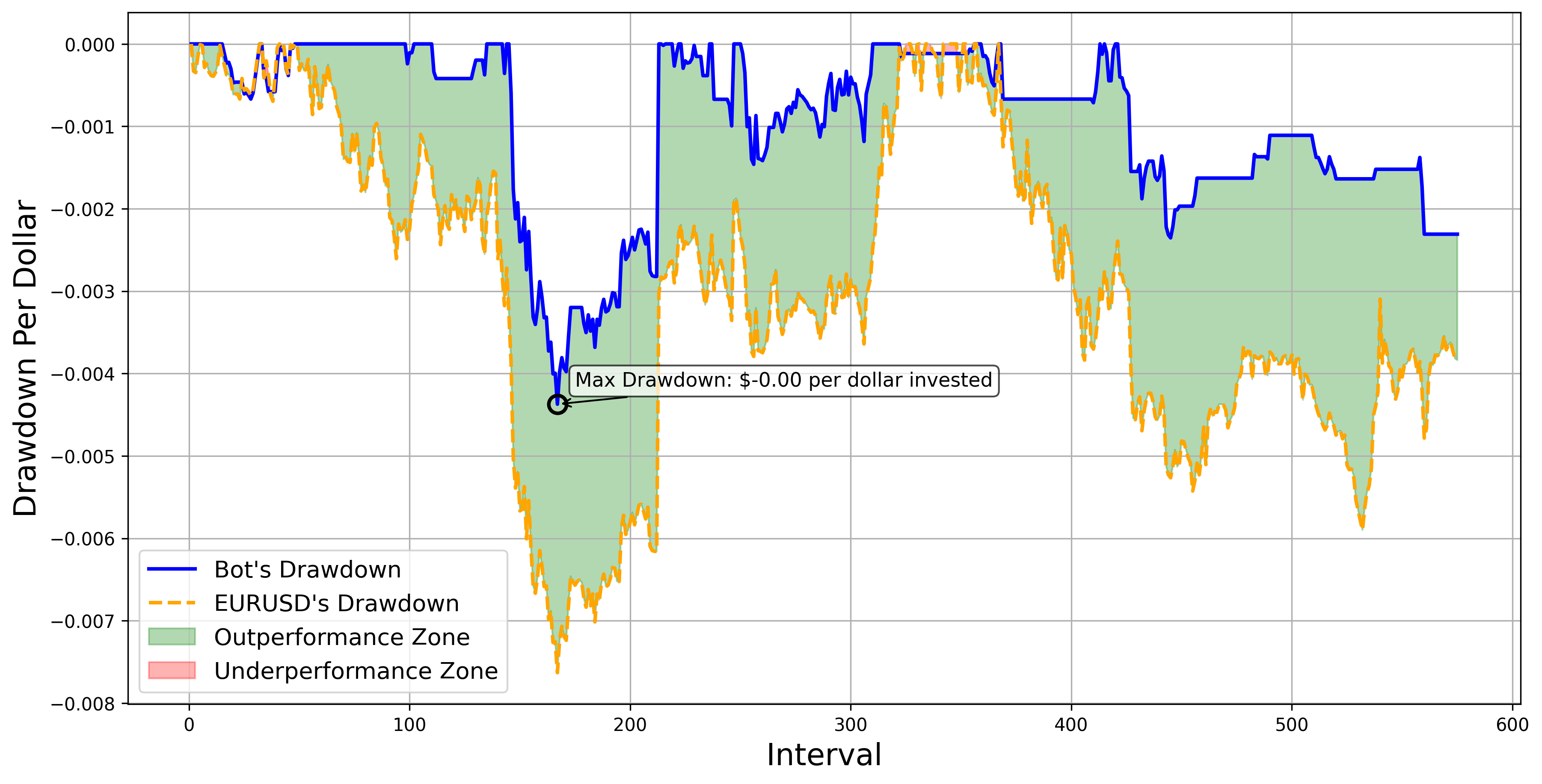}
        \caption{Drawdown}
    \end{subfigure}
    \hfill
    \begin{subfigure}[b]{0.48\textwidth}
        \includegraphics[width=\linewidth]{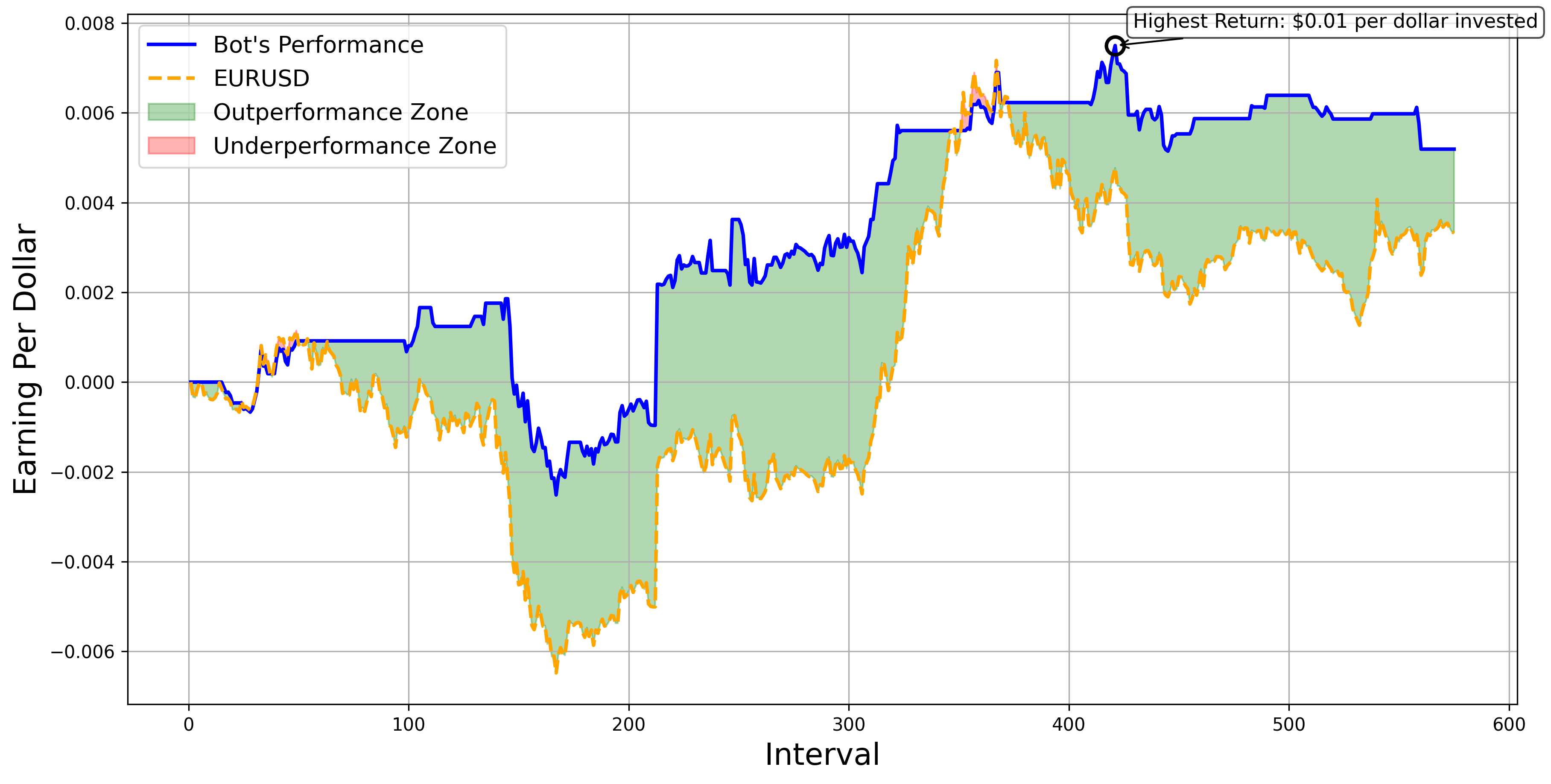}
        \caption{Portfolio}
    \end{subfigure}
    \caption{Chunk 4: Drawdown and Portfolio}
    \label{fig:eurusd-chunk4}
\end{figure}

\begin{figure}[H]
    \centering
    \begin{subfigure}[b]{0.48\textwidth}
        \includegraphics[width=\linewidth]{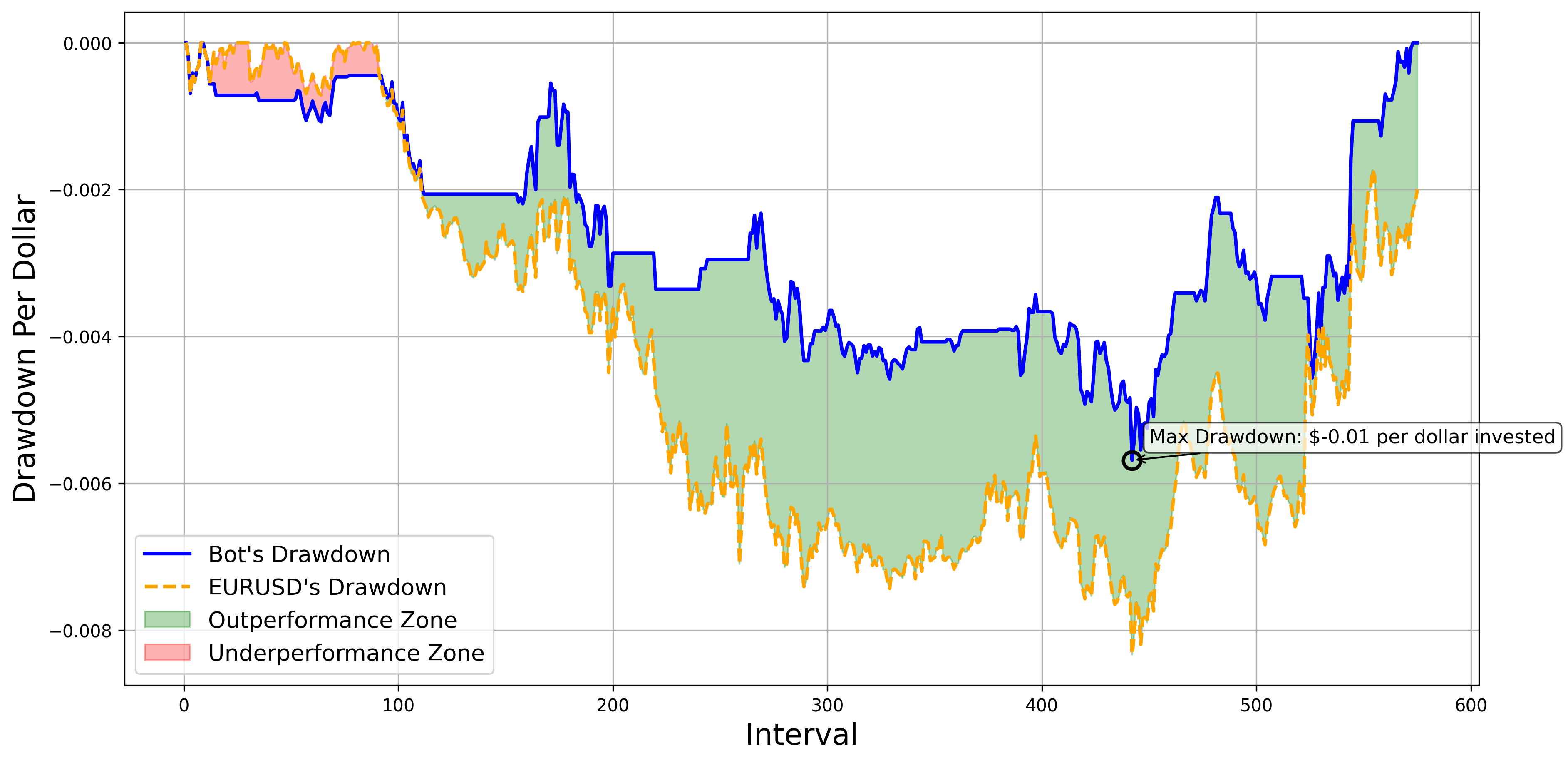}
        \caption{Drawdown}
    \end{subfigure}
    \hfill
    \begin{subfigure}[b]{0.48\textwidth}
        \includegraphics[width=\linewidth]{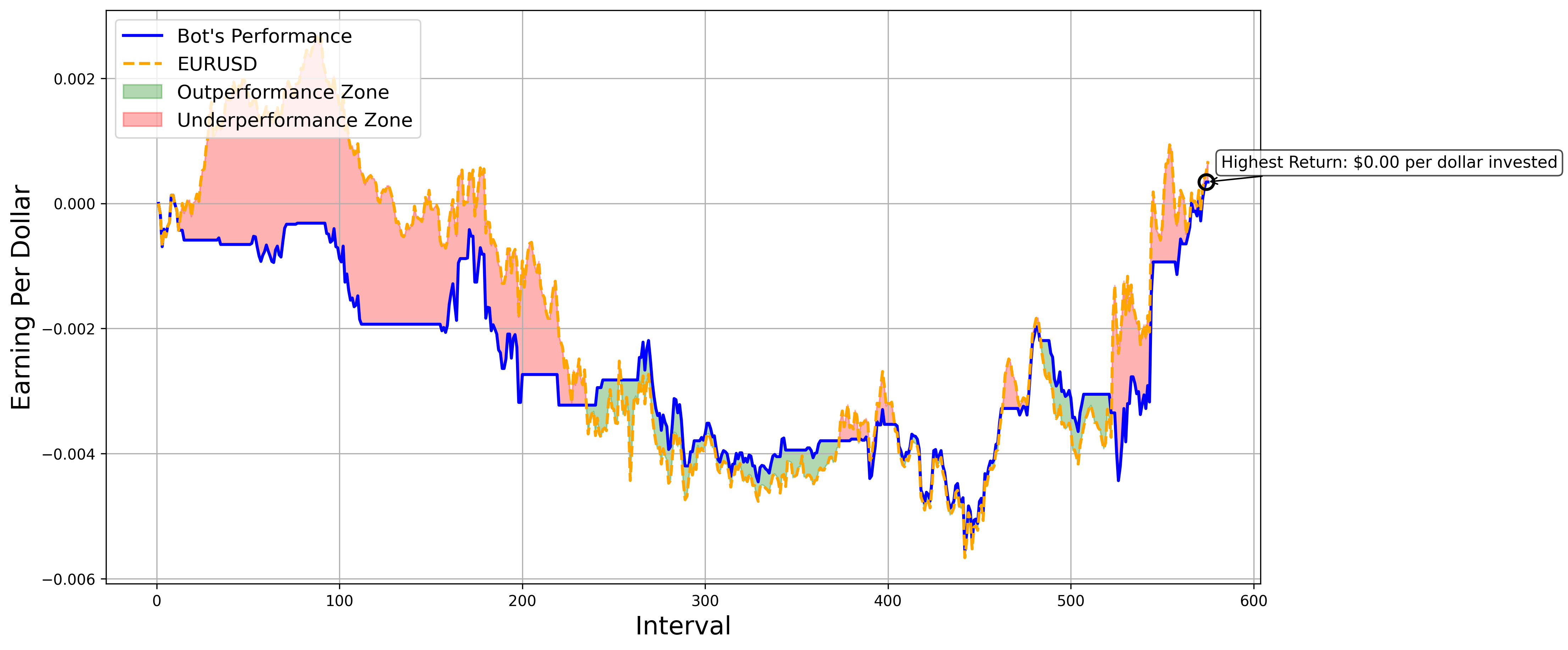}
        \caption{Portfolio}
    \end{subfigure}
    \caption{Chunk 5: Drawdown and Portfolio}
    \label{fig:eurusd-chunk5}
\end{figure}

\begin{figure}[H]
    \centering
    \begin{subfigure}[b]{0.48\textwidth}
        \includegraphics[width=\linewidth]{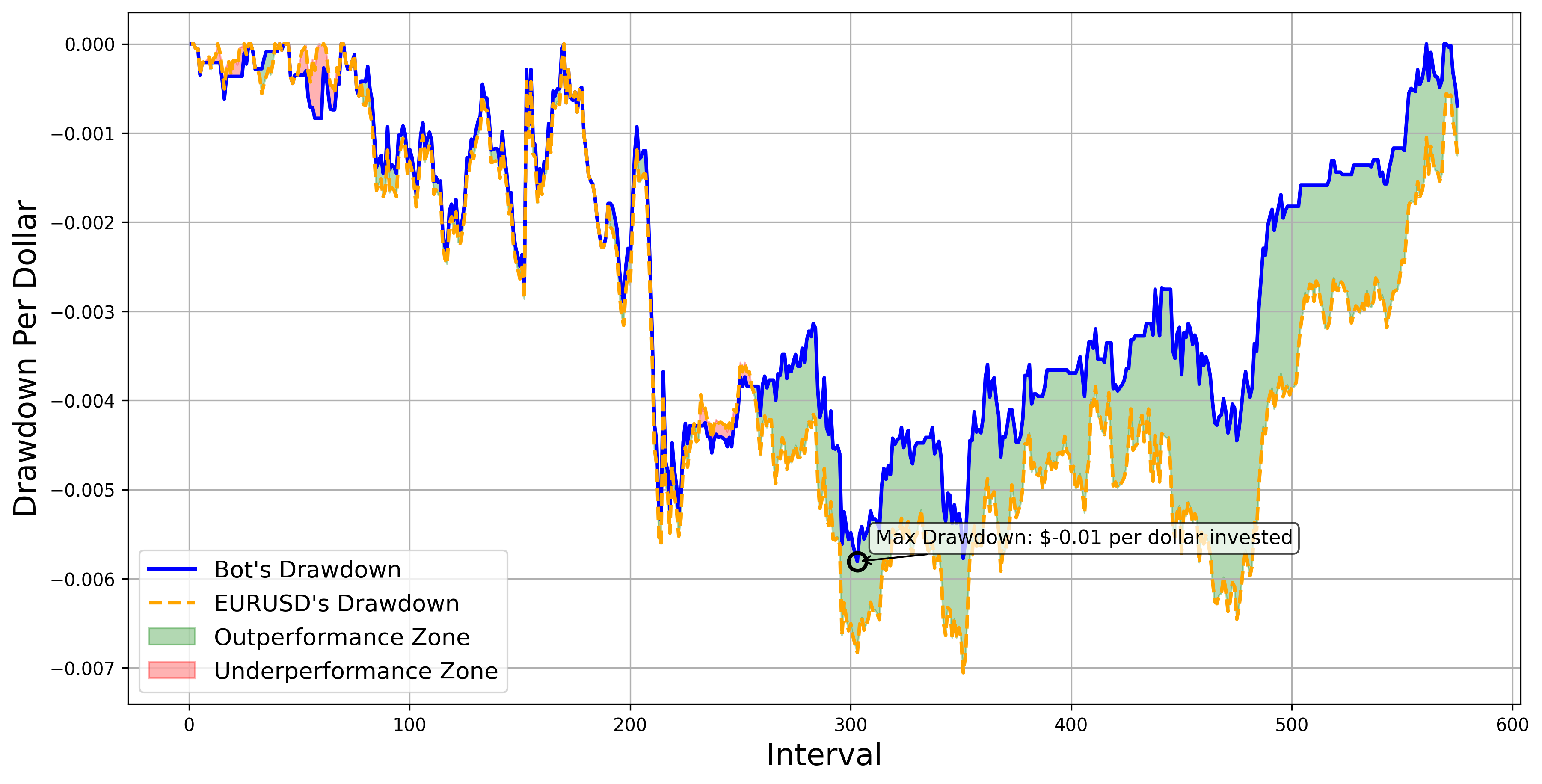}
        \caption{Drawdown}
    \end{subfigure}
    \hfill
    \begin{subfigure}[b]{0.48\textwidth}
        \includegraphics[width=\linewidth]{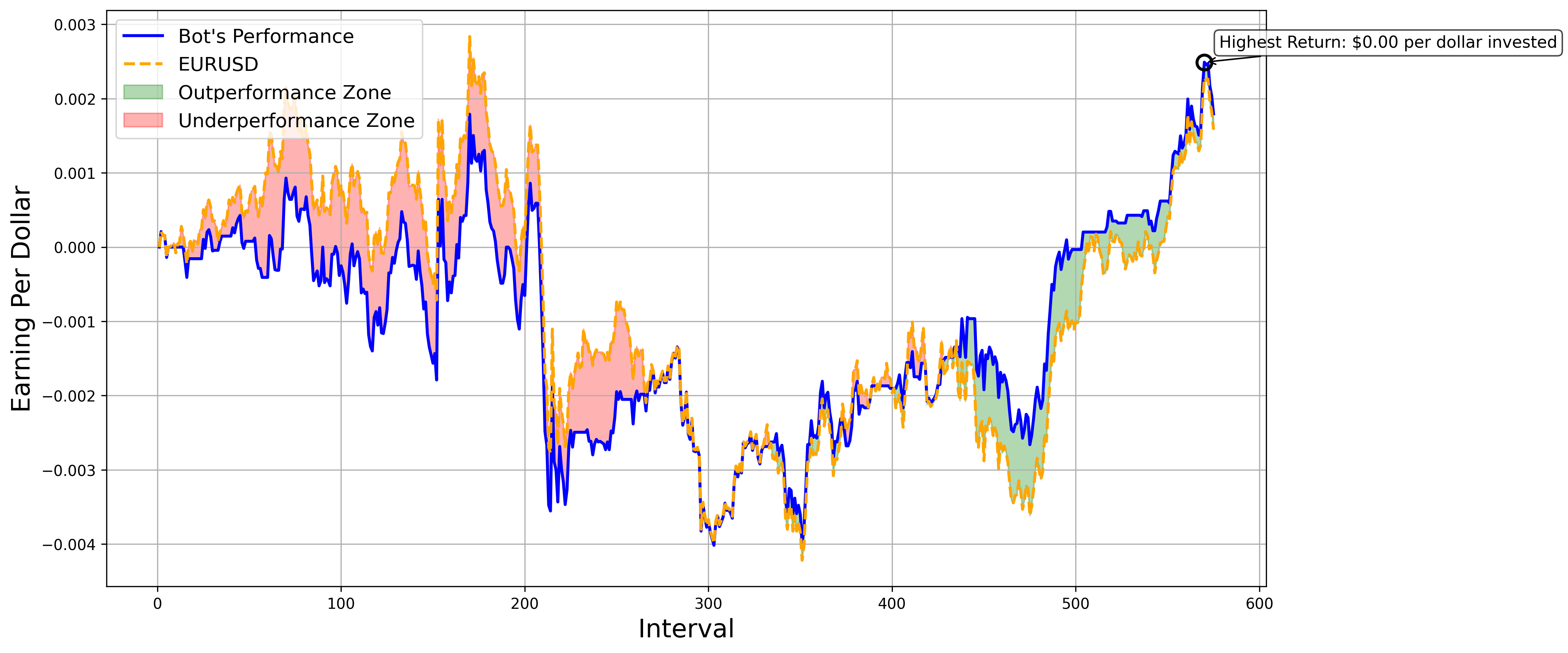}
        \caption{Portfolio}
    \end{subfigure}
    \caption{Chunk 6: Drawdown and Portfolio}
    \label{fig:eurusd-chunk6}
\end{figure}

\begin{figure}[H]
    \centering
    \begin{subfigure}[b]{0.48\textwidth}
        \includegraphics[width=\linewidth]{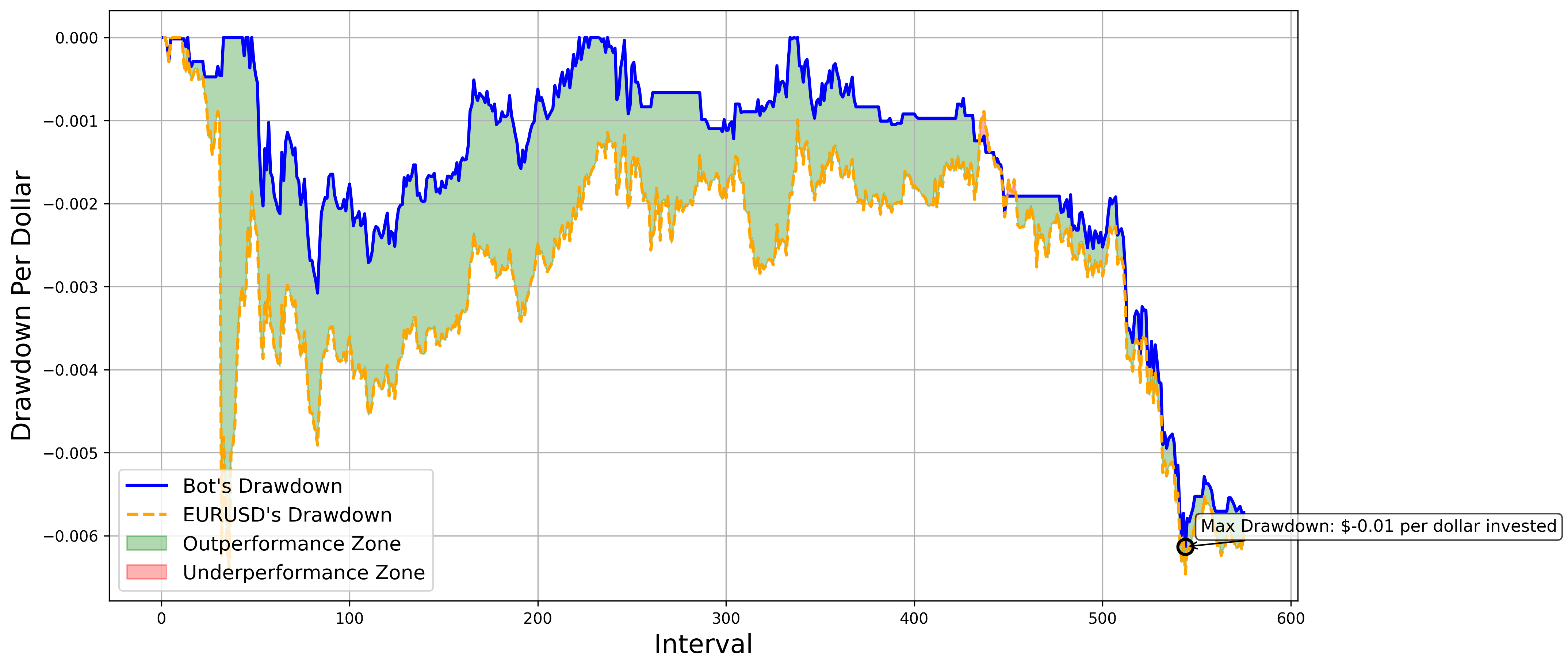}
        \caption{Drawdown}
    \end{subfigure}
    \hfill
    \begin{subfigure}[b]{0.48\textwidth}
        \includegraphics[width=\linewidth]{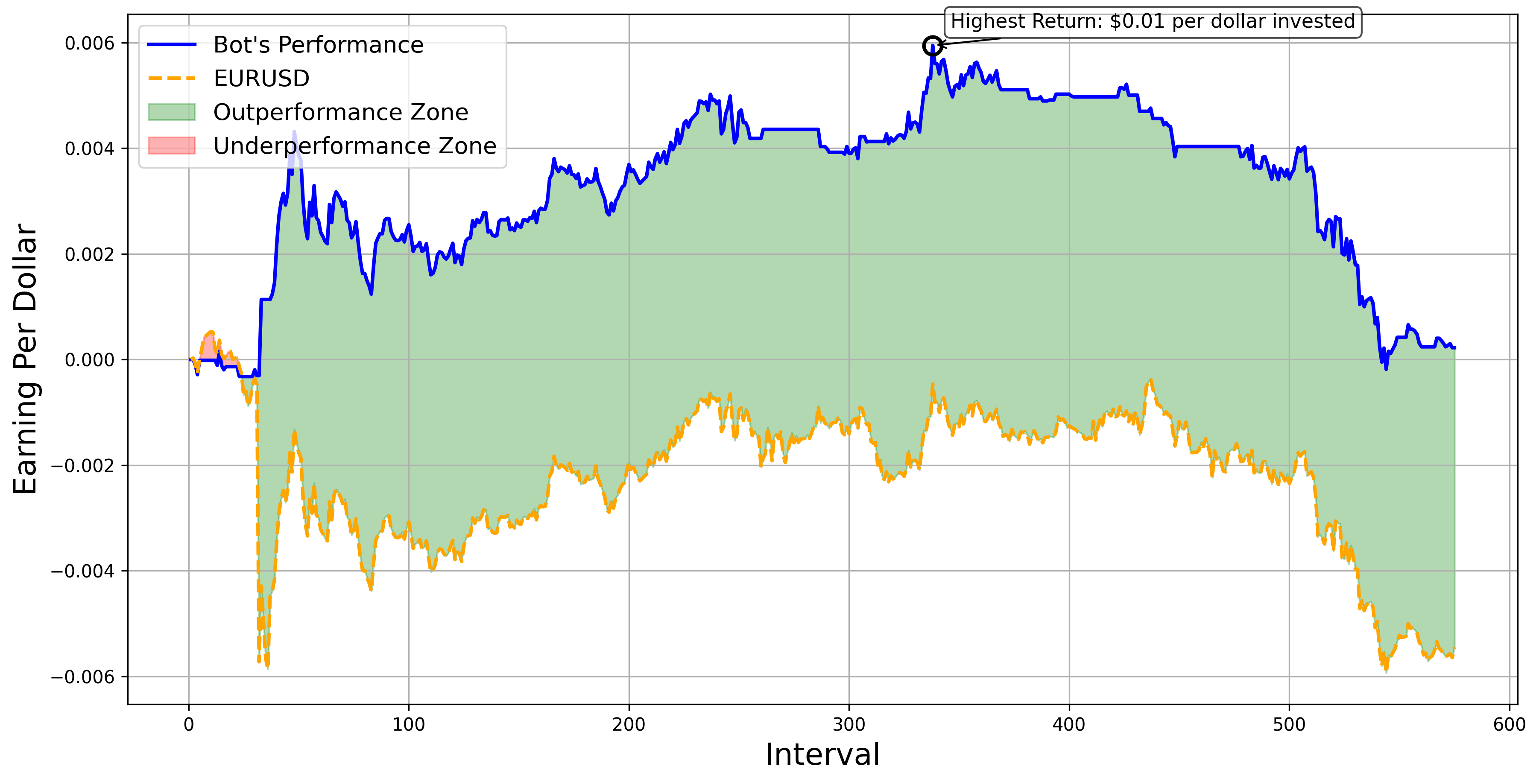}
        \caption{Portfolio}
    \end{subfigure}
    \caption{Chunk 7: Drawdown and Portfolio}
    \label{fig:eurusd-chunk7}
\end{figure}

\begin{figure}[H]
    \centering
    \begin{subfigure}[b]{0.48\textwidth}
        \includegraphics[width=\linewidth]{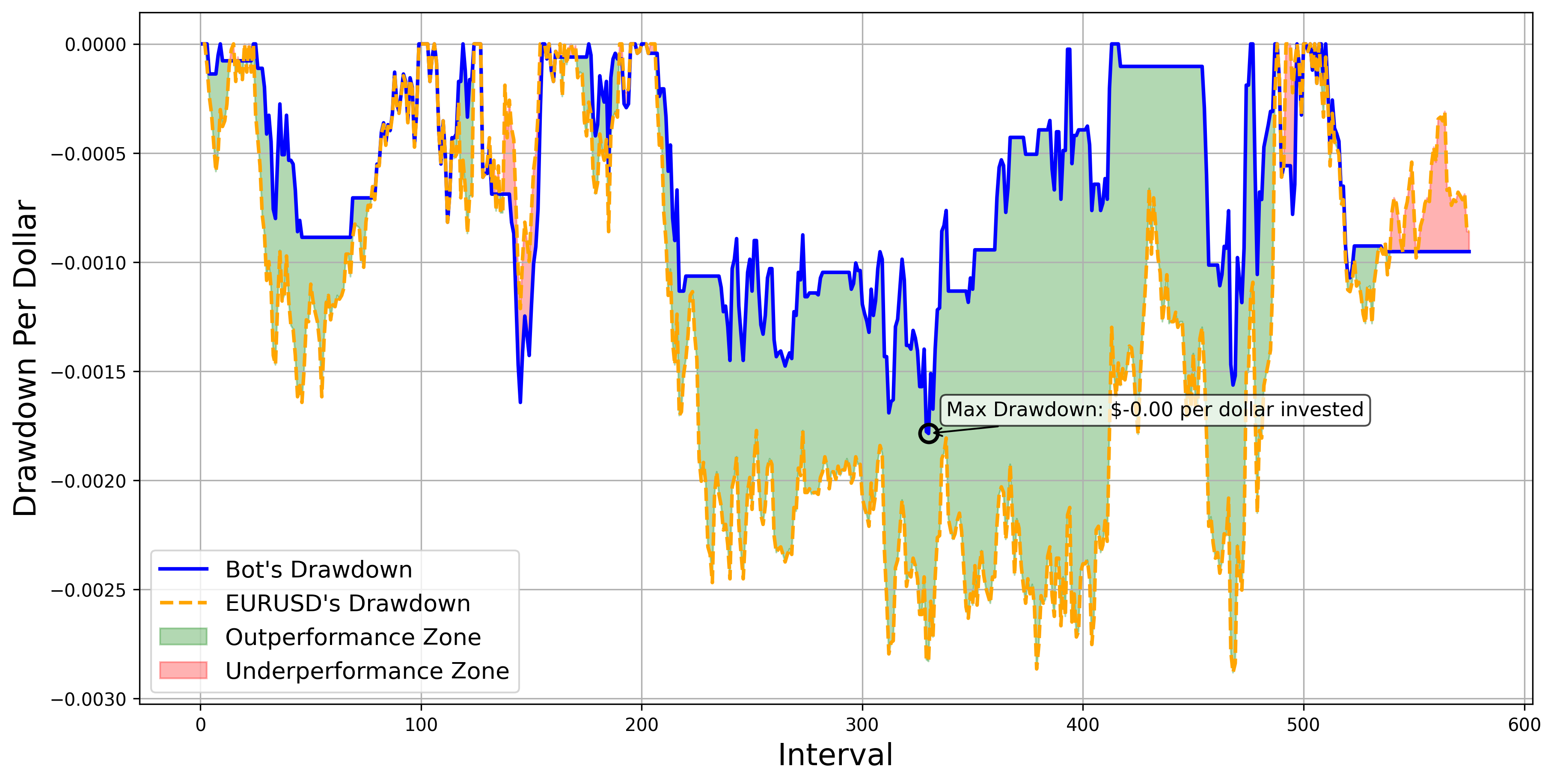}
        \caption{Drawdown}
    \end{subfigure}
    \hfill
    \begin{subfigure}[b]{0.48\textwidth}
        \includegraphics[width=\linewidth]{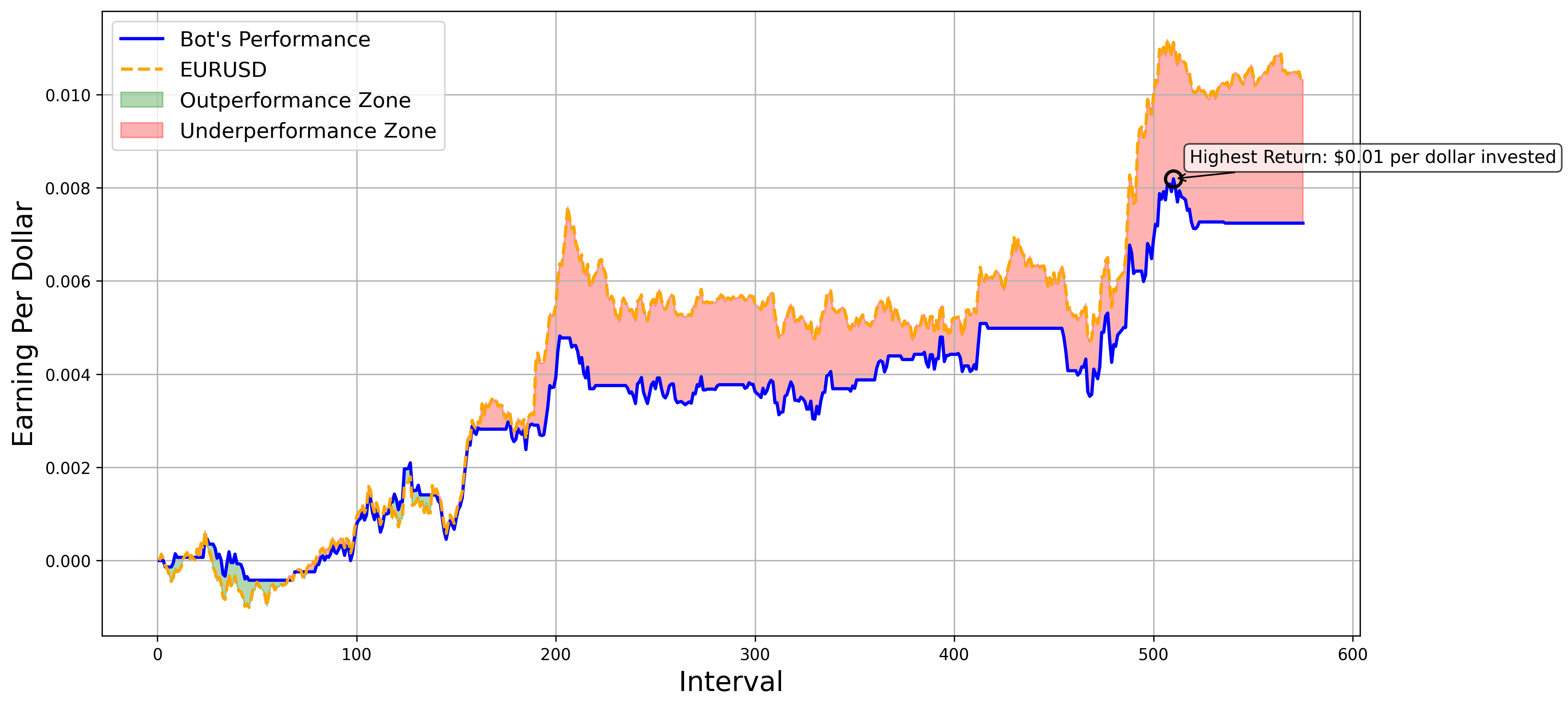}
        \caption{Portfolio}
    \end{subfigure}
    \caption{Chunk 8: Drawdown and Portfolio}
    \label{fig:eurusd-chunk8}
\end{figure}

\begin{figure}[H]
    \centering
    \begin{subfigure}[b]{0.48\textwidth}
        \includegraphics[width=\linewidth]{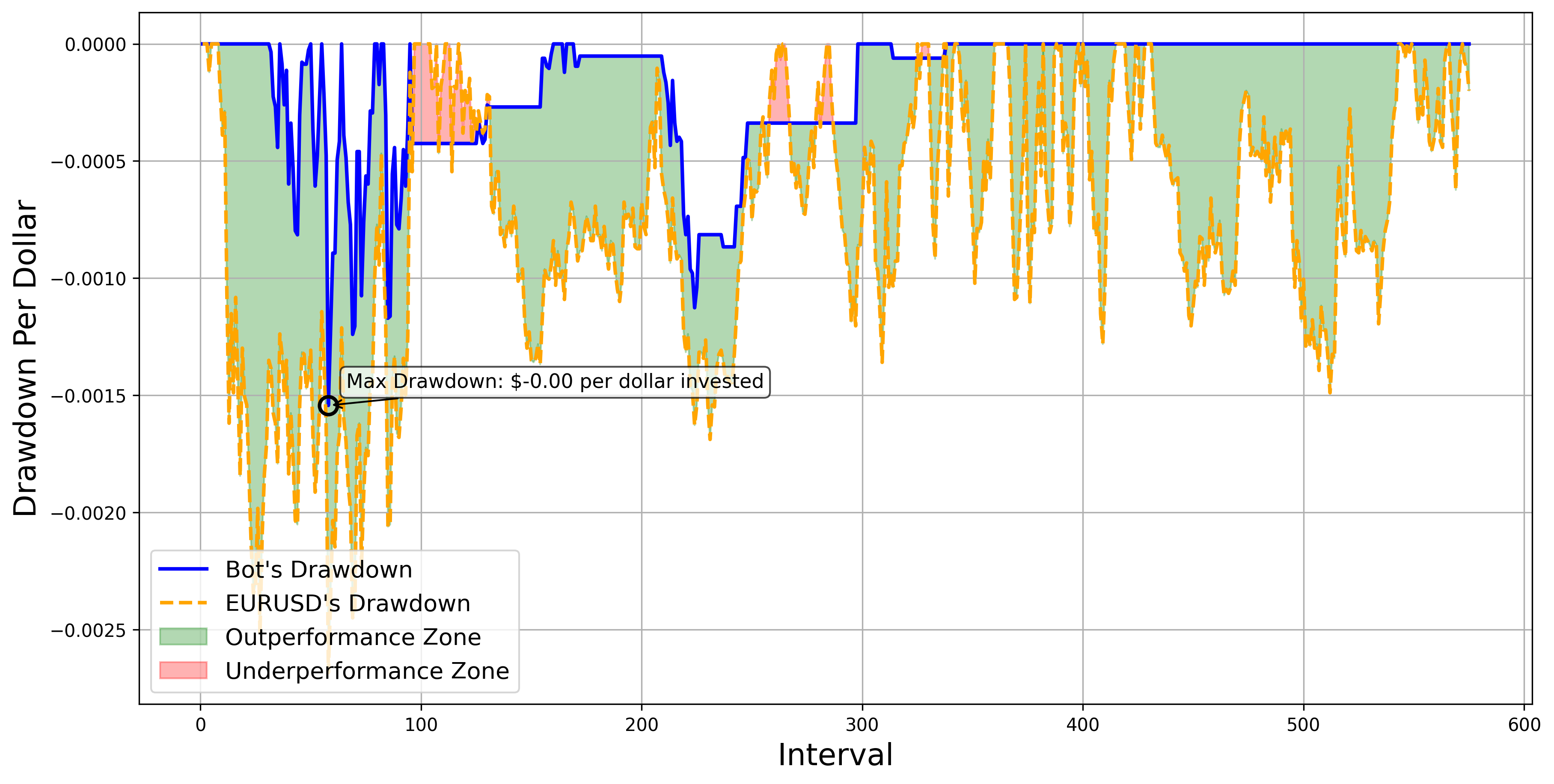}
        \caption{Drawdown}
    \end{subfigure}
    \hfill
    \begin{subfigure}[b]{0.48\textwidth}
        \includegraphics[width=\linewidth]{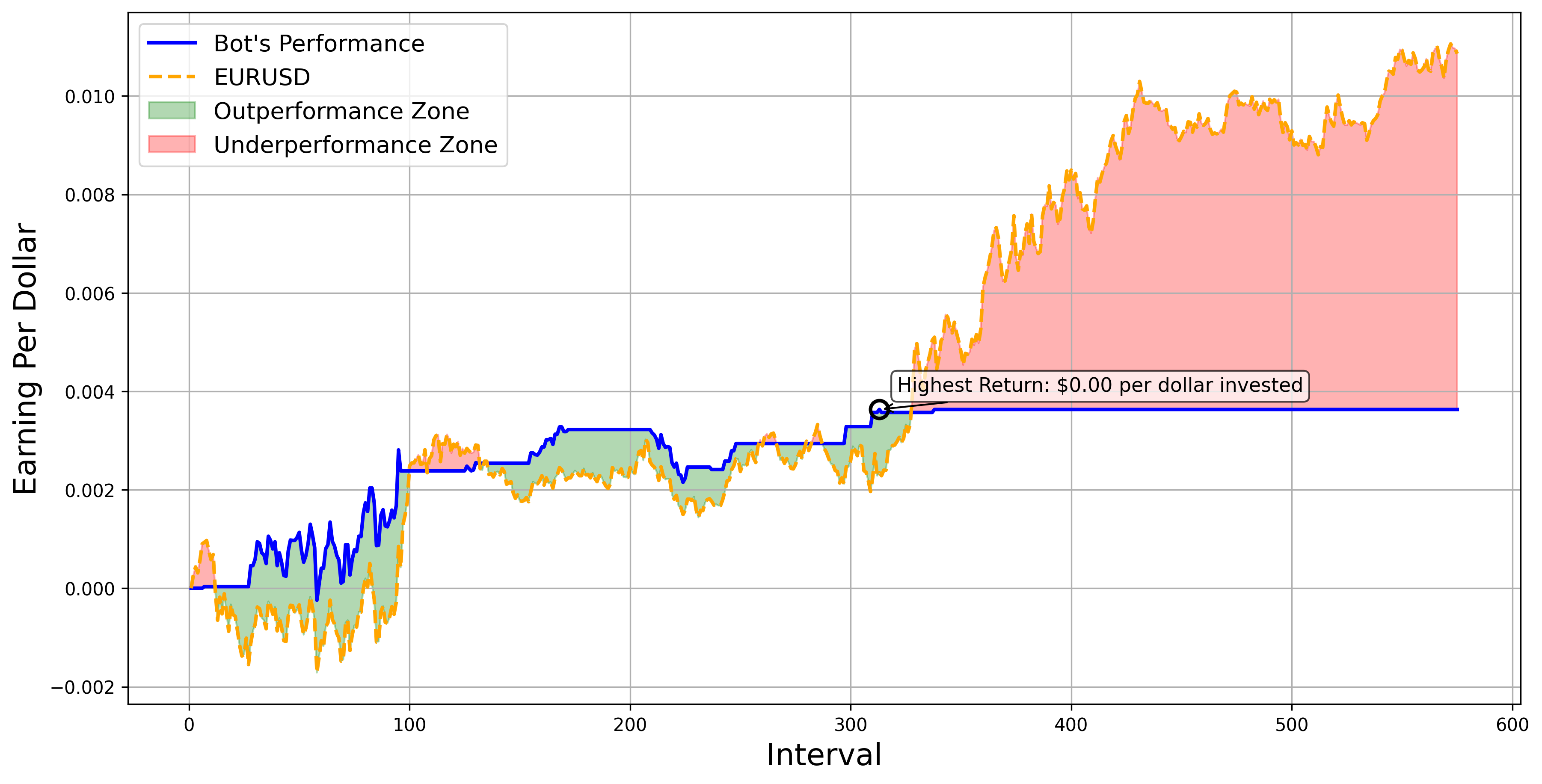}
        \caption{Portfolio}
    \end{subfigure}
    \caption{Chunk 9: Drawdown and Portfolio}
    \label{fig:eurusd-chunk9}
\end{figure}




\end{document}